\documentclass[journal]{IEEEtai}

\usepackage[colorlinks,urlcolor=blue,linkcolor=blue,citecolor=blue]{hyperref}
\usepackage{color,array}
\usepackage{graphicx}
\usepackage{cite}
\usepackage{amsmath,amssymb,amsfonts}
\usepackage{textcomp}
\def\BibTeX{{\rm B\kern-.05em{\sc i\kern-.025em b}\kern-.08em
    T\kern-.1667em\lower.7ex\hbox{E}\kern-.125emX}}
 
\usepackage{placeins} 
\usepackage{dblfloatfix}
\usepackage{ifpdf}
\usepackage{cite}
\usepackage{amstext}
\usepackage{amsthm}
\usepackage{footnote}
\usepackage{array}
\usepackage{subcaption}
\usepackage{float}
\usepackage{url}
\usepackage{cases}
\usepackage{footnote}
\usepackage{kantlipsum}
\usepackage{algpseudocode}
\usepackage{algorithm}
\usepackage{multirow}
\usepackage{capt-of}
\usepackage{color,soul}
\usepackage{balance}
\usepackage{booktabs}

\usepackage{tikz}
\newcommand\copyrighttext{%
  \footnotesize \textcopyright 2023 IEEE. Personal use of this material is permitted.
  Permission from IEEE must be obtained for all other uses, in any current or future
  media, including reprinting/republishing this material for advertising or promotional
  purposes, creating new collective works, for resale or redistribution to servers or
  lists, or reuse of any copyrighted component of this work in other works.
  DOI: \href{<http://tex.stackexchange.com>}{10.1109/TAI.2023.3234930}}
\newcommand\copyrightnotice{%
\begin{tikzpicture}[remember picture,overlay]
\node[anchor=south,yshift=5pt] at (current page.south) {\fbox{\parbox{\dimexpr\textwidth-\fboxsep-\fboxrule\relax}{\copyrighttext}}};
\end{tikzpicture}%
}

\theoremstyle{definition}

\begin{document}

\title{Towards Handling Uncertainty-at-Source in AI -- A Review and Next Steps for Interval Regression} 

\author{Shaily~Kabir, 
        Christian~Wagner,~\IEEEmembership{Senior Member,~IEEE, }
        and Zack~Ellerby
       
\thanks{S. Kabir, C. Wagner and Z. Ellerby are with the Lab for Uncertainty in Data and Decision Making (LUCID), School of Computer Science, University of Nottingham,
Nottingham, UK (e-mail: shaily.kabir3, christian.wagner, zack.ellerby@nottingham.ac.uk).}}

\markboth{Journal of IEEE Transactions on Artificial Intelligence, Vol. XX, No. X, Month XXXX}
{Kabir \MakeLowercase{\textit{et al.}}: Towards Handling Uncertainty-at-Source in AI - A Review and Next Steps for Interval Regression}

\maketitle

\copyrightnotice
\begin{abstract}
Most of statistics and AI draw insights through modelling discord or variance between sources (i.e., inter-source) of information. Increasingly however, research is focusing on uncertainty arising at the level of individual measurements (i.e., within- or intra-source), such as for a given sensor output or human response. Here, adopting intervals rather than numbers as the fundamental data-type provides an efficient, powerful, yet challenging way forward---offering systematic capture of uncertainty-at-source, increasing informational capacity, and ultimately potential for additional insight. Following progress in the capture of interval-valued data in particular from human participants, conducting machine learning directly upon intervals is a crucial next step. This paper focuses on linear regression for interval-valued data as a recent growth area, providing an essential foundation for broader use of intervals in AI. We conduct an in-depth analysis of state-of-the-art methods, elucidating their behaviour, advantages, and pitfalls when applied to synthetic and real-world data sets with different properties. Specific emphasis is given to the challenge of preserving mathematical coherence, i.e., models maintain fundamental mathematical properties of intervals. 
In support of real-world applicability of the regression methods, we introduce and demonstrate a novel visualization approach, the interval regression graph, or \emph{IRG}, which effectively communicates the impact of both position and range of variables within the regression models---offering a leap in their interpretability. Finally, the paper provides practical recommendations concerning regression-method choice for interval data and highlights remaining challenges and important next steps for developing AI with the capacity to handle uncertainty-at-source.\vspace{-1em}
\end{abstract}

\begin{IEEEImpStatement}
Capturing information as intervals provides a powerful means for handling uncertainty and inherent range in data. This paper focuses on a basic building block of statistics and AI: linear regression. In recent years, regression for intervals has become a topic of interest in AI and has been applied to domains ranging from marketing to cyber-security. It allows direct modelling of relationships not only between variables per-se, but also their associated uncertainty. For example, we can infer not only how a snack's nutritional benefits impact consumer purchase intention, but also how uncertainty about these benefits impacts purchase intention and its associated uncertainty. Nonetheless, while there are considerable upsides to interval regression and AI, substantial challenges remain. This paper reviews and extends state-of-the-art interval regression methods, presents in-depth experimental results and introduces a novel visualization approach for interval regression---enhances interpretability and facilitates real-world insight. Finally, it provides recommendations for algorithm selection based on key properties of data sets, discusses next steps for interval-valued AI beyond linear regression, and is complemented by an open-source implementation of the algorithms discussed.
\end{IEEEImpStatement}

\begin{IEEEkeywords}
Intervals, regression, uncertainty, intra-source, inter-source
\end{IEEEkeywords}

\section{Introduction}
\label{sec:introduction}
\IEEEPARstart{I}{nterval-valued} (IV) data have gained increasing attention across a variety of contexts as they offer a systematic way to capture information complete with an intrinsic representation of range or uncertainty in each individual `measurement', which is not possible using point-values, such as numbers or ranks. Such data may arise due to imprecision and uncertainty in measurement (e.g., sensor data), inherent uncertainty of outcome (e.g., estimations of stock prices ~\cite{lin2020symbolic}, climate/weather forecasting~\cite{sun2015normal}), or inherent vagueness or nuance (e.g., in linguistic terms~\cite{liu2008encoding}). A growing body of evidence also suggests that subjective judgement of humans (e.g., experts, consumers) may be better represented by intervals~\cite{wagner2015interval,ellerby2021capturing}, where the interval width/size captures the response range as a conjunctive set (e.g., the reals between 2 and 4), or degree of uncertainty with respect to an estimate or rating---a disjunctive set (e.g., confidence interval) \cite{kabir2020similarity}.

Regression analysis for estimating one or more variables is an important task in data analysis and machine learning under the wider umbrella of AI. While regression is most commonly applied to numeric data, researchers have begun to develop methods of applying (linear) regression analysis to IV data~\cite{brito2007modelling}. To date, models designed to handle data sets where \emph{both}  independent (regressor) and dependent (regressand) variables are IV have been developed within the frameworks of random sets and symbolic data analysis (SDA)~\cite{yeh2017tree,gonzalez2013constrained}. In the random set framework, intervals are viewed as compact convex sets in $\mathbb{R}$, and their interaction is modelled according to set arithmetic~\cite{gonzalez2007least,blanco2011estimation,boukezzoula2011midpoint,blanco2013set,garcia2020multiple}. On the other hand, in the SDA setting, the classical regression model is expanded to IV data where intervals are treated as bi-variate vectors~\cite{billard2000regression,billard2002symbolic,neto2008centre,neto2010constrained,wang2012linear,sun2015linear,souza2017parametrized,hao2017constrained}.  

Current linear regression models for intervals use different \emph{reference points}, such as the regressors' center values, their lower and upper bounds, or both their center and range (width), to estimate the IV  regressand~\cite{souza2017parametrized}. Earlier interval regression approaches~\cite{billard2000regression,billard2002symbolic,neto2008centre} risk `flipping' the interval bounds, respectively generating a negative interval range, thus breaking mathematical coherence (i.e., violating the definition of interval). The more recent approaches~\cite{neto2010constrained,wang2012linear,sun2015linear,souza2017parametrized} impose either positivity restrictions on parameters, or design the regression model so as to ensure that the lower bound does not exceed the upper bound. For instance, Neto and Carvalho~\cite{neto2010constrained} apply the Lawson and Hanson algorithm (LHA)~\cite{lawson1974},  Wang et al.~\cite{wang2012linear} use Moore’s linear combination~\cite{moore1966interval}, and  Souza et al.~\cite{souza2017parametrized} the Box-Cox transformation~\cite{box1964analysis} to guarantee mathematical coherence. Although Sun and Ralescu~\cite{sun2015linear} consider positivity restrictions on parameters in their model setting, they do not provide a method to maintain these restrictions. A detailed review and analysis of these models is provided in Section~\ref{sec:background} and the general issue is discussed throughout the paper. 

\begin{figure}[t!]
\begin{subfigure}{0.24\textwidth}
\includegraphics[width=0.95\textwidth]{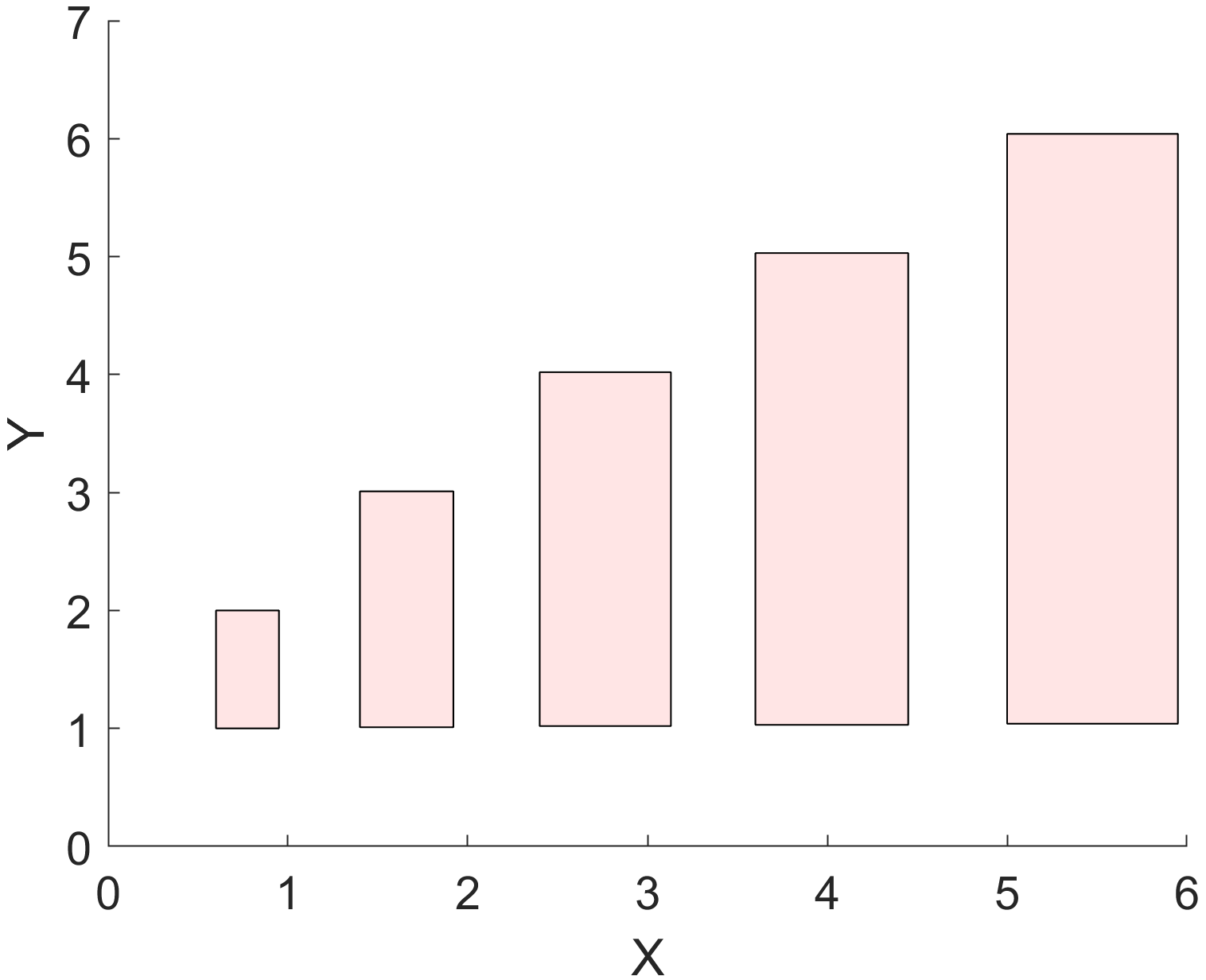}
\caption{Set-1}
\end{subfigure}
\begin{subfigure}{0.24\textwidth}
\includegraphics[width=0.98\textwidth]{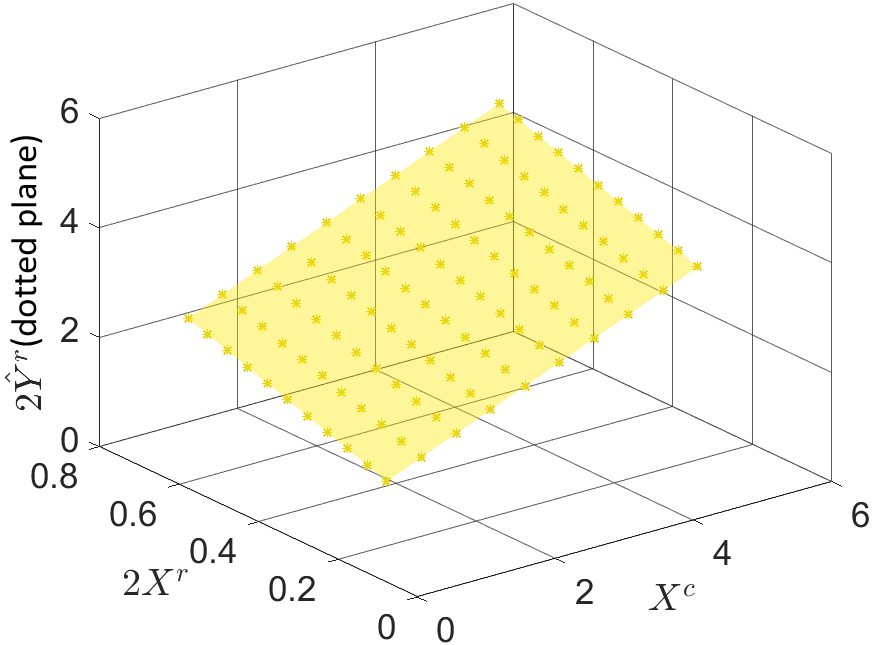}
\caption{\emph{IRG} for Set-1}
\end{subfigure}
\caption{(a) Visualization of IV data Set-1 and (b) \emph{IRG} with respect to range, $\hat{Y}^w$ ($\simeq 2\hat{Y}^r$) using the PM method \cite{souza2017parametrized}.}
\vspace{-0.8em}
\label{fig1}
\end{figure}
\begin{figure}[t!]
\begin{subfigure}{0.24\textwidth}
\includegraphics[width=0.95\textwidth]{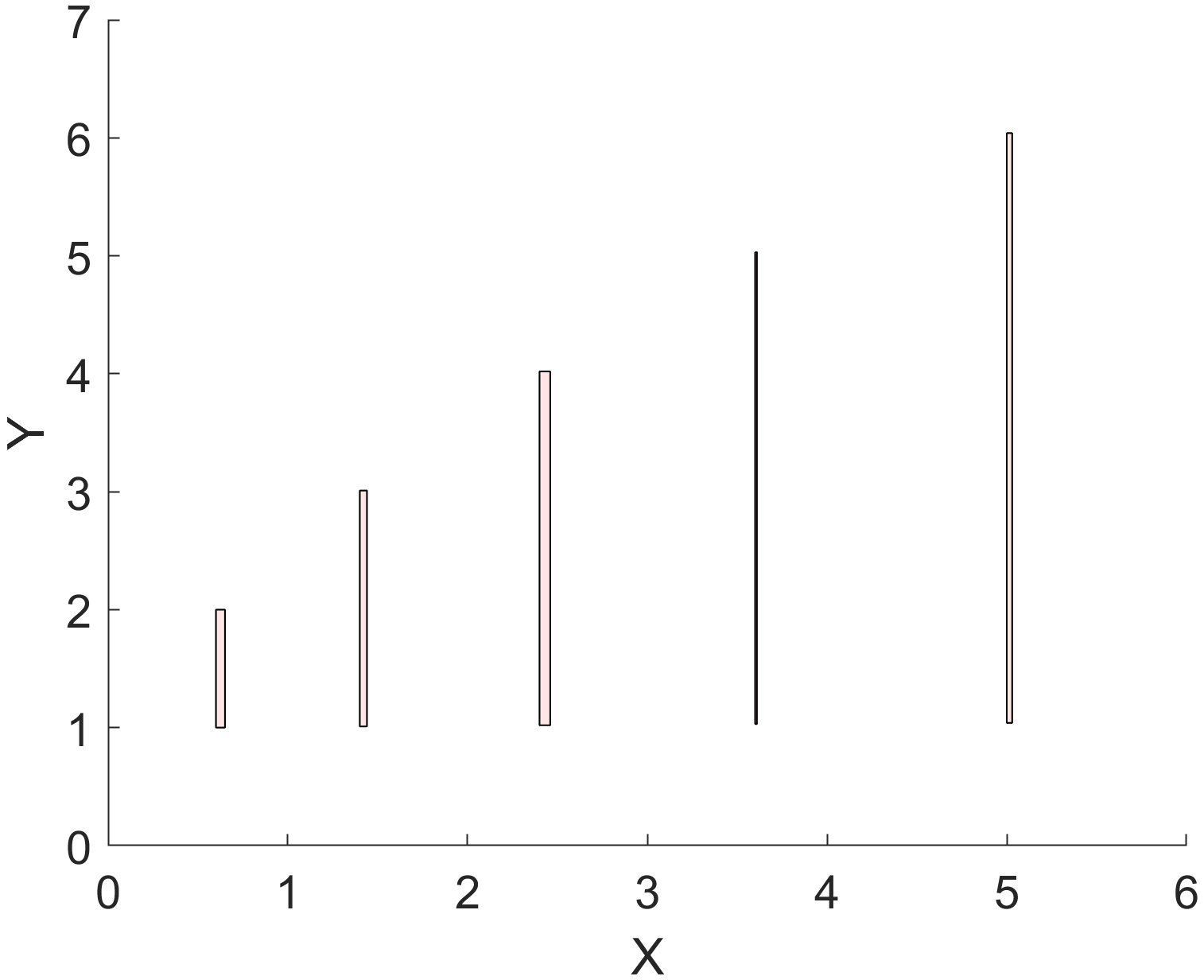}
\caption{Set-2}
\end{subfigure}
\begin{subfigure}{0.24\textwidth}
\includegraphics[width=0.98\textwidth]{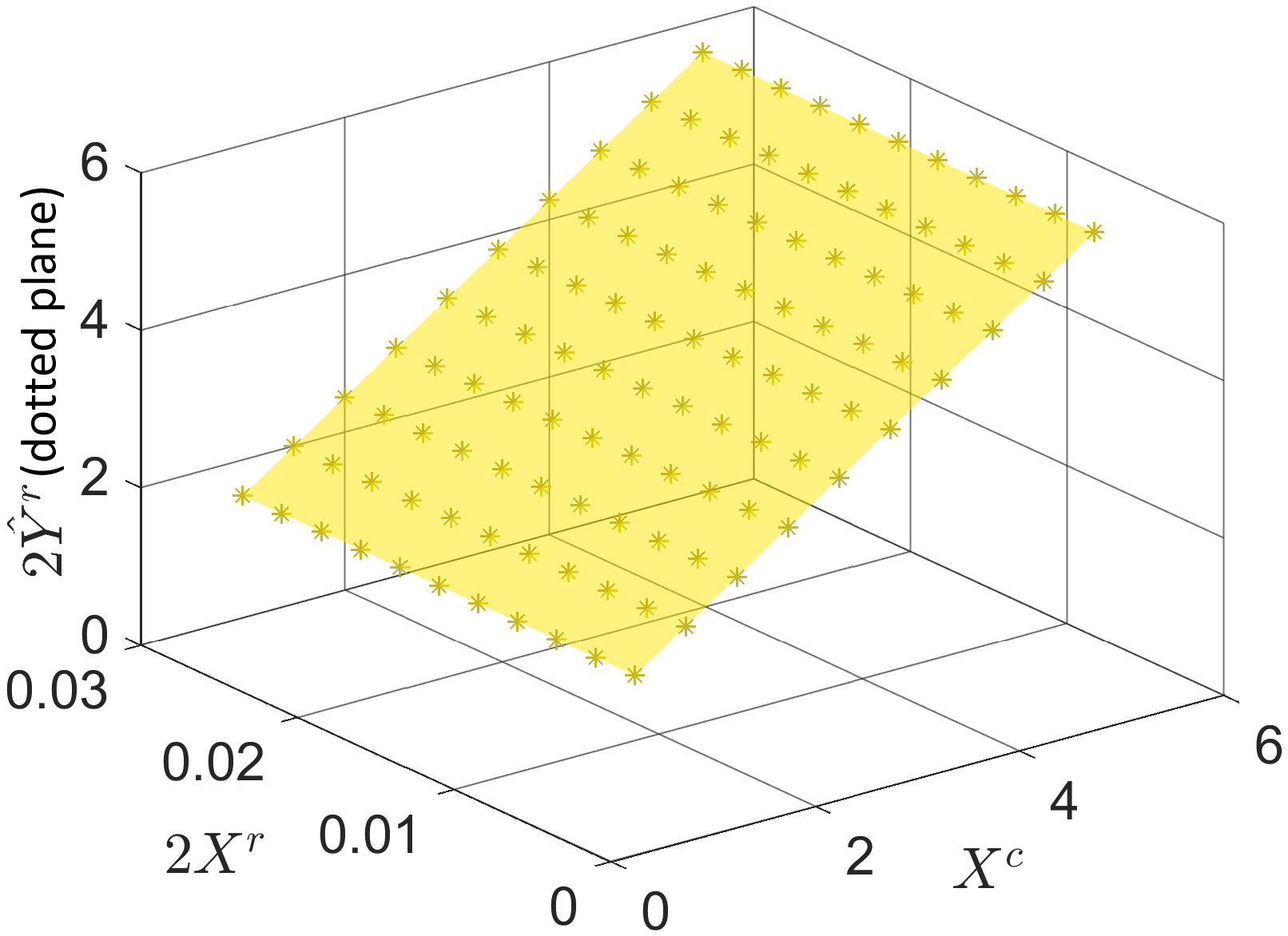}
\caption{\emph{IRG} for Set-2}
\end{subfigure}
\caption{(a) Visualization of IV data Set-2 and (b) \emph{IRG} with respect to range, $\hat{Y}^w$ ($\simeq 2\hat{Y}^r$) using the PM method \cite{souza2017parametrized}.}
\vspace{-1.8em}\label{fig11}
\end{figure}

An important aspect is that most current approaches, including the above, restrict parameters based on the assumption that an increase in uncertainty\footnote{Throughout the remainder of this paper, for simplicity, we refer to the range of the interval as representing uncertainty as a general term capturing vagueness, lack of information, etc.} of a regressor (i.e., input) leads to greater uncertainty in the dependent variable or regressand. However, there are plausible cases in the real world where this positive correlation does not hold. For example, in cyber-security, uncertainty around the vulnerability of a system may lead to certainty in the need to invest resources into controls---thus mitigating the risk of the system being compromised, that is, a negative correlation. Similarly, in a multivariate case, consider an economics example; greater uncertainty in future product prices and interest rates may cause more certainty with respect to present consumption choices.

For illustration, Figs.~\ref{fig1}(a) and~\ref{fig11}(a) show a synthetic example of two IV data sets. Here, in Set-1 (Fig.~\ref{fig1}(a)), the position (center) and range (uncertainty) of the regressand, $Y$ increases with the position (center) and range (e.g. the uncertainty) of regressor, $X$---that is, they vary in unison---this reflects the most common case as outlined above, e.g., the price of cars vs their horsepower. In contrast, for Set-2 (Fig.~\ref{fig11}(a)), the range (uncertainty) of the regressand, $Y$ increases irrespective of the range (uncertainty) of the regressor, $X$. This reflects a less common scenario, e.g., the wealth of people with respect to their age. The relationships between the variables in each figure are captured through Interval Regression Graphs (\emph{IRG}s) in Figs.~\ref{fig1}(b) and~\ref{fig11}(b) respectively. \emph{IRG}s will be introduced in detail in Section~\ref{sec:methodology.IRG}, but for now, note how Fig.~\ref{fig1}(b) shows a gentle increase in the range of the dependent variable ($2Y^r$) for an increase in the range of the independent variable ($2X^r$) for Set-1, while Fig.~\ref{fig11}(b) shows, as expected, no such change for Set-2. The \emph{IRG}s also highlight the much less drastic increase in range of Set-1's dependent variable that is attributed to an increase in the position of the independent variable ($X^c$), compared to Set-2, in which variance in regressand range ($2Y^r$) is attributed solely to regressor position ($X^c$).

Aspects such as the above highlight the complexity but also the potential of IV data in directly capturing uncertainty and range at-source, making it accessible for reasoning. With intervals encoding additional information by comparison to numbers, the choice of a `universally best' analysis approach, as can often be found for numeric cases, is however highly challenging. Instead, the selection of appropriate techniques for a given IV data set---based on its properties---is essential. Ferson~et~al. in~\cite{ferson2007experimental}, discuss the complexity of analysing real-world IV data in detail, elucidating the value of focusing on key properties for data sets to inform the choice of methods, such as whether the intervals exhibit strong overlap (i.e., `puffy'~\cite{ferson2007experimental}: wider, more uncertain intervals) or very little overlap (i.e., `skinny'~\cite{ferson2007experimental}: narrow, less uncertain); whether there are outliers; whether intervals are highly scattered, densely placed or even nested within each other. Beyond these properties, more fundamental questions arise from the semantic meaning of the given intervals, as already alluded to above; whether they represent an epistemic, disjunctive set, with the interval describing the range within which a correct answer is contained (this is arguably the most common use of intervals in science, cf.~\emph{confidence intervals}), or indeed an ontic, conjunctive set with a true range, such as the real numbers between 1 and 5, or a~time span~\cite{couso2014statistical,ellerby2021capturing}. For conciseness, we do not discuss the latter further in this manuscript, but will revisit it, and its impact on the analysis of AI and machine learning more generally, in future publications. 
    
At this stage, we also note that of course one can go beyond intervals to model uncertainty, leveraging distributions with a probabilistic or fuzzy interpretation, depending on the dis- or con-junctive nature of the information. Focusing on intervals a priori provides advantages, such as increasingly established pathways to collecting data (e.g., \cite{ellerby2021capturing}), while also laying foundations for more advanced methods. For example, in fuzzy set theory, $\alpha$-cut decomposition provides a direct translation mechanism between intervals and fuzzy sets. 

This paper makes four principal contributions:
\begin{enumerate}
\item \emph{Review of the state-of-the-art:} In view of the significance and use of intervals as a fundamental data-type in AI, the paper presents a detailed review of all well-known state-of-the-art linear regression approaches based on vector representation of data sets where \emph{both} independent and dependent variables are IV.  
\item \emph{Extensions of current methods:} As part of the review of the methods, this paper puts forward extensions to the Linear Model (LM)~\cite{sun2015linear}, enhancing its robustness efficiently for real-world applications. Specifically, we propose a dynamically selective approach to applying restrictions, letting the model's parameters vary freely unless mathematical coherence is being compromised. 
\item \emph{Practical guidance on algorithm selection and software:} To support the broader adoption of intervals as a fundamental data-type in AI, extant regression approaches and their performance are analysed using synthetic and real world data sets capturing various common and features of intervals, such as `puffy' and `skinny' intervals \cite{ferson2007experimental}. The resulting insights are distilled, for the first time, to guide readers in selecting the regression algorithms appropriate to \emph{their} data. Further, we provide open-source software at http://lucidresearch.org/software, offering, for the first time, direct access to IV regression approaches, with a view to facilitating adoption and replication.
\item \emph{Explainability and visualization:} The explainability of numeric linear regression models is one of their primary assets as an AI technique. This paper introduces a novel visualization approach for IV regression---the Interval Regression Graph (\emph{IRG})---which for the first time provides the means to visualize and succinctly communicate the relationship in terms of both position and range between IV regressor and regressand.\footnote{Note that in this paper, we focus on establishing and visualizing the relationship between the variables of a given IV data set. We do not focus on \emph{prediction}, a different problem setting with implications for example the need to establish model and prediction robustness, e.g., using cross-validation~\cite{picard1984cross}---offering scope for future research for IV data.\vspace{-1.8em}} 
\end{enumerate}

\begin{table}
\centering
\caption{Acronyms and Notation}
\begin{tabular}{ll} \hline\hline
CM & Center Method \cite{billard2000regression}\\
MinMax & MinMax Method \cite{billard2002symbolic}\\
CRM & Center and Range Method \cite{neto2008centre}\\
CCRM & Constrained Center and Range Method \cite{neto2010constrained}\\
CIM & Complete Information Method \cite{wang2012linear}\\
LM & Linear Model \cite{sun2015linear}\\
PM & Parametrized Model \cite{souza2017parametrized}\\
LM$_c$ & Constrained Linear Model \\
LM$_w$ & Weakly Constrained Linear Model\\
IV & Interval-valued\\
IRG & Interval Regression Graph\vspace{0.5mm}\\\hline
$\overline{a}$ & Interval $\{\overline{a}\subseteq\mathbb{R}:\overline{a}=[a^-,a^+],~a^-\leq a^+\}$ \\
$a^w$ & Range of Interval, $a^w=|a^+-a^-|$ \\
$a^c$ & Center of Interval, $a^c=\frac{(a^++a^-)}{2}$  \\
$a^r$ & Radius of Interval $a^r=\frac{a^w}{2}$ (Half Range)\\
$Y$ & Interval-valued Regressand\\
$\hat{Y}$ & Estimated $Y$\\
$X$ & Interval-valued Regressor\\ 
$\mu_{Y^w}$ and $\sigma_{Y^w}$ & Mean and Standard Deviation of Range of $Y$\\
$\mu_{X^w}$ and $\sigma_{X^w}$ & Mean and Standard Deviation of Range of $X$\\
$\mu_{Y^c}$ and $\sigma_{Y^c}$ & Mean and Standard Deviation of Center of $Y$\\
$\mu_{X^c}$ and $\sigma_{X^c}$  & Mean and Standard Deviation of Center of $X$\\
\hline
\end{tabular}
\vspace{-1.8em}
\label{tab:acronym}
\end{table}

The paper is organized as follows: Section~\ref{sec:background} reviews linear regression methods based on vector/matrix representation of intervals in the regression process. Section~\ref{sec:methodology} introduces the interval regression graph (\emph{IRG}) and discusses adaptations of the LM method~\cite{sun2015linear} for real-world applications, followed by IV data sets (synthetic and real-world) and evaluation metrics. Section~\ref{sec:results} demonstrates the estimation performance of all regression models with respect to both synthetic and real-world data sets. Section~\ref{sec:discussion} discusses suitability of the existing regression approaches to salient features of IV data, distilling insights into practical guidance for algorithm-selection. Lastly,  Section~\ref{sec:conclusion} concludes the paper with future works. Table~\ref{tab:acronym} presents a list of acronyms and notation used in this paper to assist the reader.

\section{Background and State of the Art}
\label{sec:background}
 A closed interval $\overline{a}$ is characterized by its two endpoints, $a^-$ and $a^+$ with $a^-\leq a^+$ where $a^-$ and $a^+$ are respectively the
lower and upper bounds of $\overline{a}$~\cite{Beliakov2014penalty}. Generally, $\overline{a}$ is denoted as $\overline{a}=[a^-,a^+]$, and its range is given by $a^w=\left|a^+-a^-\right|$. Alternatively, the same interval can be represented as $\overline{a}=[a^c-a^r,a^c+a^r]$ with its center, $a^c=\frac{(a^+ + a^-)}{2}$ and radius, $a^r=\frac{a^w}{2}$. In this section, we review linear regression methods for \emph{full} IV data sets (i.e. where both dependent and independent variables are intervals) where intervals are treated as bi-variate vectors in the regression process. We note $Y=\{\overline{y}_1,\overline{y}_2,\dots,\overline{y}_n\}$ as an IV regressand with $n$ intervals, where $\overline{y}_i=[y_i^-,y_i^+]$, $1\leq i\leq n$. Further, $\{X_1,X_2,\dots,X_p\}$ presents $p$ ($\geq1$) IV regressor(s) where each $X_j$ has also $n$ intervals, $X_j=\{\overline{x}_{j1},\overline{x}_{j2},\dots,\overline{x}_{jn}\}$ with $\overline{x}_{ji} = [x_{ji}^-,x_{ji}^+]$, $1\leq i\leq n$, $1\leq j\leq p$. Thus, there are $p$ pairs of ($Y,X$) in the data set. In addition, the estimated value of $Y$ is given by  $\hat{Y}=\{\hat{\overline{y}}_1,\hat{\overline{y}}_2,\dots,\hat{\overline{y}}_n\}$.

Since about the year 2000, a variety of linear regression approaches for \emph{full} IV data sets have been put forward. We briefly review them below in chronological order.

\subsection{The Center Method (2000)}
\label{CM_subsection}
Billard and Diday~\cite{billard2000regression} proposed a linear regression model, known as the Center Method (CM) in 2000, to fit the center of both regressand and regressors, and then apply this model to the lower and upper bounds of regressors to separately estimate the upper and lower bounds of the regressand. Equation~\eqref{CM_lower_upper} defines the CM regression model, 
\begin{equation}\label{CM_lower_upper}
\begin{split}
    y_i^-=\beta_0^c+\sum_{j=1}^p\beta_j^c x_{ji}^-+\epsilon_i^-\\
    y_i^+=\beta_0^c+\sum_{j=1}^p\beta_j^c x_{ji}^++\epsilon_i^+
\vspace{-0.8em}\end{split}
\end{equation}
where $y_i^-$ and $y_i^+$ are the lower and upper bounds of $\overline{y}_i$. Similarly,  $x_{ji}^-$ and $ x_{ji}^+$ are the the lower and upper bounds of $\overline{x}_{ji}$. $\beta_j^c\in \mathbb{R}$ are the regression coefficients for the center estimation, $0\leq j\leq p$, $p$ is the total number of regressors  while $\epsilon_i^-$ and $\epsilon_i^+$ are error terms. Both lower and upper bound models of~\eqref{CM_lower_upper} use the same coefficients, therefore, they can be captured together using~\eqref{CM_center}. 
\begin{equation}\label{CM_center}
    y_i^c=\beta_0^c+\sum_{j=1}^p\beta_j^c x_{ji}^c+\epsilon_i^c, \vspace{-0.4em}
\end{equation}
where $y_i^c$ and $x_{ji}^c$ are the center of $\overline{y}_i$ and $\overline{x}_{ji}$ respectively and $\epsilon_i^c=(\epsilon_i^-+\epsilon_i^+)/2$ is error term. Equation~\eqref{CM_center} can also be expressed for the total set of $n$ observations in the matrix format as $Y^c=X^{c}\beta^c+\epsilon^c$, where 
\resizebox{.94\linewidth}{!}{
  \begin{minipage}{0.99\linewidth}
\begin{align*}\label{CM_matrix_form}
    \begin{split}
     Y^c=(y_1^c\quad y_2^c\quad \dots\quad y_n^c)^T,\quad \beta^c=(\beta_0^c\quad \beta_1^c\quad\dots\quad\beta_p^c)^T\\
     \epsilon^c=(\epsilon_1^c\quad \epsilon_2^c\quad\dots\quad\epsilon_n^c)^T, \text{ and } X^c=\begin{pmatrix}
                     1 & x_{11}^c & \dots & x_{p1}^c\\
                     1 & x_{12}^c & \dots & x_{p2}^c\\
                     \vdots & \vdots & \dots &\vdots\\
                     1 & x_{1n}^c &\dots & x_{pn}^c\\
               \end{pmatrix}.\vspace{-0.8em}
\end{split}
\end{align*}
\end{minipage}
 }
 
The coefficients, $\beta^c$ are estimated using the least squares (LS) method~\cite{draper1998applied}, defined in~\eqref{eq:Beta_CM}. 
\begin{equation}\label{eq:Beta_CM}
    \hat{\beta}^c=((X^c)^T X^c)^{-1}(X^c)^TY^c 
\end{equation}
By using $\hat{\beta}^c$, ~\eqref{Estimated_CM_lower_upper} estimates the bounds of regressand, $Y$.
\begin{equation}\label{Estimated_CM_lower_upper}
\begin{split}
    \hat{y}_i^-=\hat{\beta}_0^c+\sum_{j=1}^p\hat{\beta}_j^c x_{ji}^-\\
    \hat{y}_i^+=\hat{\beta}_0^c+\sum_{j=1}^p\hat{\beta}_j^c x_{ji}^+.\vspace{-0.8em}
\end{split}
\end{equation}

The CM model is simple, as it follows a standard regression approach applied to the center of the intervals and then the resulting model is applied both to upper and  lower endpoints. However, by doing so, it risks and often fails to maintain mathematical coherence \cite{neto2008centre}, that is, the estimated lower bound of $\overline{y}_i$ becomes larger than the estimated upper bound. 

\subsection{The MinMax Method (2002)} \label{MM_subsection}
Billard and Diday~\cite{billard2002symbolic} presented the MinMax method in 2002 which directly uses the lower and upper bounds of the regressors to separately estimate the lower and upper bounds of the regressand. Equation~\eqref{MinMax} defines the MinMax regression model for the lower and upper bounds of the regressand, \begin{equation}\label{MinMax}
\begin{split}
    y_i^-=\beta_0^-+\sum_{j=1}^p\beta_j^- x_{ji}^-+\epsilon_i^-\\
    y_i^+=\beta_0^++\sum_{j=1}^p\beta_j^+ x_{ji}^++\epsilon_i^+  \vspace{-0.8em}
\end{split}
\end{equation}
where $\beta_j^-$ and $\beta_j^+$ are regression coefficients for the lower and upper bounds respectively ($\beta_j^-,\beta_j^+\in \mathbb{R}$, $0\leq j\leq p$). In matrix notation, the lower bound model of~\eqref{MinMax} with all $n$ observations is expressed as $Y^-=X^{-}\beta^-+\epsilon^-$, where 

\resizebox{.94\linewidth}{!}{
  \begin{minipage}{0.99\linewidth}
 \begin{align*}\label{MinMax_lower_matrix_form}
    \begin{split}
      Y^-=(y_1^-\quad y_2^-\quad \dots \quad y_n^-)^T,\quad \beta^-=(\beta_0^-\quad \beta_1^-\quad \dots \quad\beta_p^-)^T,\\
      \epsilon^-=(\epsilon_1^-\quad \epsilon_2^-\quad \dots \quad\epsilon_n^-)^T, \text{ and } X^-=\begin{pmatrix}
                     1 & x_{11}^- & \dots & x_{p1}^-\\
                      1 & x_{12}^- & \dots & x_{p2}^-\\
                     \vdots & \vdots & \dots &\vdots\\
                      1 & x_{1n}^- &\dots & x_{pn}^-\\
               \end{pmatrix}.
 \end{split}
 \end{align*}
 \end{minipage}
 }
 
Like the CM method, \eqref{eq:Beta_MinMax_lower} estimates the coefficient ($\beta^-$) for the lower bound model of~\eqref{MinMax} using the LS method~\cite{draper1998applied}.
\begin{equation}\label{eq:Beta_MinMax_lower}
    \hat{\beta}^-=((X^-)^T X^-)^{-1}(X^-)^TY^- 
\end{equation}
Now, using the estimated $\hat{\beta}^-$, \eqref{Estimated_MinMax_lower} estimates the lower bounds for the regressand, $Y$.
\begin{equation}\label{Estimated_MinMax_lower}
    \hat{y}_i^-=\hat{\beta}_0^-+\sum_{j=1}^p\hat{\beta}_j^- x_{ji}^-
\end{equation}

Similarly, the upper bound model of \eqref{MinMax} is expressed as $Y^+=X^{+}\beta^{+}+\epsilon^+$, where 

 \resizebox{0.94\linewidth}{!}{
   \begin{minipage}{\linewidth}
 \begin{align*}
     \begin{split} 
     Y^+=(y_1^+\quad y_2^+\quad\dots\quad y_n^+)^T,\quad \beta^+=(\beta_0^+\quad \beta_1^+\quad\dots\quad\beta_p^+)^T,\\
      \epsilon^+=(\epsilon_1^+\quad \epsilon_2^+\quad\dots\quad\epsilon_n^+)^T, \text{ and } X^+=\begin{pmatrix}
                      1 & x_{11}^+ & \dots & x_{p1}^+\\
                      1 & x_{12}^+ & \dots & x_{p2}^+\\
                      \vdots & \vdots & \dots &\vdots\\
                      1 & x_{1n}^+ &\dots & x_{pn}^+\\
               \end{pmatrix}.
 \end{split}
 \end{align*}
 \end{minipage}
 }
 
Equation~\eqref{eq:Beta_MinMax_upper} estimates the coefficients for the upper bound of \eqref{MinMax} by the LS method~\cite{draper1998applied}.  
\begin{equation}\label{eq:Beta_MinMax_upper}
    \hat{\beta}^+=((X^+)^T X^+)^{-1}(X^+)^TY^+
\end{equation}
Now, using $\hat{\beta}^+$, \eqref{Estimated_MinMax_upper} estimates the upper bounds of the regressand, $Y$.
\begin{equation}\label{Estimated_MinMax_upper}
      \hat{y}_i^+=\hat{\beta}_0^++\sum_{j=1}^p\hat{\beta}_j^+ x_{ji}^+
\end{equation}

In the MinMax method, the use of two separate models to estimate the lower and upper bounds of the regressand improves the model fitness and interpretation compared to the CM method~\cite{neto2008centre}. However, it still does not guarantee mathematical coherence on the estimated bounds~\cite{neto2008centre}. Further, its estimation performance can be reduced if there is no dependency between the bounds of regressand and regressors~\cite{souza2017parametrized}.

\subsection{The Center and Range Method (2008)}
\label{CRM_subsection}
Neto and Carvalho \cite{neto2008centre} extended the CM method \cite{billard2000regression} in 2008 to the Center and Range Method (CRM), by considering both interval center and half the `range' (i.e., the radius) of the regressor and regressand variables. They build two separate regression models---one for the centers and other for the ranges (or radii). Equation~\eqref{CRM_center} is used to model the interval centers.
\begin{equation}\label{CRM_center}
    y_i^c=\beta_0^c+\sum_{j=1}^p\beta_j^c x_{ji}^c+\epsilon_i^c
\end{equation}
where $y_i^c$ and $x_{ji}^c$ are the centers $\overline{y}_i$ and $\overline{x}_{ji}$ respectively and $\epsilon_i^c$ is error term. The CRM method follows the same matrix format of the CM method to represent the center model of~\eqref{CRM_center} and applies the same strategy of the CM method to estimate the coefficients $\beta_j^c$ (see~\eqref{eq:Beta_CM}). It then estimates the center values for the regressand, $Y$ using the estimated $\hat{\beta}_j^c$ by~\eqref{Estimated_CRM_center}.
\begin{equation}\label{Estimated_CRM_center}
      \hat{y}_i^c=\hat{\beta}_0^c+\sum_{j=1}^p\hat{\beta}_j^c x_{ji}^c
\end{equation}

Similarly, the interval ranges are modelled by the CRM method using~\eqref{CRM_range}, \begin{equation}\label{CRM_range}
    y_i^r=\beta_0^r+\sum_{j=1}^p\beta_j^r x_{ji}^r+\epsilon_i^r,
\end{equation}
where $y_i^r$ and $x_{ji}^r$ are the radius of $\overline{y}_i$ and $\overline{x}_{ji}$ respectively. $\beta_j^r\in\mathbb{R}$ are the coefficients for the range estimations, $0\leq j\leq p$ and $\epsilon_i^r$ is error term. Again, following the same matrix notation, the coefficients, $\beta_j^r$ are estimated by~\eqref{eq:Beta_CRM_range}, 
\begin{equation}\label{eq:Beta_CRM_range}
    \hat{\beta}^r=((X^r)^T X^r)^{-1}(X^r)^TY^r
\end{equation}
and using the estimated $\hat{\beta}^r$, \eqref{Estimated_CRM_range} computes the ranges of the regressand, $Y$.
\begin{equation}\label{Estimated_CRM_range}
      \hat{y}_i^r=\hat{\beta}_0^r+\sum_{j=1}^p\hat{\beta}_j^r x_{ji}^r
\end{equation}

Finally, the CRM method estimates the bounds of the regressand from the estimated $\hat{y}_i^c$ and $\hat{y}_i^r$ using~\eqref{CRM_lower_upper}.  
\begin{equation}\label{CRM_lower_upper}
    \hat{y}_i^-=\hat{y}_i^c-\hat{y}_i^r \text{  and  }  \hat{y}_i^+=\hat{y}_i^c+\hat{y}_i^r.
\end{equation} 
                                                 
The CRM model provides better estimation when there is a linear dependency between the ranges of regressand and regressors~\cite{neto2010constrained,souza2017parametrized}. If this is not the case, it does not ensure mathematical coherence as~\eqref{Estimated_CRM_range} can produce negative $y_i^r$~\cite{souza2017parametrized}. 

\subsection{The Constrained Center and Range Method (2010)} \label{CCRM_subsection}
Neto and Carvalho~\cite{neto2010constrained} later refined the CRM model~\cite{neto2008centre} (see Section~\ref{CRM_subsection}) in 2010 to ensure mathematical coherence. The resulting approach is known as the Constrained Center and Range Method (CCRM), where a positivity restriction is enforced on the coefficients which are estimated with respect to the relationship of the radii of regressand and regressor variables. In other words, the overall estimation process in the CCRM method remains the same as for the CRM model (see \eqref{CRM_center} to \eqref{CRM_lower_upper}) with positivity constraints on the range coefficients $\beta_j^r$. The CCRM model uses an iterative algorithm proposed by Lawson and Hanson in~\cite{lawson1974} to ensure $\beta_j^r\geq0$. For more detail, we refer the reader to~\cite{neto2010constrained}. Finally, the CCRM method estimates the bounds of the regressand using~\eqref{CRM_lower_upper}. 

Neto and Carvalho recommend to apply the CRM model in all cases, only adopting the CCRM method as a suitable strategy when the CRM method fails to maintain mathematical coherence, as the use of the CCRM model can lead to biased estimation outcomes~\cite{neto2010constrained}. In particular, the positivity restriction within the CCRM method forces any negative range coefficient to 0 (at the same time updating the remaining range coefficients), in turn leading to potentially biased estimation outcomes and poor model fitness~\cite{neto2010constrained}.

\subsection{The Complete Information Method (2012)}
\label{CIM_subsection}
Wang et al.~\cite{wang2012linear} presented the Complete Information Method (CIM) in 2012 with a focus on considering all the information contained within the intervals. The CIM model considers each interval observation of regressand and regressor variables as a hyper-cube (if a single-valued regressor and associated regressand are considered, each observation effectively reflects a rectangular, as visualized in Fig.~\ref{fig:synthetic_data_set_A}), and the regression model is built on these hyper-cubes. Further, it adopts Moore’s linear combination algorithm~\cite{moore1966interval} to ensure the mathematical consistency of the bounds. In this model, $Y$ is presented by a linear combination of $X_{j}$, 
\begin{equation}\label{CIM}
Y=\beta_0 1_n+\sum_{j=1}^p\beta_j X_{j} +\epsilon_i,
\end{equation}
where $\beta_j$ represents the coefficients with $0\leq j\leq p$ and $1_n$ is an n-dimensional column constant vector of ones. Equation~\eqref{CIM_lower_upper} defines the lower and upper bound estimations of $Y$.
\begin{equation}\label{CIM_lower_upper}
\begin{split}
    y_i^-=\beta_0+\sum_{j=1}^p\beta_j\left(\tau_j x_{ji}^-+(1-\tau_j)x_{ji}^+\right)\\
    y_i^+=\beta_0+\sum_{j=1}^p\beta_j\left((1-\tau_j) x_{ji}^-+\tau_j x_{ji}^+\right)
\end{split}
\end{equation}
where,
\[\tau_j = \left\{
  \begin{array}{lr}
    0, & \text{if } \beta_j\leq 0\\
    1, & \text{otherwise. }
  \end{array}\right.
\]
Using~\eqref{CIM}, the LS method~\cite{draper1998applied} of~\eqref{CIM_lower_upper} generates the following linear system, 
\begin{equation}\label{CIM1}
    M\beta = b
\end{equation}
where,
\begin{equation*}
M=\begin{pmatrix}
         \langle 1,1\rangle & \langle 1,X_1\rangle& \dots & \langle 1,X_p\rangle\\
         \langle X_1,1\rangle & \langle X_1,X_1\rangle& \dots & \langle X_1,X_p\rangle\\
          \vdots & \vdots & \dots & \vdots\\
         \langle X_p,1\rangle& \langle X_p,1\rangle&\dots & \langle X_p,X_p\rangle\\
   \end{pmatrix},
\end{equation*}
\begin{equation*}
 \beta=\begin{pmatrix}
         \beta_0\\
         \beta_1\\
         \vdots\\
         \beta_p\\
       \end{pmatrix}, 
 \text{ and } 
 b=\begin{pmatrix}
      \langle 1,Y\rangle\\
      \langle X_1,Y\rangle\\
      \vdots\\
      \langle X_p,Y\rangle\\
   \end{pmatrix}. 
\end{equation*}
\noindent Here, $\langle X_a,X_b\rangle$ is the inner product between $X_a$ and $X_b$ ($0\leq a,b\leq p$ and $X_0=1$)  which is computed using~\eqref{CIM_More}.
\begin{equation}\label{CIM_More}
\resizebox{.48\textwidth}{!}{$
 \langle X_a,X_b\rangle = \left\{
       \begin{array}{lr}
        \frac{1}{3}\sum\limits_{i=1}^n\left(x^{-2}_{ai}+x^{-}_{ai} x^{+}_{ai}+x^{+2}_{ai}\right), & \text{if } X_a=X_b\\
       \frac{1}{4}\sum\limits_{i=1}^n\left(x^{-}_{ai}x^{-}_{bi} +x^{-}_{ai}x^{+}_{bi} +x^{+}_{ai}x^{-}_{bi}+x^{+}_{ai}x^{+}_{bi}\right), & \text{if } X_a\neq X_b
  \end{array}\right. $}
\end{equation}
From~\eqref{CIM1}, the coefficients $\beta$ can be estimated by the inverse of matrix $M$ as defined in~\eqref{CIM11}.
\begin{equation}\label{CIM11}
    \hat{\beta} = (M)^{-1} b
\end{equation}
Thus, using $\hat{\beta}$ and $\hat{\tau}_j$, the lower and upper bounds of the regressand, $Y$ can be estimated by~\eqref{CIM_lower_upper}.

The CIM method satisfies mathematical coherence of the estimated bounds of the regressand using $\tau_j$ where its value (1 or 0) depends on the sign (+ or -) of $\hat{\beta}_j$~\cite{wang2012linear}.   

\subsection{The Linear Model (2015)}\label{LM_subsection}
Sun and Ralescu~\cite{sun2015linear} developed the Linear Model (LM) in 2015 based on the affine operator in the cone $\mathcal{C}=\{(x,y)\in \mathbb{R}^2|x\leq y\}$, aiming to maximize model flexibility while preserving its interpretability. This model considers both lower and upper bounds of regressors and their ranges for estimating the bounds of the regressand. In this model, $\overline{y}_i$ is expressed as a linear transformation of $\overline{x}_{ji}$ such that, 
\begin{align}\centering
\begin{split}\label{LM}
y_i^-&=\sum_{j=1}^p\left(\alpha_j x_{ji}^-+\beta_j x_{ji}^+\right)+\eta+\epsilon_i^-,\\
y_i^+&=\sum_{j=1}^p\left((\alpha_j-\gamma_j) x_{ji}^-+(\beta_j+\gamma_j) x_{ji}^+\right)+\eta+\theta+\epsilon_i^+,
\end{split}
\end{align}
where the regressor coefficients, $\alpha_j,\beta_j, \eta\in \mathbb{R}$ and $\gamma_j,\theta\ge0$, $j=1,2,...,p$. Equation~\eqref{LM} can also be written in the following matrix form as $Y=X\beta+\epsilon$, where 

$Y=\begin{pmatrix}
        Y^-\\
        Y^+
        \end{pmatrix}$ with $Y^-=\begin{pmatrix}
        y_1^-\\
        \vdots \\
        y_n^-
        \end{pmatrix}$ and $Y^+=\begin{pmatrix}
        y_1^+\\
        \vdots \\
        y_n^+
        \end{pmatrix}$,
        
\resizebox{.45\textwidth}{!}{$X=\begin{pmatrix}
        X^* & 0\\
        X^* & X^w
        \end{pmatrix} \text{ with } X^*=\begin{pmatrix}
        1 & x_{11}^-& x_{11}^+&\dots & x_{p1}^+\\
        1 & x_{12}^-& x_{12}^+&\dots & x_{p2}^+\\
        \vdots & \vdots & \vdots&\vdots &\vdots \\
        1 & x_{1n}^-&x_{1n}^+&\dots & x_{pn}^+
        \end{pmatrix}$}  
        \text{ and } $X^w=\begin{pmatrix}
        1 & x_{11}^w& \dots & x_{p1}^w\\
        1 & x_{12}^w& \dots & x_{p2}^w\\
        \vdots & \vdots &\vdots &\vdots \\
        1 & x_{1n}^w&\dots & x_{pn}^w
        \end{pmatrix}$, and  $\beta=\begin{pmatrix}
        \beta^* \\
        \beta^w 
        \end{pmatrix}$ with
        
        $\beta^*=\begin{pmatrix}
        \eta \\
        \alpha_1\\
        \beta_1\\
        \vdots \\
        \alpha_p\\
        \beta_p
        \end{pmatrix}$ and $\beta^w=\begin{pmatrix}
        \theta \\
        \gamma_1\\
        \vdots \\
        \gamma_p
        \end{pmatrix}$.

Again, the LS estimate of the coefficients, $\hat{\beta}$ are computed by~\eqref{eq:Beta_Linear}, 
\begin{equation}\label{eq:Beta_Linear}
    \hat{\beta}=(X^T X)^{-1}X^TY
\end{equation}
and~\eqref{Estimated_Linear} estimates the bounds of the regressand, $Y$.
\begin{align}\label{Estimated_Linear}\centering
\begin{split}
\hat{y}_i^-&=\sum_{j=1}^p\left(\hat{\alpha}_j x_{ji}^-+\hat{\beta}_j x_{ji}^+\right)+\hat{\eta},\\
\hat{y}_i^+&=\sum_{j=1}^p\left((\hat{\alpha}_j-\hat{\gamma}_j) x_{ji}^-+(\hat{\beta}_j+\hat{\gamma}_j) x_{ji}^+\right)+\hat{\eta}+\hat{\theta}
\end{split}
\end{align}

Even though the authors assume positive constraints on range coefficients (i.e., $\gamma_j\ge0$ and $\theta\ge0$) in~\eqref{LM} to maintain mathematical coherence, the actual model does not ensure compliance with these constraints. As a result, \eqref{eq:Beta_Linear} can result in negative range coefficients ($\hat{\gamma}_j,\hat{\theta}<0$), which may lead to flipped interval bounds ($\hat{y}_i^->\hat{y}_i^+$) through~\eqref{Estimated_Linear}. The authors do not discuss how to maintain these constraints in practice, though they expect that if any of the estimates of $\hat{\gamma_j}$ and $\hat{\theta}$ turns out to be negative, forcing it to be positive may lead to poor fitness of the LM model\cite{sun2015linear}. 

\subsection{The Parametrized Model (2017)}
Souza et al. \cite{souza2017parametrized} proposed the Parametrized Model (PM) in 2017, which also builds two different models for the regressand bounds. This approach extracts the best reference points from the regressors and uses them to build linear regression models for both lower and upper bounds of the regressand, unlike earlier approaches that use specific reference points on the regressors, such as center, range, interval bounds. 

In this method, an interval is considered as a line segment, useful to reach all points within it. For instance, given an interval $\overline{a}$, any point $q\in\overline{a}$ can be  computed as $q=a^- (1-\lambda) + a^+\lambda$, $0\leq \lambda \leq 1$. By setting $\lambda$, $\overline{a}$ is turned into a single point. Hence, when $\lambda= 0$,  $q=a^-$ (lower bound of $\overline{a}$) and when $\lambda= 1$, $q=a^+$ (upper bound of $\overline{a}$). Similarly, $q=a^c$ (center of $\overline{a}$) when $\lambda = 0.5$. Utilizing this concept, the PM method specifies the linear regression models for the lower and upper bounds of $Y$ in~\eqref{PM_lower_upper}. 
\begin{equation}\label{PM_lower_upper}
\begin{split}
    y_i^-=\beta_0^-+\sum_{j=1}^p{{\beta_j^-(1-\lambda_j) x_{ji}^-+\beta_j^-\lambda_j x_{ji}^+}}+\epsilon_i^-,\\
    y_i^+=\beta_0^++\sum_{j=1}^p{{\beta_j^+(1-\lambda_j) x_{ji}^-+\beta_j^+\lambda_j x_{ji}^+}}+\epsilon_i^+.
\end{split}
\end{equation}

Equation~\eqref{PM_lower_upper1} simplifies \eqref{PM_lower_upper} by replacing $\beta_j^-(1-\lambda_j)$ by $\alpha_j^-$ and $\beta_j^-\lambda_j$ by $\omega_j^-$ for lower bounds, and $\beta_j^+(1-\lambda_j)$ and $\beta_j^+\lambda_j$  by $\alpha_j^+$ and $\omega_j^+$  respectively. 
\begin{equation}\label{PM_lower_upper1}
\begin{split}
    y_i^-=\beta_0^-+\sum_{j=1}^p{\alpha_j^- x_{ji}^-+\omega_j^- x_{ji}^+}+\epsilon_i^-\\
    y_i^+=\beta_0^++\sum_{j=1}^p{\alpha_j^+ x_{ji}^-+\omega_j^+ x_{ji}^+}+\epsilon_i^+.
\end{split}
\end{equation}

In matrix notation, the lower bound model can be expressed for all $n$ observations as $Y^- = X^*\beta^- + \epsilon^{-}$, where

\resizebox{.95\linewidth}{!}{
  \begin{minipage}{\linewidth}
\begin{align*}\label{PM_form}
    \begin{split}
     Y^-=(y_1^-\quad y_2^-\quad\dots\quad y_n^-)^T,\quad \beta^-=(\beta_0^-\quad \alpha_1^-\quad\omega_1^-\quad\dots\quad\alpha_p^-\quad\omega_p^-)^T,\\ \epsilon^-=(\epsilon_1^-\quad \epsilon_2^-\quad\dots\quad\epsilon_n^-)^T,
     \text{ and } X^*=\begin{pmatrix}
                     1 & x_{11}^- & x_{11}^+ &\dots & x_{p1}^+\\
                     1 & x_{12}^- & x_{12}^+ &\dots & x_{p2}^+\\
                     \vdots & \vdots &\vdots &\dots &\vdots\\
                     1 & x_{1n}^- &x_{1n}^+ &\dots & x_{pn}^+\\
                     \end{pmatrix}.
\end{split}
\end{align*}
\end{minipage}}

The LS estimate of the coefficients for the lower bound model, $\beta^-$ is computed by~\eqref{eq:Beta_PM_lower}.
\begin{equation}\label{eq:Beta_PM_lower}
    \hat{\beta}^-=((X^*)^T X^*)^{-1}(X^*)^TY^-
\end{equation}

The matrix expression follows the same pattern for the upper bound model, $Y^+= X^*\beta^+ + \epsilon^{+}$, and the LS estimate of the coefficients for the upper bound model, $\beta^+$ in defined in~\eqref{eq:Beta_PM_upper}.
\begin{equation}\label{eq:Beta_PM_upper}
    \hat{\beta}^+=((X^*)^T X^*)^{-1}(X^*)^TY^+
\end{equation}
Finally, using $\hat{\beta}^-$ and $\hat{\beta}^+$, the lower and upper bounds of $Y$ are estimated using \eqref{PM_lower_upper2}.
\begin{equation}\label{PM_lower_upper2}
\begin{split}
    \hat{y}_i^-=\hat{\beta}_0^-+\sum_{j=1}^p{\hat{\alpha}_j^- x_{ji}^-+\hat{\omega}_j^- x_{ji}^+}\\
    \hat{y}_i^+=\hat{\beta}_0^++\sum_{j=1}^p{\hat{\alpha}_j^+ x_{ji}^-+\hat{\omega}_j^+ x_{ji}^+}.
\end{split}
\end{equation}

The PM method does not automatically guarantee the mathematical coherence of the bounds~\cite{souza2017parametrized}. To avoid flipping the interval bounds, the approach estimates the range of $Y$ using \eqref{eq:PM_range_estimate} before performing the regression.
\begin{equation}\label{eq:PM_range_estimate}
    \hat{Y}^w=X^*((X^*)^T X^*)^{-1}(X^*)^TY^w
\end{equation}
If all estimated ranges are positive ($\hat{y}^w\in\hat{Y}^w$), the model automatically ensures mathematical coherence. However, if at least one of the estimated ranges is negative, it applies the Box-Cox transformation~\cite{box1964analysis}, extended to intervals by the authors \cite{souza2017parametrized} (defined in~\eqref{eq:Extended_Box_cox}), to transform the regressand so that the desirable coherence is achieved by the PM method. 
\begin{equation}\label{eq:Extended_Box_cox}
  \overline{y}_i^k=\begin{cases}
    \left[\frac{(y_i^-+k_2)^{k_1}-1}{k_1},\frac{(y_i^++k_2)^{k_1}-1}{k_1}\right], & \text{if $k_1\neq 0$}.\\
    \left[log(y_i^-+k_2),log(y_i^++k_2)\right], & \text{if $k_1=0$}.
  \end{cases}
\end{equation}
where $\overline{y}_i^k$ is the transformed interval of $\overline{y}_i$. $k_1$ is any real value and $k_2$ maintains the following restriction: $y_i^-+k_2>0$.

\begin{table*}
\centering
\caption{Summary of well-known linear regression models for IV data}
\resizebox{0.97\textwidth}{!}{
\begin{tabular}[t]{llcl} 
\hline
 \small{Methods} &  \small{Interval reference point(s) used} &  \small{Also applicable to numeric data} &\small{Ensure mathematical coherence on } \\
  & {\small{in the regression process}} & {\small{(interval range = 0)}}   & {\small{interval bounds}}\\\hline
CM \cite{billard2000regression}&Center&Yes&No\\
MinMax \cite{billard2002symbolic}&Lower and upper bounds&Yes&No\\
CRM \cite{neto2008centre}&Center and radius (half of range)     &No&No\\
CCRM \cite{neto2010constrained}& Center and radius (half of range)&No&Yes with positivity restrictions on range coefficients\\
CIM \cite{wang2012linear}& All points within intervals &Yes&Yes with Moore’s  linear  combination algorithm \cite{moore1966interval}\\
LM \cite{sun2015linear}  & Lower and upper bounds, and range &No&Yes with positivity restrictions as proposed in this paper\\
PM \cite{souza2017parametrized}& Any point within intervals &Yes&Yes with the interval Box-Cox transformation \cite{box1964analysis}\\\hline
 \end{tabular}}\vspace{-1.4em}
\label{tab:regression_models}
\end{table*}

\section{Data and Methodology}
\label{sec:methodology}
In Section~\ref{sec:background}, we discussed linear regression models for intervals considering their vector/matrix representation. Table~\ref{tab:regression_models} provides a summary of these models and their features. In this section, we initially put forward an extension to the LM method \cite{sun2015linear}  which ensures mathematical coherence, thus making it more suitable for real-world data. Next, we introduce a novel approach to intuitively visualizing IV regression---the Interval Regression Graph (\emph{IRG}). \emph{IRG}s are designed to facilitate the interpretation and communication of the regression model and thus the relationship between an independent and dependent variable of interest. The latter is a crucial asset of traditional (numeric) regression models, accounting for a substantial part of their utility and popularity. Going beyond the traditional visualization of numeric regression, \emph{IRG}s communicate the relationships of both position and range (uncertainty) of IV regressor and regressand. Following the introduction of \emph{IRG}s, we introduce both synthetic and real-world data sets to support the empirical exploration and analysis of the regression approaches in Section~\ref{sec:background}. Finally, we discuss evaluation metrics employed for performance assessment during the remainder of the paper.

\subsection[Extension of the Linear Model]{Extension of the Linear Model~\cite{sun2015linear}}
As pointed out in Section~\ref{sec:background}, the LM method~\cite{sun2015linear} considers positivity restrictions on the range coefficients to consistently estimate the regressand bounds. However, it neither guarantees these restrictions through its model setting, nor provides algorithmic solutions to maintaining them. Thus, to make the model suitable for practical real-world deployment, we put forward two alternative extensions to the LM method: (1) universally forcing positive restrictions on parameters as~\cite{sun2015linear} suggested---that is proposing a constrained LM method, and (2) enforcing restrictions only when needed to avoid unnecessary estimation bias~\cite{sun2015linear}---that is, proposing a weakly constrained LM method.

\begin{algorithm} 
\caption{A constrained LM method ($LM_c$)}
\noindent\textbf{Input:} Let $Y=\{\overline{y}_1,\overline{y}_2,...,\overline{y}_n\}$ be the regressand variable's samples and $\{X_1, X_2,..., X_p\}$ are the $p$ ($\geq 1$) be the regressor variables. \vspace{0.2cm}\\
\noindent\textbf{Output:} Let $\hat{Y}=\{\hat{\overline{y}}_1,\hat{\overline{y}}_2,...,\hat{\overline{y}}_n\}$ be the estimated $Y$.
\label{algorithm_1}
\begin{algorithmic}[1]
 \State Apply the LM method at \eqref{eq:Beta_Linear} on $Y$ and $X_j$ ($1\leq j\leq p$) to estimate $\hat{\beta}^*$ and $\hat{\beta}^w$ 
 \If {$\hat{\theta}<0$ or $\hat{\gamma}_j<0$ where $\hat{\theta},\hat{\gamma}_j\in \hat{\beta}^w$,}  
 \State Apply LHA \cite{lawson1974} to make $\hat{\theta}$ and $\hat{\gamma}_j$ positive \EndIf
 \State Estimate $\hat{Y}^-$ and $\hat{Y}^+$ using \eqref{Estimated_Linear}
 \State \Return $\hat{Y}=[\hat{Y}^-,\hat{Y}^+]$
\end{algorithmic}%
\end{algorithm}

\emph{(1)} We propose a constrained LM method ($LM_c$) following the notion of the CCRM method \cite{neto2010constrained} where the Lawson and Hanson algorithm (LHA) \cite{lawson1974} is applied to universally impose 
the positivity restriction on the range coefficients, $\gamma_j$ and $\theta$. Algorithm~\ref{algorithm_1} presents this constrained LM variant, $LM_c$. Note that universally (or blindly) applying the restrictions for maintaining mathematical coherence risks poor performance in terms of model fit~\cite{sun2015linear}. 

\emph{(2)} To minimize the impact on performance, we also develop a more weakly constrained LM method (LM$_w$), in line with the rationale of the PM approach~\cite{souza2017parametrized} to avoid unnecessary bias in the estimation results due to universally restricting parameters. Here, we first regress the range of the regressand variable ($y_i^w$) on the range of regressors ($x_{ji}^w$) before performing the estimation using~\eqref{eq:PM_range_estimate} as the estimated range values of the regressand guarantee mathematical coherence of the estimated bounds \emph{if} they are positive~\cite{souza2017parametrized}. Therefore, if all estimated range values are positive ($\hat{y}_i^w\geq 0, \forall i$), we compute the least squares estimation of \eqref{eq:Beta_Linear} without any parameter restrictions. In contrast, if any of the estimated range values are negative ($\hat{y}_i^w<0, \exists i$), we enforce positivity restrictions on $\hat{\gamma}_j$ and $\hat{\theta}$ using the Lawson and Hanson algorithm (LHA)~\cite{lawson1974,neto2010constrained}. \emph{Algorithm~\ref{algorithm_2} summarizes this adapted approach, LM$_w$}. 

\begin{algorithm}
\caption{A weakly-constrained LM method ($LM_w$)}
\noindent\textbf{Input:} Let $Y=\{\overline{y}_1,\overline{y}_2,...,\overline{y}_n\}$ be the regressand variable's samples and $\{X_1, X_2,..., X_p\}$ are the $p$ ($\geq 1$) regressor variables. \\\vspace{0.2cm}
\noindent\textbf{Output:} Let $\hat{Y}=\{\hat{\overline{y}}_1,\hat{\overline{y}}_2,...,\hat{\overline{y}}_n\}$ be the estimated $Y$.
\label{algorithm_2}
\begin{algorithmic}[1]
\State Estimate the range values of regressand variable, $\hat{Y}^w=\{\hat{y}^w_1,...,\hat{y}^w_n\}$ using \eqref{eq:PM_range_estimate}
\If {all $\hat{y}^w_i\geq 0$}
   \State Apply the LM method at \eqref{eq:Beta_Linear} on $Y$ and $X_j$ ($1\leq j\leq p$) to generate $\hat{\beta}^*$ and $\hat{\beta}^w$ 
   \State Estimate $\hat{Y}^-$ and $\hat{Y}^+$ using \eqref{Estimated_Linear}
\Else
   \State Apply \textbf{Algorithm 1} to estimate $\hat{Y}^-$ and $\hat{Y}^+$
\EndIf 
\State \Return $\hat{Y}=[\hat{Y}^-,\hat{Y}^+]$
\end{algorithmic}
\end{algorithm}

\subsection{The Interval Regression Graph (IRG)}
\label{sec:methodology.IRG}
Visualization of traditional regression models provides a powerful mechanism for interpretability and communication of the relationship between numeric variables. While intervals provide a richer model, enabling the capture of uncertain information, the crucial interpretation and communication of any insights is complex, yet just as vital as for the numeric case. We introduce a novel 3D visualization approach, the interval regression graph (\emph{IRG}) which succinctly and clearly captures the change in a regressand’s key features (center and range) for given changes in a regressor’s key features (center and range). The \emph{IRG} visualizes the correlation between a regressor's and regressand’s position (or range), based on a given regression model \cite{kabir2022visualization}. Note that in this paper, for conciseness, we only introduce and focus on \emph{IRG}s generated for visualizing the relationship between one regressor and one regressand. Algorithm~\ref{algorithm_3} presents the pseudocode for constructing \emph{IRG}s for a regressand's center and range with respect to a regressor's center and range---for a given regression model. 

\begin{algorithm} [t!]
\caption{Interval Regression Graph (\emph{IRG}) Generation 
}
\noindent\textbf{Input:} An IV regression model (IVRM). The IV regressor's (e.g., from the original data set) minimum range  and maximum range are $\text{range}X_{\text{min}}$ and $\text{range}X_{\text{max}}$, as well as its minimum center and maximum center coordinates are $\text{center}X_{\text{min}}$ and $\text{center}X_{\text{max}}$. \vspace{0.02cm}\\
\noindent\textbf{Output:} Two \emph{IRG} plots mapping the regressor to the regressand's center, $\hat{Y}^c$ and range, $\hat{Y}^w$.\vspace{0.2cm}\\
\label{algorithm_3}
\begin{minipage}{0.48\textwidth}
\renewcommand\footnoterule{} 
\begin{algorithmic}[1] 
\State Generate the set $\boldsymbol{X^w}$ of $p$ discretizations of the interval $[\text{range}X_{\text{min}}, \text{range}X_{\text{max}}]$\footnote{Note that the \emph{IRG} generation algorithm is designed to allow for the visualization of both linear and non-linear relationships between variables. In this paper, we focus on linear relationships/regression models only.}
\State Generate the set $\boldsymbol{X^c}$ of $q$ discretizations of the interval $[\text{center}X_{\text{min}}, \text{center}X_{\text{max}}]$
\For {each discretized $X^w_i, 1\leq i\leq p$}
 \For {each discretized $X^c_j, 1\leq j\leq q$}
 \State Compute $X^{-}_{ij}=X^c_j-\frac{X^w_i}{2}$
 \State Compute $X^{+}_{ij}=X^c_j+\frac{X^w_i}{2}$ 
 \State Compute $\hat{Y}^{-}_{ij}$, $\hat{Y}^{+}_{ij}$ with $X^{-}_{ij}$, $X^{+}_{ij}$ using IVRM
 \State Compute $\hat{Y}^c_{ij} =\frac{\hat{Y}^{-}_{ij}+\hat{Y}^{+}_{ij}}{2}$
 \State Compute $\hat{Y}^w_{ij}=\hat{Y}^{+}_{ij}-\hat{Y}^{-}_{ij}$ 
 \EndFor
 \EndFor
\State Generate $\hat{Y}^c$ \emph{IRG} plot with $\boldsymbol{\hat{Y}^c}$ on the vertical axis, $\boldsymbol{X^w}$ on the bottom left axis and $\boldsymbol{X^c}$ on the bottom right axis 
\State Generate $\hat{Y}^w$ \emph{IRG} plot with $\boldsymbol{\hat{Y}^w}$ on the vertical axis, $\boldsymbol{X^w}$ on the bottom left axis and $\boldsymbol{X^c}$ on the bottom right axis
 \State \Return \emph{IRG} plots for $\hat{Y}^c$ and $\hat{Y}^w$
\end{algorithmic}
\end{minipage}
\end{algorithm}

\begin{figure}[t!]
\begin{subfigure}{0.24\textwidth}
\includegraphics[width=0.985\textwidth]{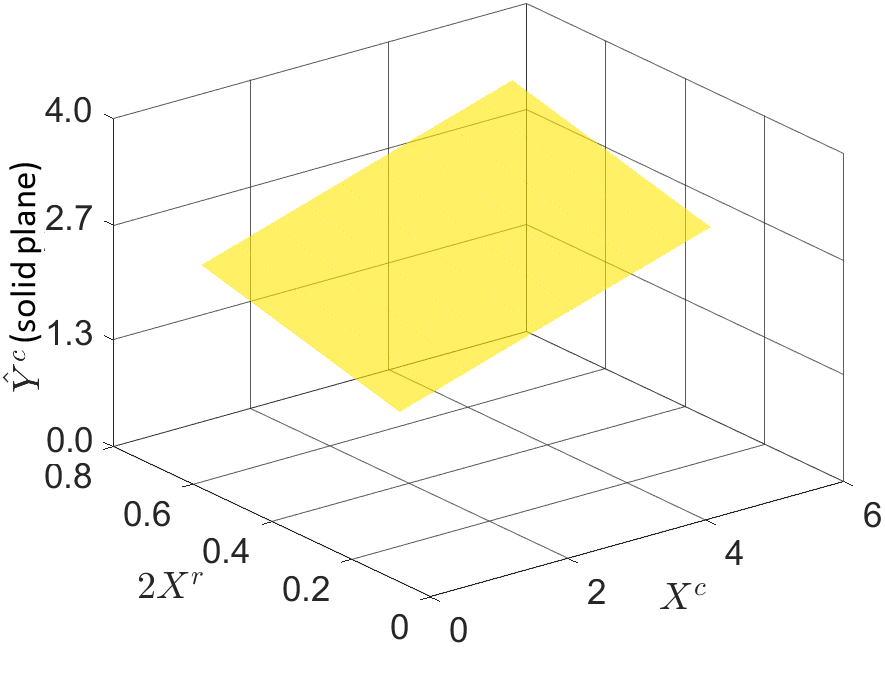}
\caption{\emph{IRG} as to $\hat{Y}^c$}
\end{subfigure}
\begin{subfigure}{0.24\textwidth}
\includegraphics[width=0.985\textwidth]{2.png}
\caption{\emph{IRG} as to $\hat{Y}^w$ ($\simeq 2\hat{Y}^r$)}
\end{subfigure}
\caption{Relationship between regressand and regressor with respect to center~(position) (a) and range (b) using the PM method for Set-1 (Fig.~\ref{fig1}(a)).}\vspace{-1.4em}
\label{fig:synthetic2-case11}
\end{figure}

To illustrate the \emph{IRG} and its utility, consider the IV Set-1 (Fig.~\ref{fig1}(a)), presented in the introductory section. Figures~\ref{fig:synthetic2-case11}(a) and (b) individually show the two different aspects of the \emph{IRG} for Set-1 based on the PM regression method.  The bottom-left and bottom-right axes show the regressor’s center ($X^c$) and range ($X^w\simeq 2\hat{X}^r$) respectively. In Fig.~\ref{fig:synthetic2-case11}(a), the vertical axis denotes the regressand’s estimated center ($\hat{Y}^c$ \text{(solid plane)}), while in Fig.~\ref{fig:synthetic2-case11}(b), it reflects the regressand’s estimated range ($\hat{Y}^w\simeq 2\hat{Y}^r \text{(dotted plane)}$). 

Interpreting these figures, we can see how Fig.~\ref{fig:synthetic2-case11}(a) visualizes that the regressand’s center, $\hat{Y}^c$ increases with respect to increasing values of both the regressor's range, $2X^r$ (or $X^w$) and its center, $X^c$. Fig.~\ref{fig:synthetic2-case11}(b) shows how the regressand’s range, $2\hat{Y}^r$ also increases with respect to both increasing values of the regressor's range, $2X^r$ and center, $X^c$, and that it does so at a greater rate in each case than does $\hat{Y}^c$. For compact and complete visualization, \emph{IRGs} commonly combine the visualizations with respect to both the regressand's center and range as shown in Fig.~\ref{fig:case-1}(f). We will leverage \emph{IRGs} throughout the remainder with additional examples. 

\begin{figure}[t!]
\begin{subfigure}{0.23\textwidth}
\centering
\includegraphics[width=1\textwidth]{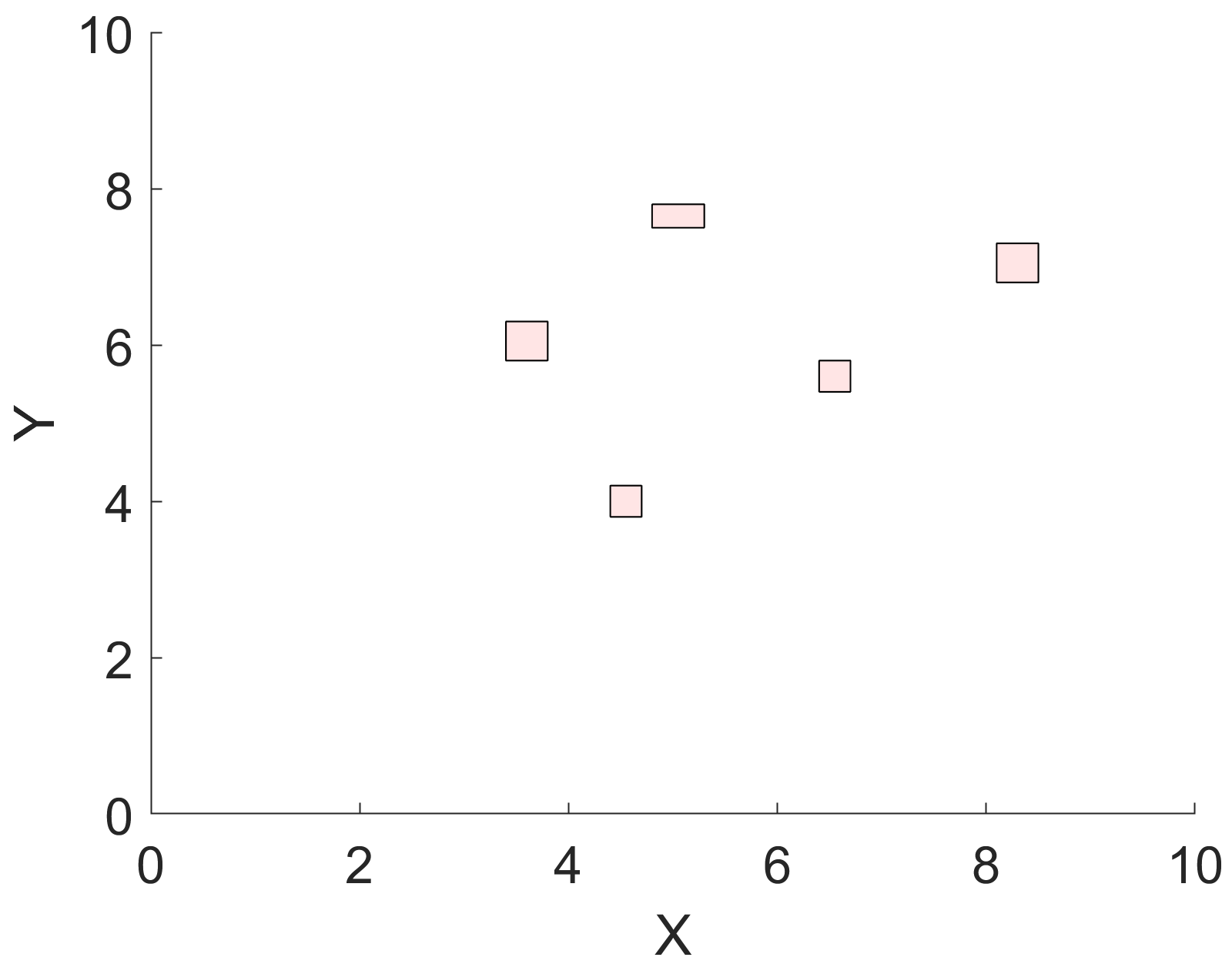}
\caption{\begin{footnotesize}Set-3 (All disjoint SIs)
\end{footnotesize}}
\end{subfigure}\quad
\begin{subfigure}{0.23\textwidth}
\centering
\includegraphics[width=1\textwidth]{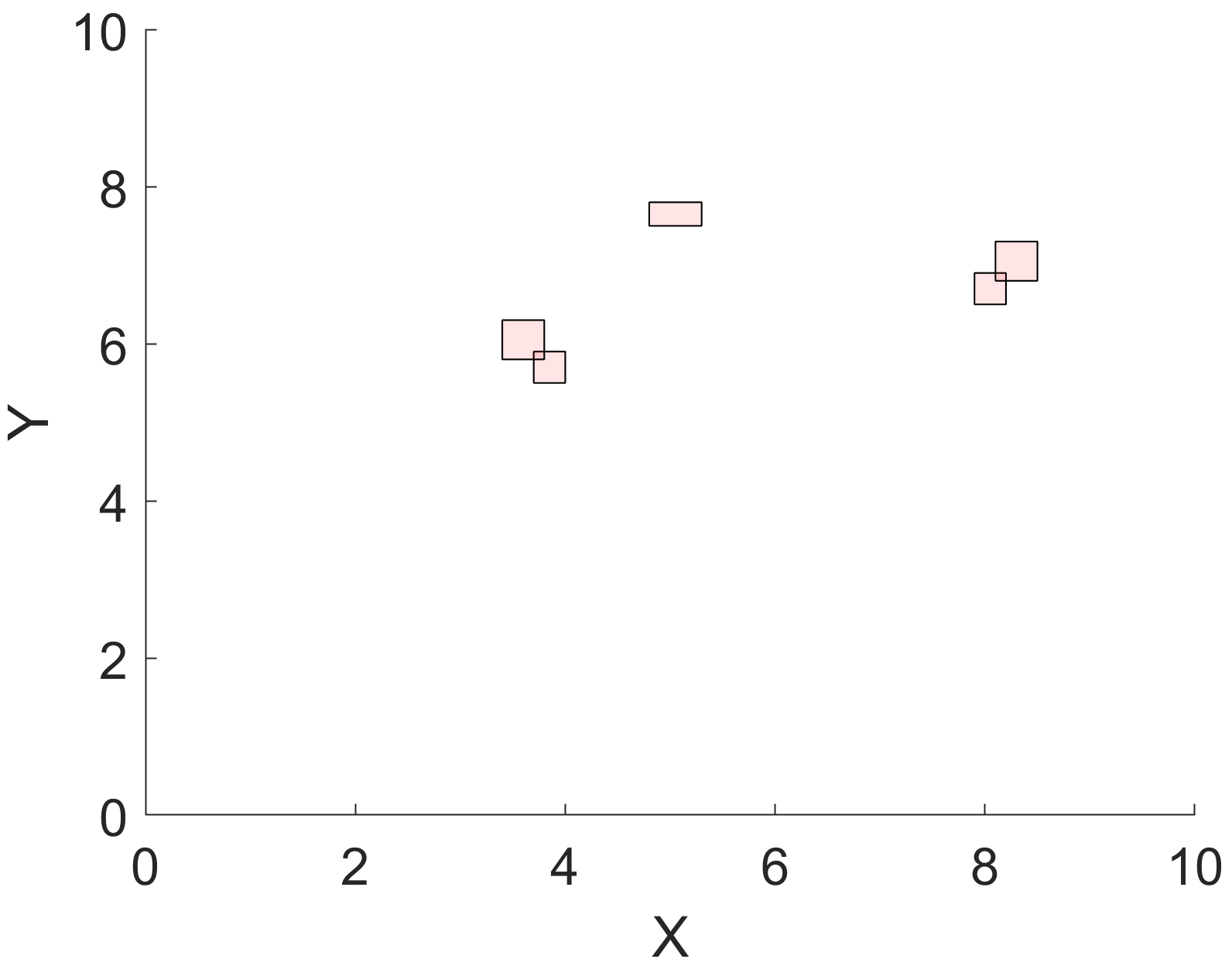}
\caption{\begin{footnotesize}Set-4 (Some overlapping SIs)\end{footnotesize}}
\end{subfigure}\\
\begin{subfigure}{0.23\textwidth}
\centering
\includegraphics[width=1\textwidth]{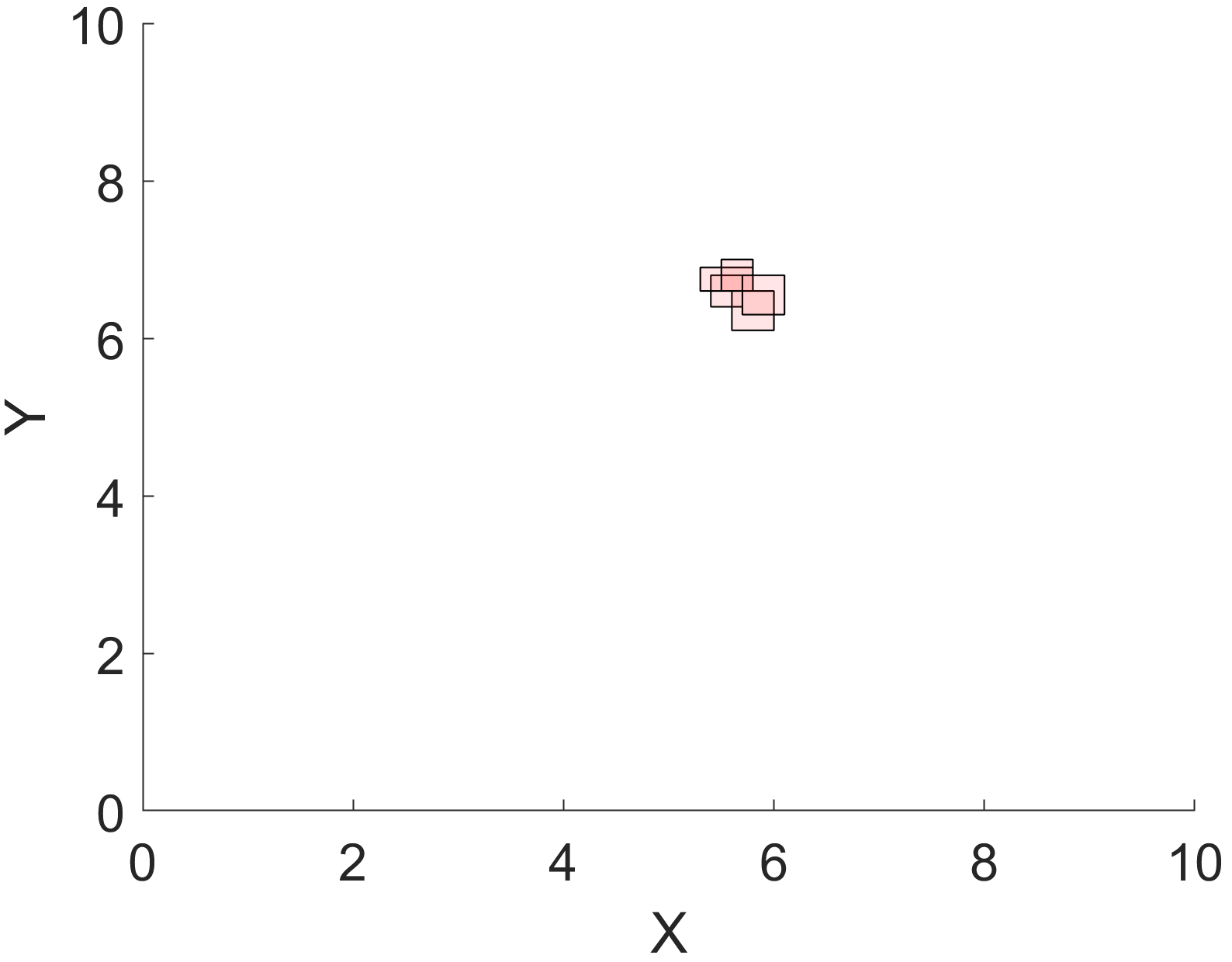}
\caption{\begin{footnotesize}Set-5 (All overlapping SIs)\end{footnotesize}}
\end{subfigure}\quad
\begin{subfigure}{0.23\textwidth}
\centering
\includegraphics[width=1\textwidth]{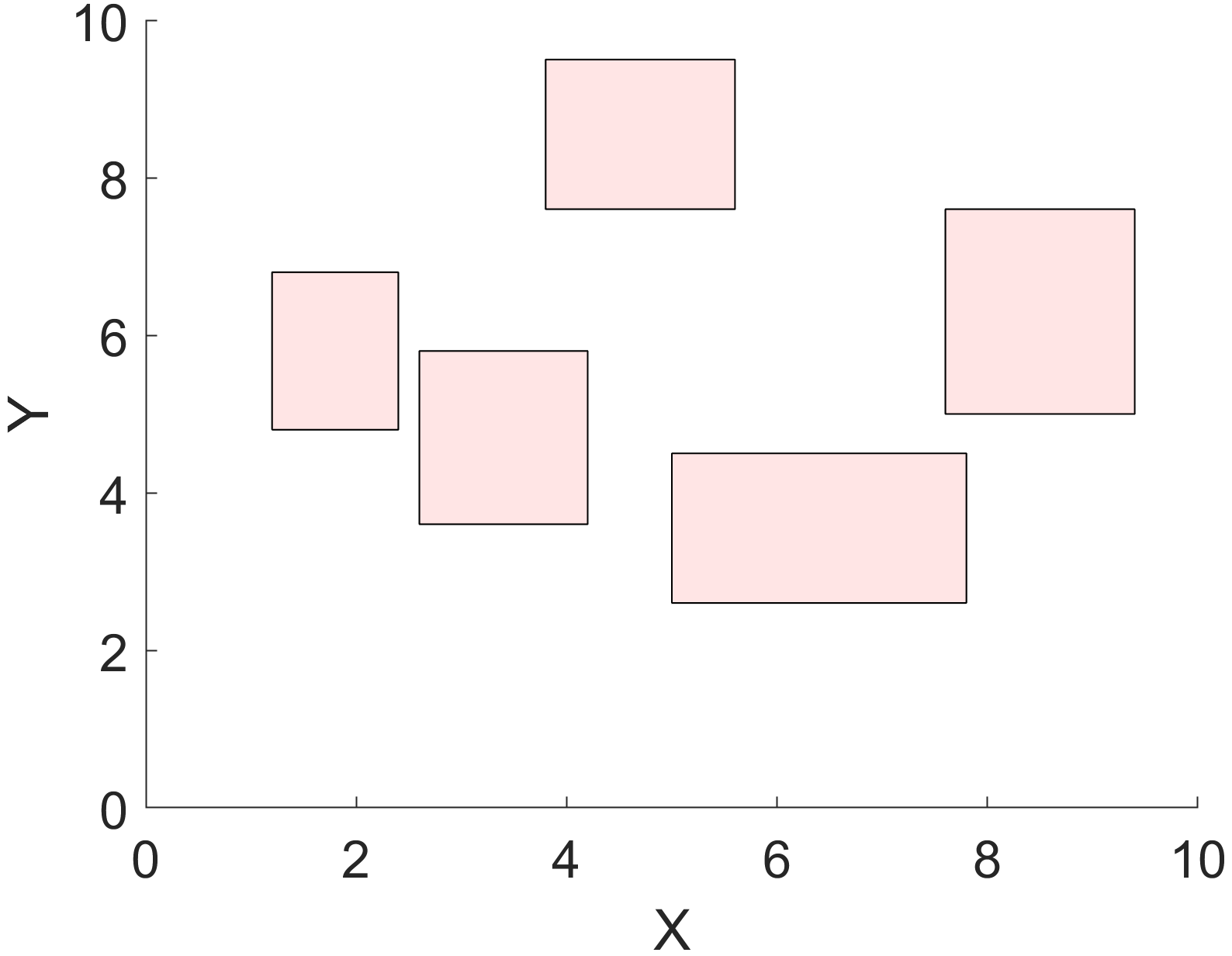}
\caption{\begin{footnotesize}Set-6 (All disjoint PIs)\end{footnotesize}}
\end{subfigure}\\
\begin{subfigure}{0.23\textwidth}
\centering
\includegraphics[width=1\textwidth]{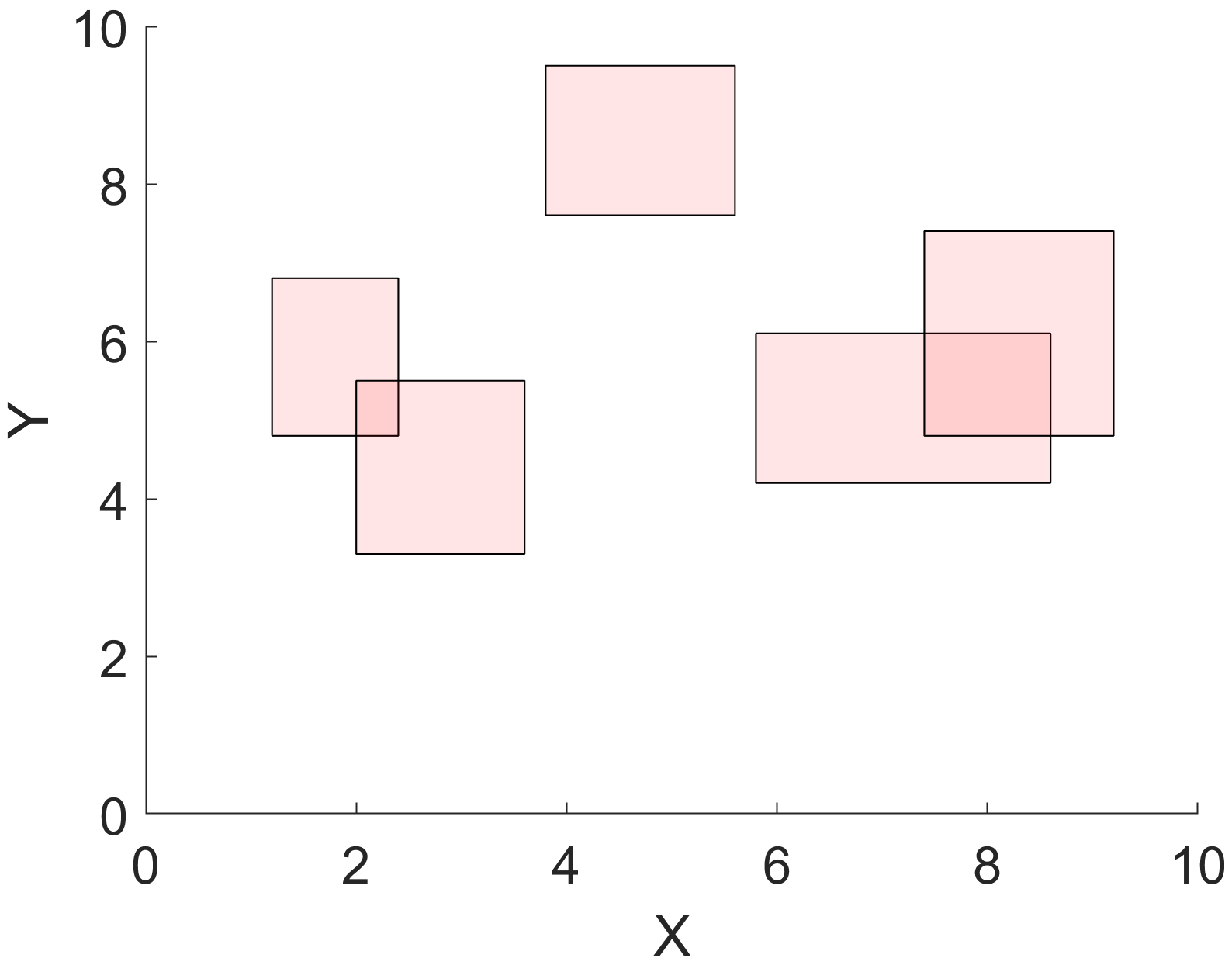}
\caption{\begin{footnotesize}Set-7 (Some overlapping PIs)\end{footnotesize}}
\end{subfigure}\quad
\begin{subfigure}{0.23\textwidth}
\centering
\includegraphics[width=1\textwidth]{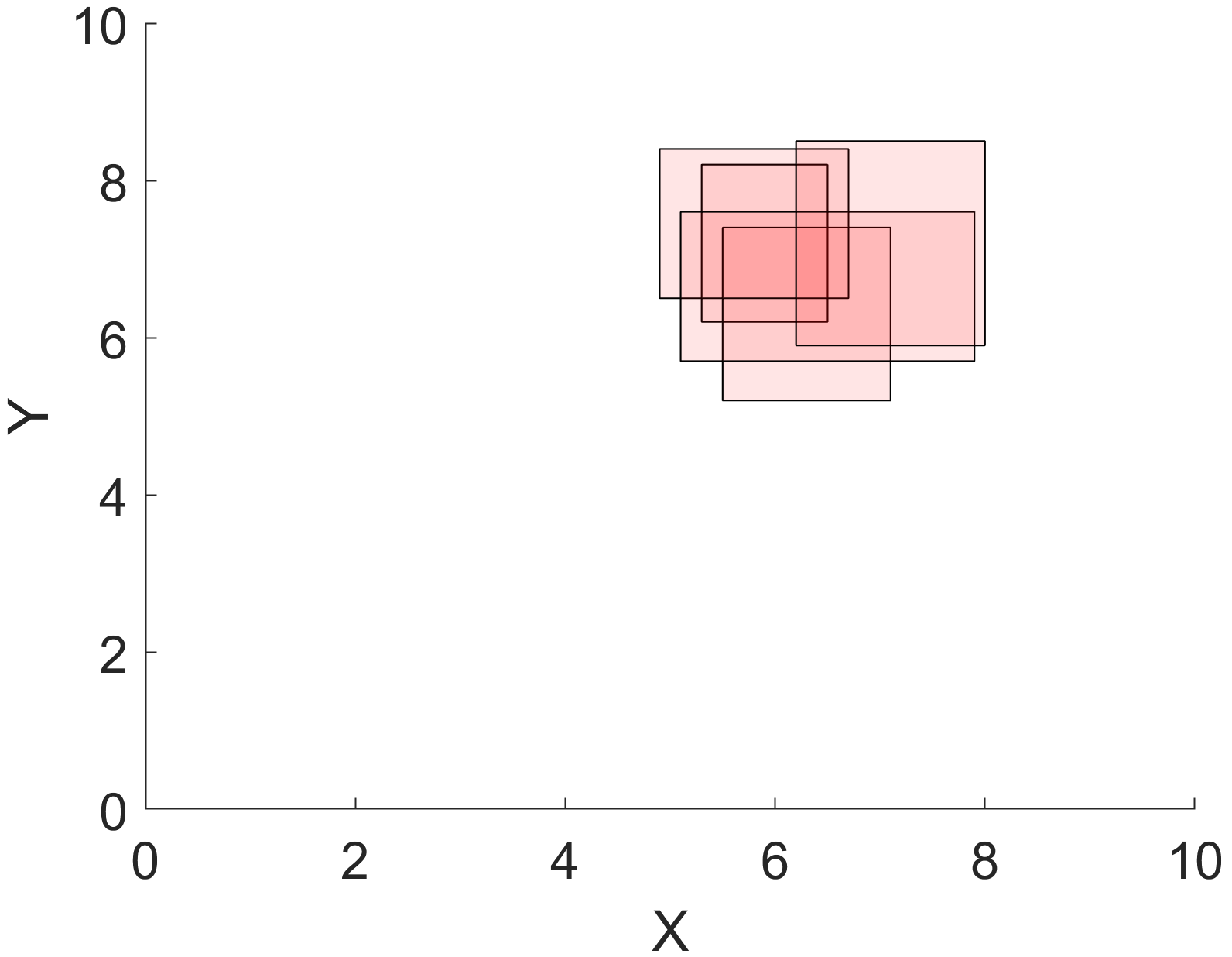}
\caption{\begin{footnotesize}Set-8 (All overlapping PIs)\end{footnotesize}}
\end{subfigure}\\
\begin{subfigure}{0.23\textwidth}
\centering
\includegraphics[width=1\textwidth]{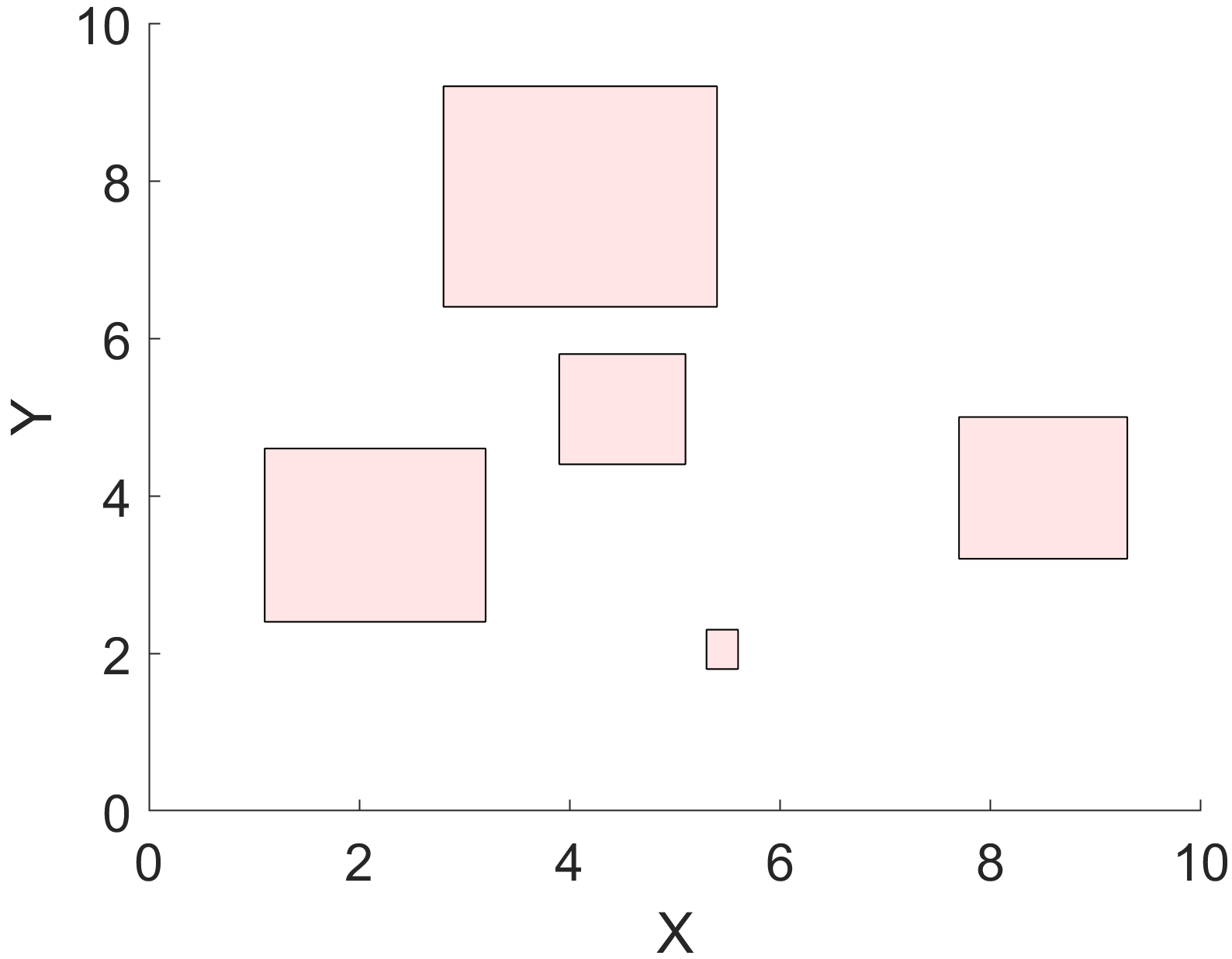}
\caption{\begin{footnotesize}Set-9 (Disjoint MIs)\end{footnotesize}}
\end{subfigure}\quad
\begin{subfigure}{0.23\textwidth}
\centering
\includegraphics[width=1\textwidth]{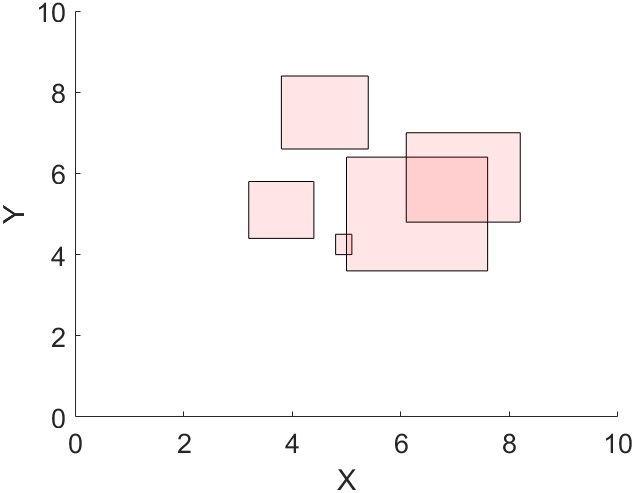}
\caption{\begin{footnotesize}Set-10 (Some overlapping MIs)\end{footnotesize}}
\end{subfigure}\\
\begin{subfigure}{0.46\textwidth}
\centering
\includegraphics[width=0.5\textwidth]{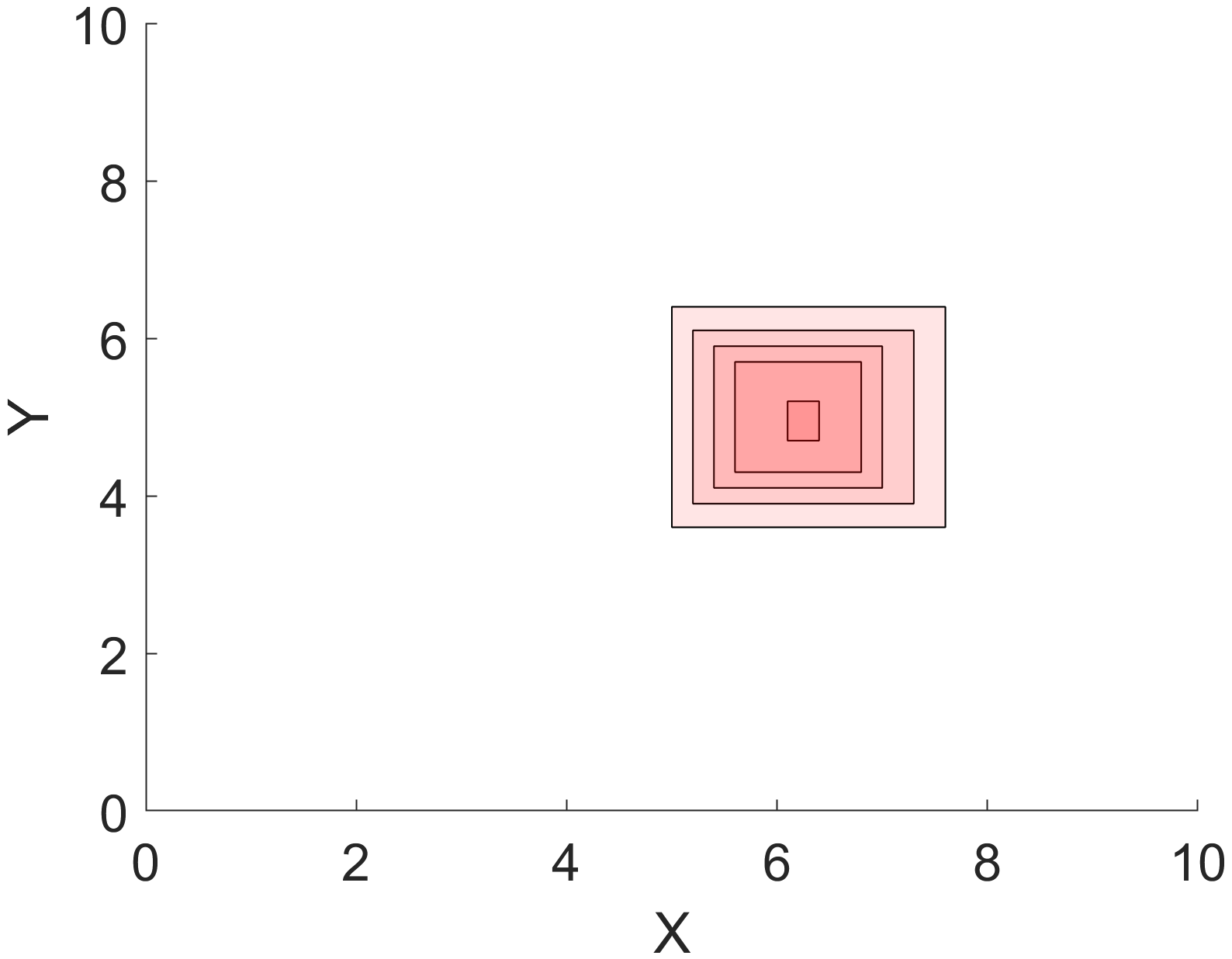}
\caption{\begin{footnotesize}Set-11 (Nested/all overlapping MIs)\end{footnotesize}}
\end{subfigure}
\caption{2D visualization of synthetic data sets, Set-3 to Set-11. SI=skinny interval, PI=puffy interval, and MI=mixed interval.} \vspace{-1.5em}
\label{fig:synthetic_data_set_A}
\end{figure}

\subsection{Data Sets}
\label{sec:methodology.data sets}
We use both synthetic and real-world IV data sets to investigate the performance of linear regression methods. The synthetic data sets are used to explore and visualize the sensitivity of the regression models with respect to different properties of the interval data, whereas the real-world data sets are used to investigate their suitability to real-world scenarios. All data sets are described below.   

\subsubsection{Synthetic Data Sets}
\label{sec:methodology.syn.data sets}
We specifically design a series of synthetic IV data sets to investigate the behavior of the regression models. First, Set-1 and Set-2, shown at the beginning of the paper (Figs.~\ref{fig1}(a) and~\ref{fig11}(a)), are used to explore the relationship between regressor and regressand with respect to the primary features (i.e., center and range). Second, an additional nine sets of intervals (Set-3 to Set-11), exhibiting different properties are used: we use three sets of \emph{skinny} intervals (Set-3 to Set-5) with smaller range\footnote{The `skinny' and `puffy' designs have been inspired by~Ferson et al.~\cite{ferson2007experimental}.} and three \emph{puffy} sets with wider range (Set-6 to Set-8) having features of disjointedness, some, and full overlap. We also consider three mixed cases (Set-9 to Set-11), containing puffy and skinny intervals in separated, partially overlapping and nested states. All of these data sets are produced within the range of 0 to 10. They are given in Table~\ref{tab:data set2}~(a)~to~(k) in the Appendix including the mean ($\mu$) and standard deviation ($\sigma$) of the center and range of intervals. Figure~\ref{fig:synthetic_data_set_A} graphically presents the data sets, Set-3 to Set-11.

\subsubsection{Real-world Data Sets}
\label{sec:methodology.real.data sets}
We focus on three real-world IV data sets with different properties and arising in different application contexts, i.e., medical, consumer food ratings/marketing, and cyber-security, reflecting the broad applicability of IV data in real-world applications. 

The \emph{first} data set represents medical observations and includes interval-valued \emph{systolic} and \emph{diastolic} blood pressure (BP) data of 59 patients of the Hospital Valle del Nalón in Asturias, Spain~\cite{blanco2011estimation,garcia2020multiple}.

The \emph{second} data set presents IV consumer ratings of eight different (UK market) snack-food products~\cite{ellerby2020insights}. In this case, 40 consumers rated each product---using `DECSYS' interval survey software \cite{ellerby2019decsys}---on six different attributes: \emph{visual appeal}, \emph{value for money}, \emph{healthiness}, \emph{taste}, \emph{branding}, and \emph{ethics}, along with their \emph{overall purchase intention} (i.e., how likely they were to buy each product). 

The \emph{third} and final data set captures assessments by cyber-security experts of the vulnerability of system components in a real-world scenario. Specifically, 38 cyber-security experts at the UK CESG (Communications-Electronics Security Group) assessed a range of components (referred to as hops) that are commonly encountered during a cyber-attack~\cite{miller2016modelling,ellerby2019exploring}. They rated the hops on overall difficulty for an attacker to either attack or evade them, as well as rating each on several attributes (seven for attack, and three for evade) that might affect this difficulty. The experts provided their IV ratings on a scale from 0 to 100. For more details of these data sets, see~\cite{blanco2011estimation,garcia2020multiple,ellerby2020insights,miller2016modelling,ellerby2019exploring}.

\subsection{Evaluation Metrics}
\label{sec:methodology.metrics}
Evaluation poses one of the key challenges for IV data analysis and AI more broadly. While `quality of fit' for numeric data is straightforward, the same is not the case for intervals. Comparing the latter requires application-led assumptions (on the nature of the intervals and whether for example interval size is more important than position), which in turn feed into similarity measures or arithmetic means of comparison. We do not delve further into this challenge in this paper, instead following three fairly standard metrics to evaluate the performance of the linear regression models. Each of the evaluation metrics is defined with respect to the regressand (samples), $Y=\{\overline{y}_1,\overline{y}_2,...,\overline{y}_n\}$ and its estimated value(s),  $\hat{Y}=\{\hat{\overline{y}}_1,\hat{\overline{y}}_2,...,\hat{\overline{y}}_n\}$.

\noindent\emph{1) Root Mean Squared Error (RMSE)} estimates error by considering the average deviation of the estimated values from the actual ones~\cite{neto2010constrained}. The \emph{RMSE} for the lower and upper bounds of $Y$ is defined as,  
\begin{equation*}\label{eq:RMSE}
\resizebox{0.99\columnwidth}{!}{$
    RMSE^-=\sqrt{\frac{1}{n}\sum\limits_{i=1}^{n}(y_i^--\hat{y}_i^-)^2} \text{ and }
    RMSE^+=\sqrt{\frac{1}{n}\sum\limits_{i=1}^{n}(y_i^+-\hat{y}_i^+)^2}, 
$}
\end{equation*}
where 
a smaller \emph{RMSE} value indicates a better fitted model.

\noindent\emph{2) Mean Absolute Error (MAE)} calculates the average of absolute difference between the actual and estimated values~\cite{iccen2016error}. For the upper and lower bounds of $Y$, the \emph{MAE} is defined as,  
\begin{equation*}\label{eq:MAE}
\resizebox{0.99\columnwidth}{!}{ $
MAE^-=\frac{1}{n}\sum\limits_{i=1}^n\left|y_i^--\hat{y}_i^-\right| \text{ and } MAE^+=\frac{1}{n}\sum\limits_{i=1}^n\left|y_i^+-\hat{y}_i^+\right|,
$}
\end{equation*}
where $0\leq MAE^-,MAE^+\leq\mathbb{R}^+$. Again, a lower value of \emph{MAE} implies a better fitted model. 

\noindent\emph{3) Mean Magnitude of Relative Error (MMRE)} measures the mean of dispersion of the actual and estimated values with respect to the actual values~\cite{fagundes2016quantile}. It is calculated as follows:
\begin{equation*} \label{eq:MMRE}
    MMRE=\frac{1}{n}\sum\limits_{i=1}^n\frac{1}{2}\left\{\left|\frac{y^-_i-\hat{y}^-_i}{y^-_i}\right|+\left|\frac{y^+_i-\hat{y}^+_i}{y^+_i}\right|\right\},
\end{equation*}
where $0\leq MMRE\leq \mathbb{R}^+$. In this case, the lower the \emph{MMRE}, the better the model.

\section{Results}
\label{sec:results}
We explore the performance of existing regression approaches together with two proposed variants of the LM method (LM$_c$ and LM$_w$) for synthetic and real-world IV data sets (see Section~\ref{sec:methodology.data sets}). Their performance is assessed using different metrics  described in Section~\ref{sec:methodology.metrics}. All  these regression methods are implemented in Matlab and the full source code will be available at http://lucidresearch.org/software with the publication of the paper.

\subsection{Performance Analysis with Synthetic Data Sets}
In Section \ref{sec:methodology.IRG}, we briefly introduced the \emph{IRG}s using IV Set-1 (Fig.~1(a)). We expand on this example here, reporting performance of all regression models covered above with this set, as well as for Set-2 (Fig.~2(a)). 

\begin{figure}[b!]
\vspace{-0.9em}
\begin{subfigure}{0.24\textwidth}
\includegraphics[width=0.985\textwidth]{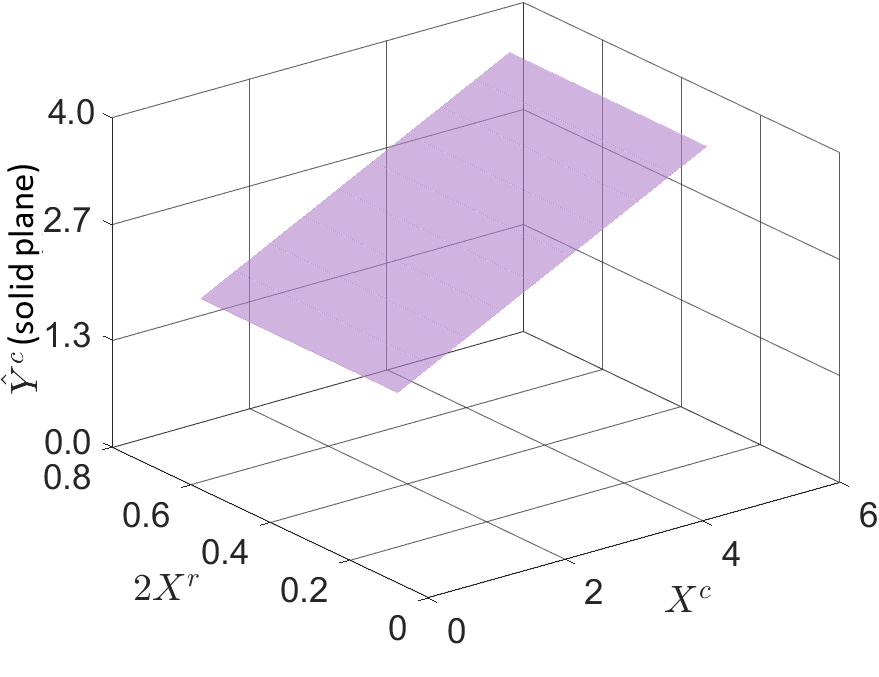}
\caption{\emph{IRG} as to $\hat{Y}^c$}
\end{subfigure}
\begin{subfigure}{0.24\textwidth}
\includegraphics[width=0.985\textwidth]{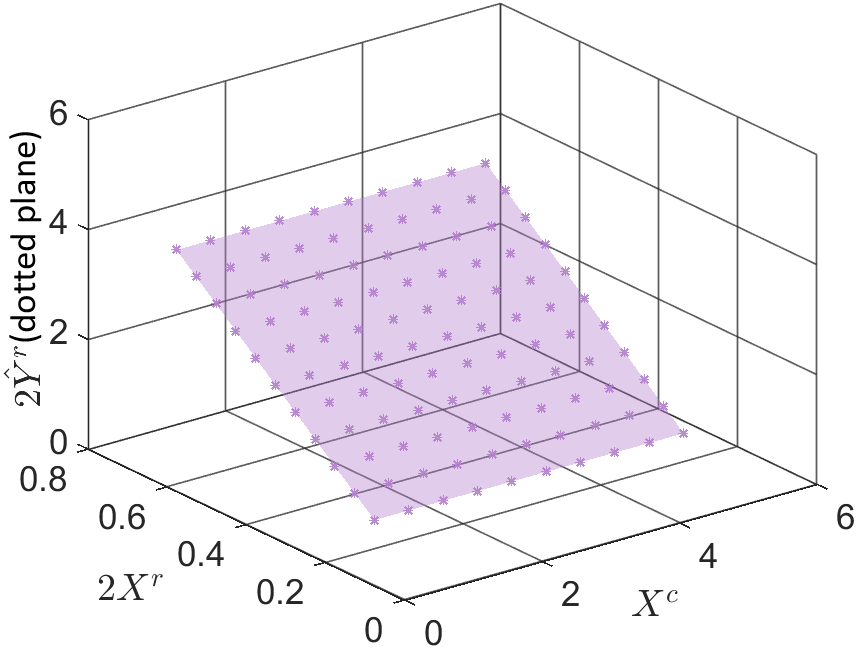}
\caption{\emph{IRG} as to $\hat{Y}^w$ ($\simeq 2\hat{Y}^r$)}
\end{subfigure}
\caption{Relationship between regressand and regressor with respect to center~(position) (a) and range (b) using the CRM method for Set-1 (Fig.~\ref{fig1}(a)).}\vspace{-1.2em}
\label{fig:synthetic2-case1}
\end{figure}
\begin{table}[b!]
    \centering
\caption{Regression performance for Set-1}
\resizebox{0.9\hsize}{!}{
\begin{tabular}{cccccc} \hline
Methods & $RMSE^-$ & $RMSE^+$ & $MAE^-$ & $MAE^+$ & $MMRE$ 
\\\hline
CM &	1.506&	1.510&	1.354&	1.354&	0.816\\
MinMax&	0.002&	0.113&	0.001&	0.109&	0.017\\
CRM	&0.160&	0.055&	0.140&	0.047&	0.075\\
CCRM&	0.251&	0.227&	0.218&	0.198&	0.140\\
CIM	&1.503&	1.517&	1.356&	1.356&	0.817\\
PM	&\textbf{0.0}&	\textbf{0.040}&	\textbf{0.0}&	\textbf{0.031}&	\textbf{0.006}\\
LM$_c$	&0.097&	0.428&	0.085&	0.381&	0.104\\
LM$_w$	&0.097&	0.105&	0.085&	0.088&	0.053\\\hline
\begin{scriptsize}\textbf{Bold} = Best\end{scriptsize}&&&&&
\end{tabular}}
\label{tab:my_label1}
\end{table}

As discussed earlier, Set-1 shows the more commonly considered and intuitive scenario where uncertainty and position in regressor and regressand vary in unison. On the other hand, Set-2 presents a less commonly considered scenario, where the uncertainty of the regressand increases irrespective of the uncertainty of the regressor. Said differently, for Set-2, uncertainty and position of the regressand vary quasi exclusively with respect to the \emph{position} of the regressor variable.

\begin{figure}
\label{fig:synthetic2-case2}
\begin{subfigure}{0.24\textwidth}
\includegraphics[width=0.99\textwidth]{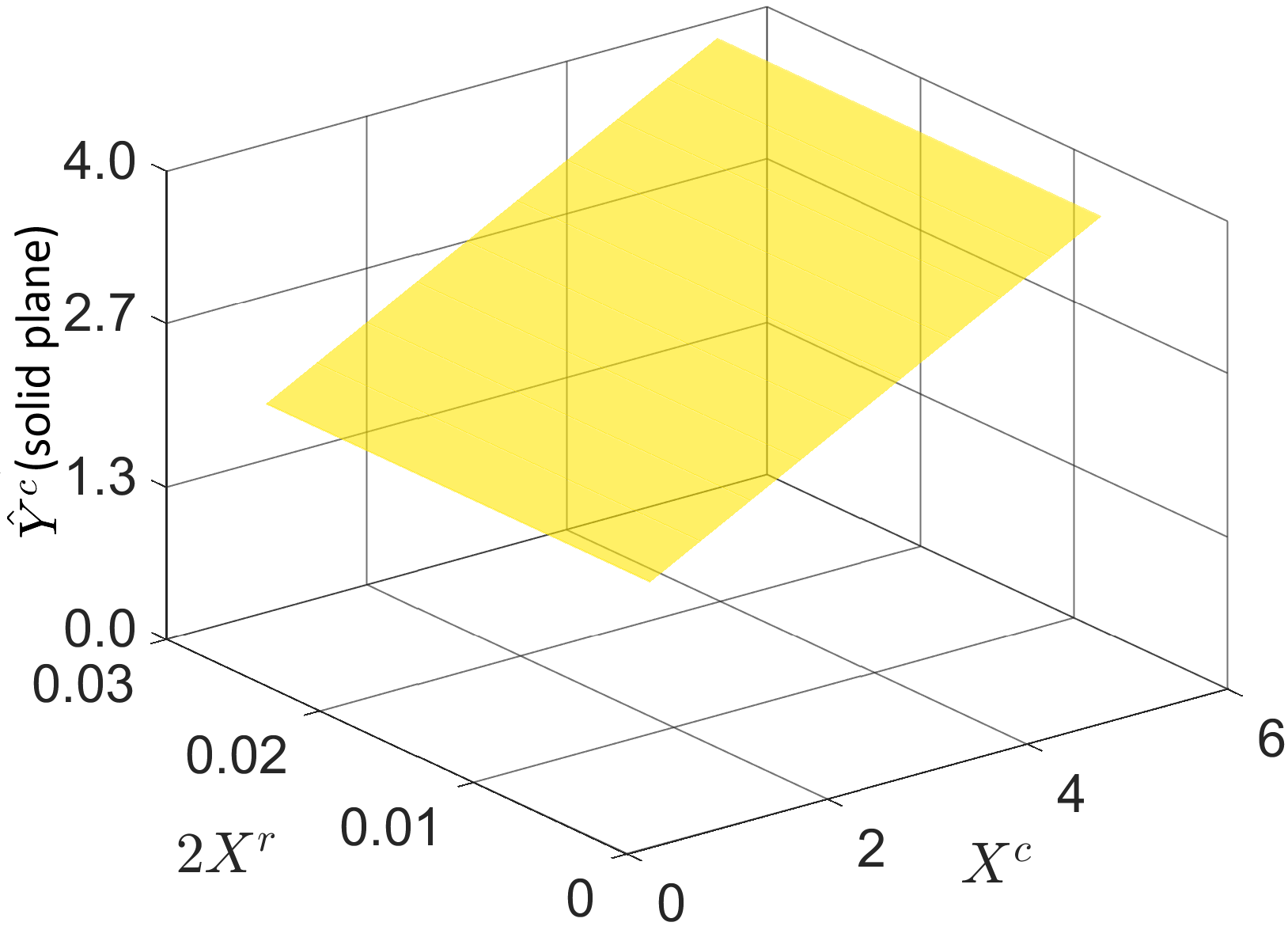}
\caption{\emph{IRG} as to $\hat{Y}^c$}
\end{subfigure}
\begin{subfigure}{0.24\textwidth}
\includegraphics[width=0.99\textwidth]{4.png}
\caption{\emph{IRG} as to $\hat{Y}^w$ ($\simeq 2\hat{Y}^r$)}
\end{subfigure}
\caption{Relationship between regressand and regressor with respect to center~(position) (a) and range (b) using the PM method for Set-2 (Fig.~\ref{fig11}(a)).}\vspace{-0.8em}
\label{fig:synthetic2-case22}
\end{figure}
\begin{figure}
\begin{subfigure}{0.24\textwidth}
\includegraphics[width=0.99\textwidth]{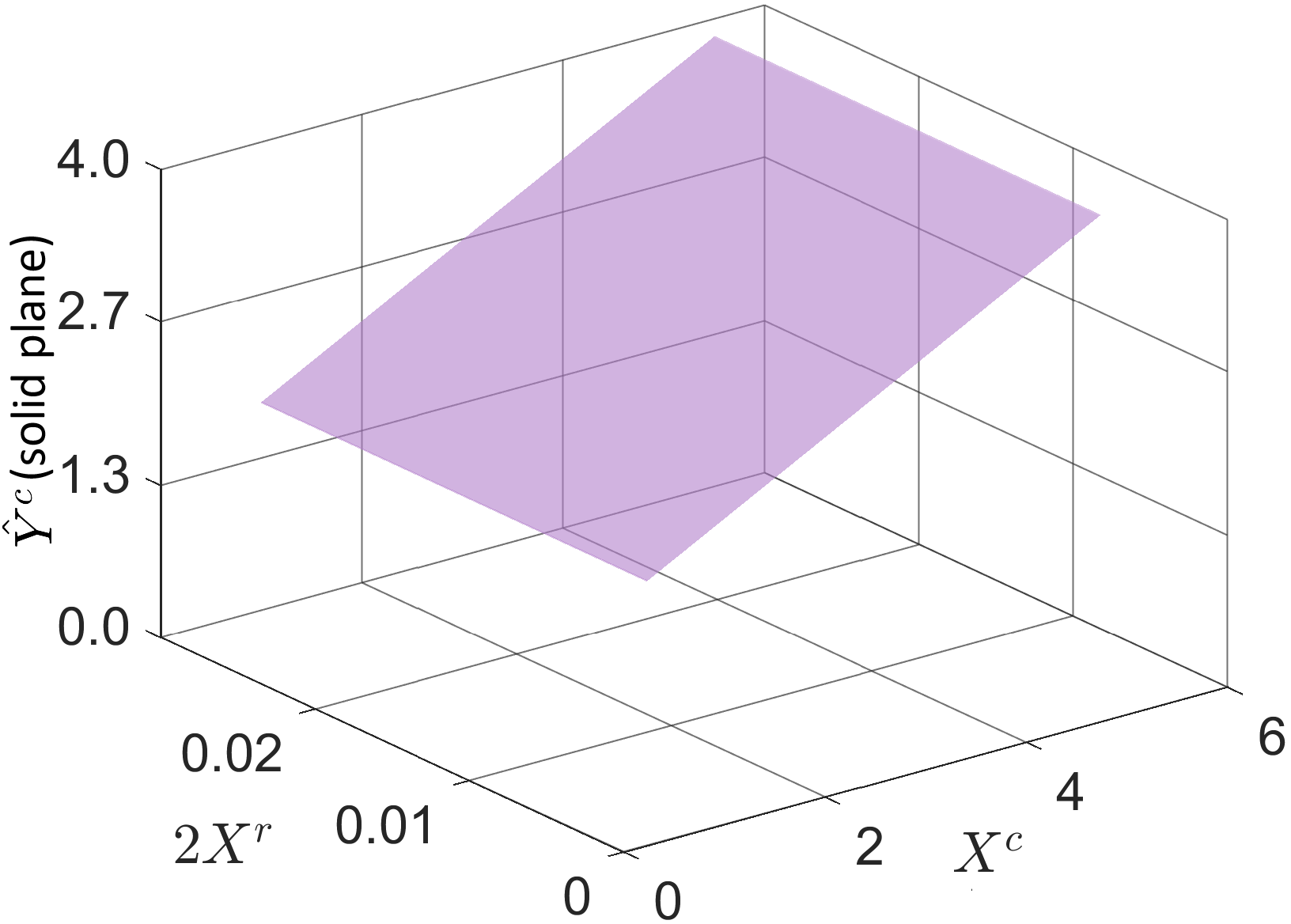}
\caption{\emph{IRG} as to $\hat{Y}^c$}
\end{subfigure}
\begin{subfigure}{0.24\textwidth}
\includegraphics[width=0.99\textwidth]{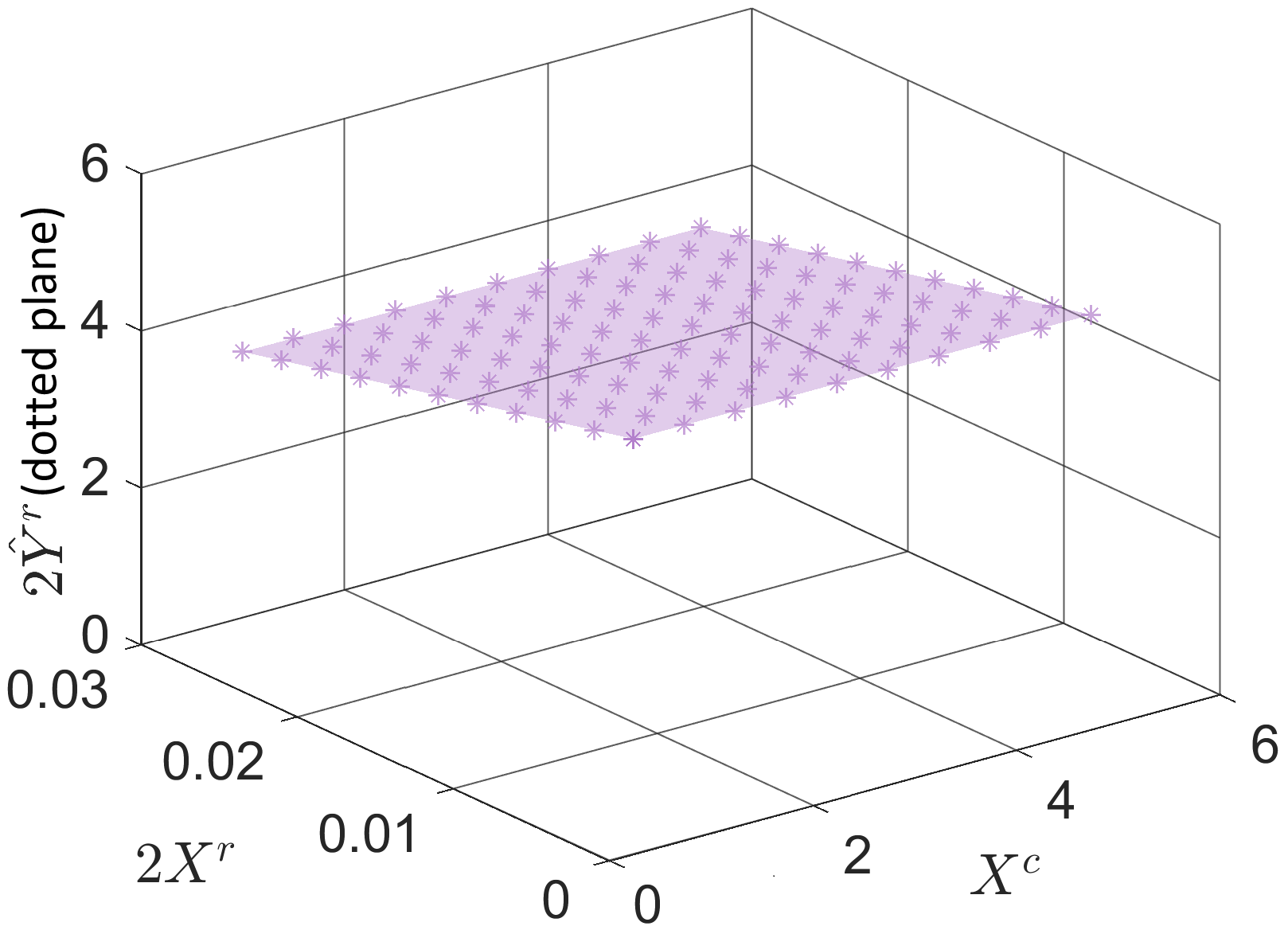}
\caption{\emph{IRG} as to $\hat{Y}^w$ ($\simeq 2\hat{Y}^r$)}
\end{subfigure}
\caption{Relationship between regressand and regressor with respect to their center~(position) (a) and range (b) using the CRM method for Set-2 (Fig.~\ref{fig11}(a)).}\vspace{-0.8em}
\label{fig:synthetic2-case21}
\end{figure}

\begin{table}[t!]
\centering
\caption{Regression performance for Set-2}
\resizebox{0.9\hsize}{!}{
\begin{tabular}{cccccc} \hline
Methods & $RMSE^-$ & $RMSE^+$ & $MAE^-$ & $MAE^+$ & $MMRE$ \\\hline
CM	&1.650	&1.657	&1.491	&1.491	&0.90\\
MinMax	&\textbf{0.002}	&\textbf{0.153}	&\textbf{0.001}	&0.148	&\textbf{0.023}	\\
CRM	&0.575	&0.592	&0.53	&0.535	&0.344	\\
CCRM	&0.703	&0.719	&0.61	&0.625	&0.398	\\
CIM	&1.65	&1.657	&1.491	&1.491	&0.90\\
PM	&\textbf{0.002}	&\textbf{0.153}	&\textbf{0.001}	&\textbf{0.145}	&\textbf{0.023}	\\
LM$_c$	&0.573	&1.007	&0.528	&0.930	&0.398\\
LM$_w$	&0.573	&0.593	&0.528	&0.537	&0.343\\\hline
\begin{scriptsize}\textbf{Bold} = Best\end{scriptsize}&&&&&
\end{tabular}}\vspace{-1.6em}
\label{tab:my_label2}
\end{table}
\begin{table*}
\caption{Performance of different regression models for IV synthetic data sets (Set-3 to Set-11)}
\begin{subtable}{.33\linewidth}
\centering
\caption{Set-3 (All disjoint skinny intervals)}
\renewcommand{\arraystretch}{1.1}
\resizebox{0.95\columnwidth}{!}{
\begin{Huge}
\begin{tabular}{ccccccc} \hline
Methods & $RMSE^-$ & $RMSE^+$ & $MAE^-$ & $MAE^+$ & $MMRE$ \\\hline
CM&1.215	&1.181	&0.972	&1.035	&0.185\\
MinMax&1.204	&1.169	&1.004	&1.002	&0.185\\
CRM&1.185	&1.189	&0.986	&1.021	&0.185\\
CCRM&1.195	&1.179	&0.995	&1.011	&0.185\\
CIM&1.215	&1.181	&0.972	&1.035	&0.185\\
PM&\textbf{0.343}	&\textbf{0.355}	&\textbf{0.291}	&0.312	&\textbf{0.059}\\
LM$_c$&\textbf{0.343}	&0.356	&0.293	&0.317	&\textbf{0.059}\\
LM$_w$&\textbf{0.343}	&\textbf{0.355}	&0.293	&\textbf{0.310}	&\textbf{0.059}\\\hline\vspace{0.1mm}
\end{tabular}\end{Huge}}
\end{subtable}%
\begin{subtable}{.33\linewidth}
\centering
\caption{Set-4  (Some overlapping  skinny  intervals)}
\renewcommand{\arraystretch}{1.1}
\resizebox{0.95\columnwidth}{!}{
\begin{Huge}
\begin{tabular}{ccccccc} \hline
Methods & $RMSE^-$ & $RMSE^+$ & $MAE^-$ & $MAE^+$ & $MMRE$ \\\hline
CM&0.656	&0.601	&0.592	&0.407	& 0.074	\\
MinMax&0.631&	0.574&	0.483&	0.432&	0.068	\\
CRM&0.616&	0.591&	0.475&	0.451&	0.069\\
CCRM&0.626&	0.581&	0.482&	0.443&	0.069\\
CIM&0.656&	0.601	&0.592&	0.407&	0.074\\
PM&\textbf{0.263}&\textbf{0.192}&	\textbf{0.249}&	\textbf{0.181}&	\textbf{0.032}\\
LM$_c$&\textbf{0.263}&	0.193&\textbf{0.249}&	0.185&0.033\\
LM$_w$&\textbf{0.263}&	\textbf{0.192}&	\textbf{0.249}&	\textbf{0.181}&	\textbf{0.032}\\\hline\vspace{0.1mm}
\end{tabular}\end{Huge}}
\end{subtable}%
\begin{subtable}{.33\linewidth}
\centering
\caption{Set-5 (All overlapping skinny intervals)}
\renewcommand{\arraystretch}{1.1}
\resizebox{0.95\columnwidth}{!}{
\begin{Huge}
\begin{tabular}{ccccccc} \hline
Methods & $RMSE^-$ & $RMSE^+$ & $MAE^-$ & $MAE^+$ & $MMRE$ \\\hline
CM&	0.370&	0.362&	0.341&	0.341&	0.052\\
MinMax&	0.141&	\textbf{0.115}&\textbf{0.120}&	\textbf{0.107}&	\textbf{0.017}\\
CRM&	0.143&	0.120&	0.125&	0.108&	0.018\\
CCRM&	0.143&	0.121&	0.128&	0.111&	0.018\\
CIM&	0.204&	0.177&	0.159&	0.149&	0.024\\
PM&	\textbf{0.139}&	\textbf{0.115}&	0.128&	\textbf{0.107}&	0.018\\
LM$_c$&	0.143&	0.122&	0.126&	0.113&	0.018\\
LM$_w$&	0.143&	0.120&	0.126&	0.109&	0.018\\\hline\vspace{0.1mm}
\end{tabular}\end{Huge}}
\end{subtable}\\%
\begin{subtable}{.33\linewidth}
\centering
\caption{Set-6 (All disjoint puffy intervals)}
\renewcommand{\arraystretch}{1.1}
\resizebox{0.95\columnwidth}{!}{
\begin{Huge}
\begin{tabular}{ccccccc} \hline
Methods & $RMSE^-$ & $RMSE^+$ & $MAE^-$ & $MAE^+$ & $MMRE$ \\\hline
CM&	1.996&	1.999&	1.796&	1.568&	0.350\\
MinMax&	1.681&	1.685&	1.299&	1.396&	0.268\\
CRM&	1.692&	1.673&	1.314&	1.389&	0.270\\
CCRM&	1.679&	1.687&	1.297&	1.391&	0.268\\
CIM&	1.976&	1.986&	1.777&	1.564&	0.347\\
PM&	\textbf{1.456}&	\textbf{1.344}&	\textbf{1.143}&	1.061&	0.188\\
LM$_c$&	1.460&	1.350&	1.147&	1.059&	0.188\\
LM$_w$&	1.460&	1.349&	1.147&	\textbf{1.057}&	\textbf{0.187}\\\hline\vspace{0.1mm}
\end{tabular}\end{Huge}}
\end{subtable}%
\begin{subtable}{.33\linewidth}
\centering
\caption{Set-7 (Some overlapping puffy intervals)}
\renewcommand{\arraystretch}{1.1}
\resizebox{0.95\columnwidth}{!}{
\begin{Huge}
\begin{tabular}{ccccccc} \hline
Methods & $RMSE^-$ & $RMSE^+$ & $MAE^-$ & $MAE^+$ & $MMRE$ \\\hline
CM&	1.751&	1.697&	1.673&	1.220&	0.266\\
MinMax&	1.434&	1.367&	1.070&	1.034&	0.180\\
CRM&	1.442&	1.357&	1.1067&	1.036&	0.184\\
CCRM&	1.437&	1.364&	1.090&	1.038&	0.183\\
CIM&	1.753&	1.698&	1.674&	1.220&	0.267\\
PM&	\textbf{1.411}&	\textbf{1.283}&	\textbf{1.067}&	0.984&	0.173\\
LM$_c$&	1.415&	1.289&	1.071&	0.983&	0.173\\
LM$_w$&	1.415&	1.288&	1.071&	\textbf{0.981}&	0.172\\\hline\vspace{0.1mm}
\end{tabular}\end{Huge}}
\end{subtable}%
\begin{subtable}{.33\linewidth}
\centering
\caption{Set-8 (All overlapping puffy intervals)}
\renewcommand{\arraystretch}{1.1}
\resizebox{0.95\columnwidth}{!}{
\begin{Huge}
\begin{tabular}{ccccccc} \hline
Methods & $RMSE^-$ & $RMSE^+$ & $MAE^-$ & $MAE^+$ & $MMRE$ \\\hline
CM&	1.308&	1.315&	1.238&	1.238&	0.184\\
MinMax&	0.416&	0.437&	0.387&	0.398&	0.059\\
CRM&	0.416&	0.445&	0.365&	0.387&	0.056\\
CCRM&	0.410&	0.452&	0.348&	0.404&	0.056\\
CIM&	1.073&	1.084&	0.986&	0.986&	0.147\\
PM&	\textbf{0.396}&	\textbf{0.406}&	0.321&	\textbf{0.328}&	\textbf{0.049}\\
LM$_c$&	0.415&	0.429&\textbf{0.313}&	0.369&	0.051\\
LM$_w$&	0.415&	0.425&	\textbf{0.313}&	0.348&	\textbf{0.049}\\\hline\vspace{0.1mm}
\end{tabular}\end{Huge}}
\end{subtable}\\%
\begin{subtable}{.33\linewidth}
\centering
\caption{Set-9 (Disjoint mixed intervals)}
\renewcommand{\arraystretch}{1.1}
\resizebox{0.95\columnwidth}{!}{
\begin{Huge}
\begin{tabular}{ccccccc} \hline
Methods & $RMSE^-$ & $RMSE^+$ & $MAE^-$ & $MAE^+$ & $MMRE$ \\\hline
CM&	1.871&	2.449&	1.616&	1.882&	0.509\\
MinMax&	1.596&	2.236&	1.349&	1.686&	0.430\\
CRM&	1.886&	1.911&	1.497&	1.526&	0.433\\
CCRM&	1.886&	1.911&	1.497&	1.526&	0.433\\
CIM&	1.828&	2.298&	1.550&	1.796&	0.487\\
PM&	\textbf{1.210}&\textbf{1.238}&\textbf	{0.065}&	1.089&	0.251\\
LM$_c$&	\textbf{1.210}&\textbf{1.238}&	1.066&	\textbf{1.087}&	\textbf{0.250}\\
LM$_w$&	\textbf{1.210}&\textbf{1.238}&	1.066&	\textbf{1.087}&	\textbf{0.250}\\\hline
\end{tabular}\end{Huge}}
\end{subtable}%
\begin{subtable}{.33\linewidth}
\centering
\caption{Set-10 (Some overlapping mixed intervals)}
\renewcommand{\arraystretch}{1.1}
\resizebox{0.95\columnwidth}{!}{
\begin{Huge}
\begin{tabular}{ccccccc} \hline
Methods & $RMSE^-$ & $RMSE^+$ & $MAE^-$ & $MAE^+$ & $MMRE$ \\\hline
CM&	1.361&	1.572&	1.293&	1.307&	0.152\\
MinMax&	0.994&	1.235&	0.906&	0.891&	0.105\\
CRM&	1.107&	1.105&	0.938&	0.918&	\textbf{0.10}\\
CCRM&	1.107&	1.105&	0.938&	0.918&	\textbf{0.10}\\
CIM	&1.344&	1.549&	1.278&	1.281&	0.165\\
PM	&\textbf{0.988}     & \textbf{0.977}&	\textbf{0.883}&	0.869&	\textbf{0.098}\\
LM$_c$ & \textbf{0.988} & \textbf{0.977}&	0.886&	\textbf{0.866} &\textbf{0.098}\\
LM$_w$	&\textbf{0.988}	& \textbf{0.977}&   0.886&  \textbf{0.866} &\textbf{0.098}\\\hline
\end{tabular}\end{Huge}}
\end{subtable}%
\begin{subtable}{.33\linewidth}
\centering
\caption{Set-11 (Nested mixed intervals)}
\renewcommand{\arraystretch}{1.1}
\resizebox{0.95\columnwidth}{!}{
\begin{Huge}
\begin{tabular}{ccccccc} \hline
Methods & $RMSE^-$ & $RMSE^+$ & $MAE^-$ & $MAE^+$ & $MMRE$ \\\hline
CM&	1.008&	1.022&	0.926&	0.926&	0.194\\
MinMax&	0.048&	0.048&	0.046&	0.045&	0.010\\
CRM&	0.027&	0.028&	0.019&	0.026&	\textbf{0.004}\\
CCRM&	0.027&	0.028&	0.019&	0.026&	\textbf{0.004}\\
CIM&	0.945&	0.958&	0.870&	0.870&	0.183\\
PM&	\textbf{0.018}&	\textbf{0.020}&	0.014&	\textbf{0.016}&	\textbf{0.003}\\
LM$_c$&	\textbf{0.018}&	\textbf{0.020}&	\textbf{0.013}&	\textbf{0.016}&	\textbf{0.003}\\
LM$_w$&	\textbf{0.018}&	\textbf{0.020}&	\textbf{0.013}&	\textbf{0.016}&	\textbf{0.003}\\\hline
\end{tabular}\end{Huge}}
\end{subtable}\vspace{0.5mm}\\
\qquad\begin{scriptsize}\textbf{Bold} = Best\end{scriptsize}
\vspace{-1.4em}
\label{tab:synthetic_data_set_A_Table}
\end{table*}

Tables~\ref{tab:my_label1} and~\ref{tab:my_label2} present the fitness results for all regression approaches for Sets 1 and 2, respectively. These show that the PM method provides the best fitness in each case, though the MinMax method provides almost equal fitness for Set-2 (they do differ with respect to $MAE^+$). For conciseness, we focus further discussion of the regression results on two of the regression approaches: the CRM---representing an earlier and more basic approach which does not guarantee mathematical coherence, and the PM---the most recent approach, which does. \emph{IRG}s for all methods are shown in the appendix (Figs.~\ref{fig:case-1}~and~\ref{fig:case-2}). We independently present the relationship of the center and range of the test regressor on the center (subfigures (a)) and the range (subfigures (b)) of the regressand, for the CRM and PM methods in Figs.~\ref{fig:synthetic2-case11},
~\ref{fig:synthetic2-case1},~\ref{fig:synthetic2-case22}, and~\ref{fig:synthetic2-case21}. Note that, Figs.~\ref{fig:synthetic2-case11}(b) and \ref{fig:synthetic2-case22}(b) are repeated for clarity from Figs.~\ref{fig1}(b) and \ref{fig11}(b) in Section~\ref{sec:introduction}.
The \emph{IRG}s in  Figs.~\ref{fig:synthetic2-case11}(a)~and~(b) show that, according to the PM method, both changes in the estimated center and range of Set-1's regressand are associated with the changes in the center and range of the regressor---in-line with our expectation. On the other hand, the \emph{IRG}s in Figs.~\ref{fig:synthetic2-case1}(a)~and~(b) show that for the CRM method, the estimated center of Set-1's regressand is influenced only by the center of the regressor, and the estimated range of Set-1 is influenced only by the range of the regressor. Thus, the CRM method is not in line with our expectation for the Set-1 data set. The rest of the methods also do not perform according to the expectation (see Appendix, Fig.~\ref{fig:case-1}).

Further, the \emph{IRG}s in Figs.~\ref{fig:synthetic2-case22}(a)~and~(b) show how, according to the PM model, both center and range of the regressand of Set-2 vary solely with respect to the center of the regressor---in line with expectation. The \emph{IRG} in Fig.~\ref{fig:synthetic2-case21}(a) visualizes how, using the CRM method, the estimated center of Set-2's regressand is similarly positively influenced by the changes in the center, but not the range of the regressor---as expected. However, in contrast to the PM model outputs, the CRM-derived \emph{IRG} for the same data, shown in Fig.~\ref{fig:synthetic2-case21}(b), indicates no strong relationship between regressand range and either facet of the regressor. If any relationship is present at all, then this is a negative effect of the range of the regressor on that of the regressand---these results do not fit those expected when viewing the data. Fig.~\ref{fig:case-2} (in the Appendix) visualizes that the MinMax method shows similar performance to the PM method. The remaining methods perform less well. 

Next, we will briefly discuss equivalent analysis and model visualizations, using the \emph{IRG}s, for an additional nine synthetic data sets (Set-3---Set-11), each designed to reflect a key property regularly encountered in IV data sets. This will highlight how different data sets---with different features---are differently suitable for different regression methods, as later results will expand upon in more detail. For data Set-3 to Set-11, all shown in Fig.~\ref{fig:synthetic_data_set_A}, we briefly summarize individual performance results across methods below before discussing performance more broadly. Table~\ref{tab:synthetic_data_set_A_Table}(a)~--~(i) present the fitness of regression methods for each of these data sets. 

\begin{itemize}
\item \emph{Set-3 (All disjoint skinny intervals)}: 
The PM and LM$_w$ methods produce the best fit for all disjoint skinny intervals. The LM$_c$ method results in equally good fit well with respect to two metrics ($RMSE^-$ and $MMRE$).  

\item \emph{Set-4 (Some overlapping skinny intervals)}:  Again, the PM and LM$_w$ methods produce the best fit. The LM$_c$ method also gives the same fitness with respect to three metrics ($RMSE^-$, $MAE^-$, and $MMRE$). The fitness of other methods improves with some overlapping of intervals.  

\item \emph{Set-5 (All overlapping skinny intervals)}: 
The PM and MinMax methods come out as the best approach. Among others, CRM, CCRM, LM$_c$ and LM$_w$ methods produce quite similar results as to some of the evaluation metrics.  

\item \emph{Set-6 (All disjoint puffy intervals)}: 
The PM method gives the best fit, though the LM$_c$ and LM$_w$ methods also produce quite similar results for puffy intervals.  

\item \emph{Set-7 (Some overlapping puffy intervals)}: 
The PM and LM$_w$ emerge as the best methods as to different metrics, though the MinMax,CRM, CCRM and LM$_c$ methods also produce similarly good fit in terms of $MMRE$. 

\item \emph{Set-8 (All overlapping puffy intervals)}: For all overlapping puffy intervals, the PM method gives the best fit with respect to three metrics and the LM$_w$ method is for two evaluation metrics. Except CM and CIM methods, growing overlap among intervals appears to lower estimation errors by the CRM, CCRM and MinMax methods.

\begin{figure*}[t!]
\centering\vspace{-0.8em}
\begin{subfigure}{0.29\textwidth}
\centering
\includegraphics[width=0.97\textwidth]{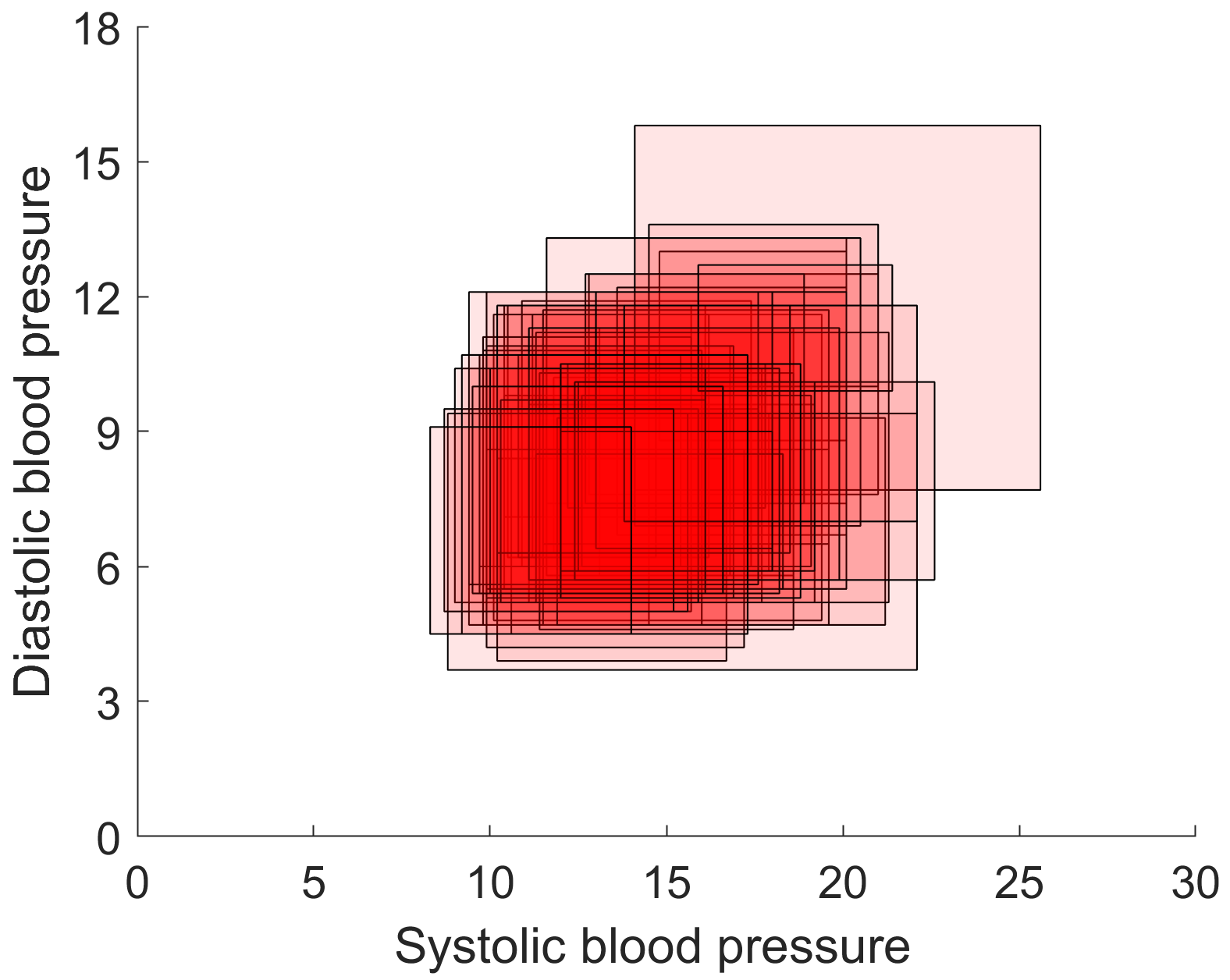}
\caption{Blood pressure data set}
\end{subfigure}\hfill
\begin{subfigure}{0.69\textwidth}
\centering\vspace{1.3em}
\renewcommand{\arraystretch}{1.2}
\resizebox{1\hsize}{!}{
\begin{Huge}
\begin{tabular}{ccccccc} \hline
Methods & $RMSE^-$ & $RMSE^+$ & $MAE^-$ & $MAE^+$ & $MMRE$ \\\hline\hline
CM	& 1.259 [0.937  1.806]&	1.617 [1.302  2.039] & 1.097 [0.772  1.658] &	1.395 [1.087  1.756] &	0.167 [0.121  0.238]\\
MinMax	&0.793 [0.631  0.905] &	1.275 [1.046  1.438] &	0.651 [0.499  0.758] &	1.058 [0.846  1.216] &	0.106 [0.086  0.120]\\
CRM	&0.881 [0.729  1.151] &	1.256 [1.006  1.377] &	0.653 [0.525  0.888] &	1.029 [0.814  1.173] &	0.105 [0.088  0.129]\\
CCRM	&0.881 [0.728  1.148]	&1.256 [1.008  1.377] &	0.653 [0.525  0.884] &	1.029 [0.815  1.174] &	0.105 [0.088  0.129]\\
CIM		&2.104	[1.869  2.386] &2.249 [1.986  2.522]	&1.951 [1.712  2.240]	&1.907 [1.681  2.199]	&0.268 [0.232  0.314]	\\	
PM	&\textbf{0.780}	[0.616  0.887] &\textbf{1.208}	[0.977  1.356]  &0.638 [0.483  0.739] &\textbf{0.999} [0.799  1.149] & \textbf{0.102} [0.083  0.115]	\\
LM$_c$	&0.784 [0.620  0.897] &1.211 [1.310  3.232]	&\textbf{0.631}	[0.480  0.736] & 1.008 [1.006  2.917]	&\textbf{0.102}	[0.10  0.194]\\
LM$_w$	&0.784	[0.620  0.897] & 1.211 [1.307  3.232]	&\textbf{0.631} [0.480  0.736] & 1.008 [1.004  2.917]	&\textbf{0.102}	[0.10  0.194]\\\hline
\end{tabular}
\end{Huge}}
\begin{scriptsize}\textbf{Bold} = Best; Bootstrapped 95\% confidence intervals at 10000 simulations are provided.\end{scriptsize}
\caption{Model Performance}
\end{subfigure}
\caption{Graphical presentation of \emph{systolic} and \emph{diastolic} BP data set~\cite{blanco2011estimation} and performance of different regression models as to evaluation metrics using this set. }\vspace{-1.4em}
\label{fig:systolic}
\end{figure*}
\begin{figure}[t!]
\centering \vspace{-0.5em}
\begin{subfigure}{0.24\textwidth}
\includegraphics[width=0.99\textwidth]{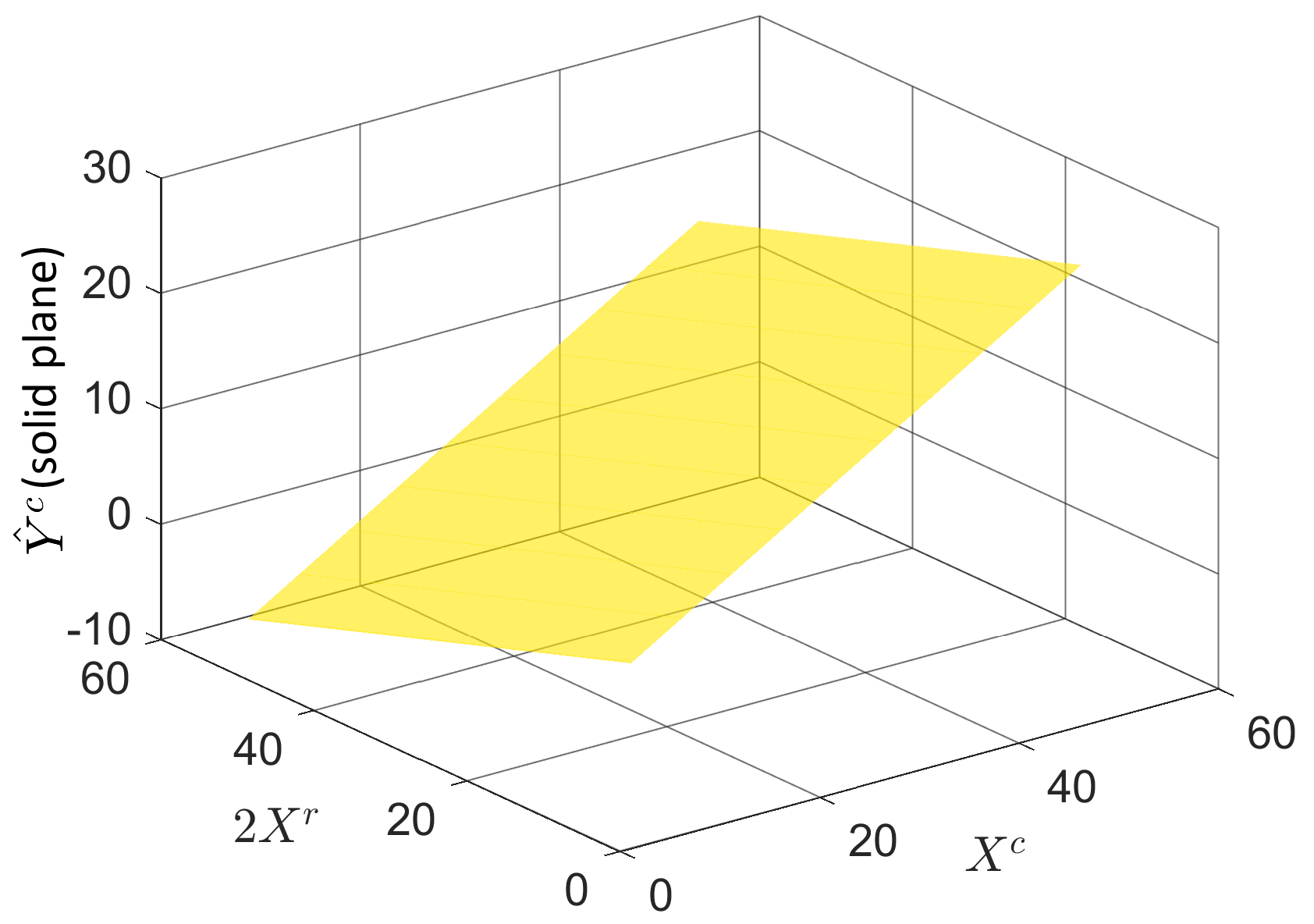}
\caption{with respect to $\hat{Y}^c$}
\end{subfigure}
\begin{subfigure}{0.24\textwidth}
\includegraphics[width=0.99\textwidth]{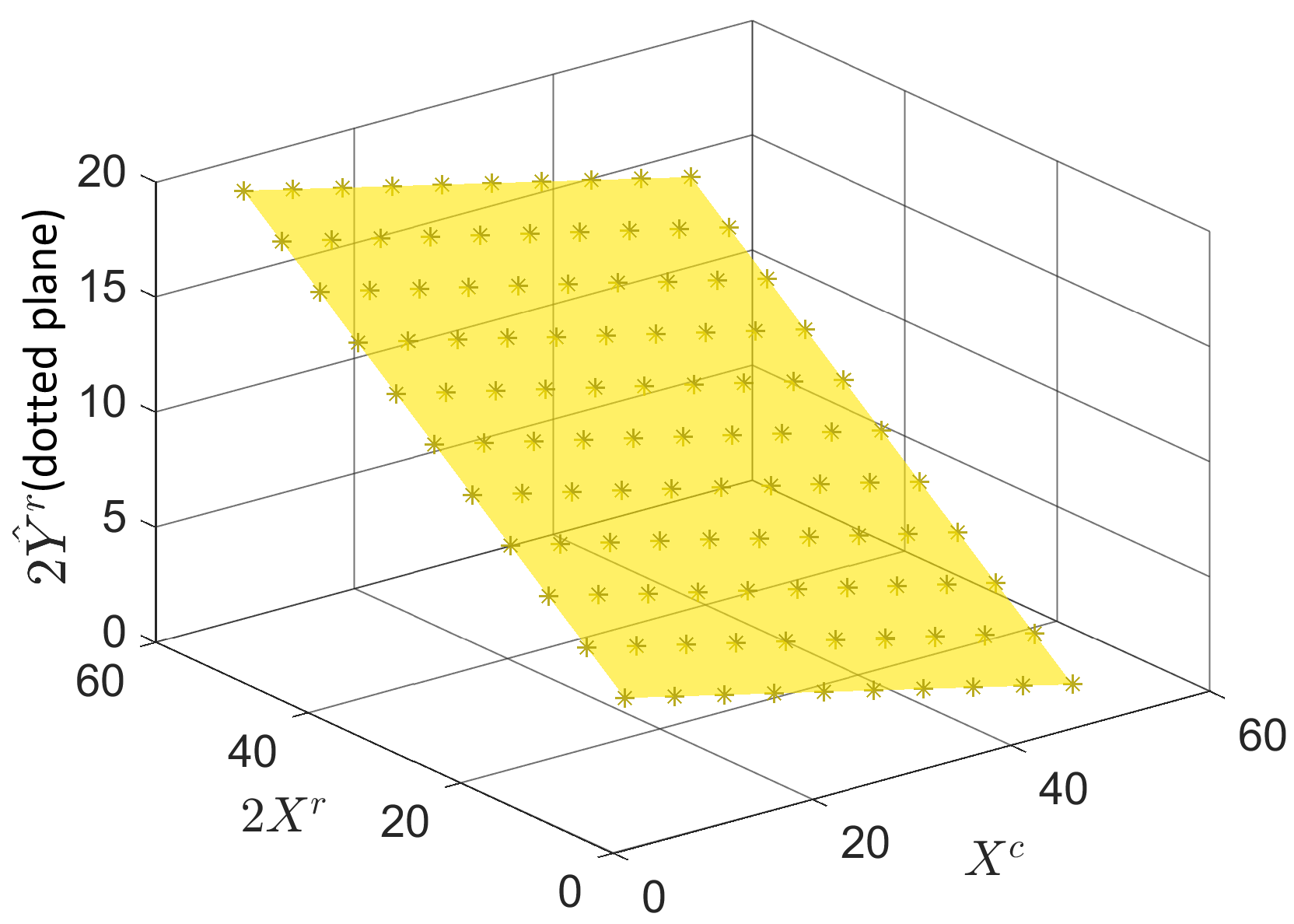}
\caption{with respect to $\hat{Y}^w$ ($\simeq 2\hat{Y}^r$)}
\end{subfigure}
\caption{\emph{IRG}s showing the relationship between \emph{systolic} and \emph{diastolic} BP in terms of center and range for the PM method.} \vspace{-1.4em}
\label{fig:systolic-IRG}
\end{figure}
\begin{figure*}[t!]
\centering \vspace{-0.4em}
\begin{subfigure}{0.29\textwidth}
\centering
\includegraphics[width=0.96\textwidth]{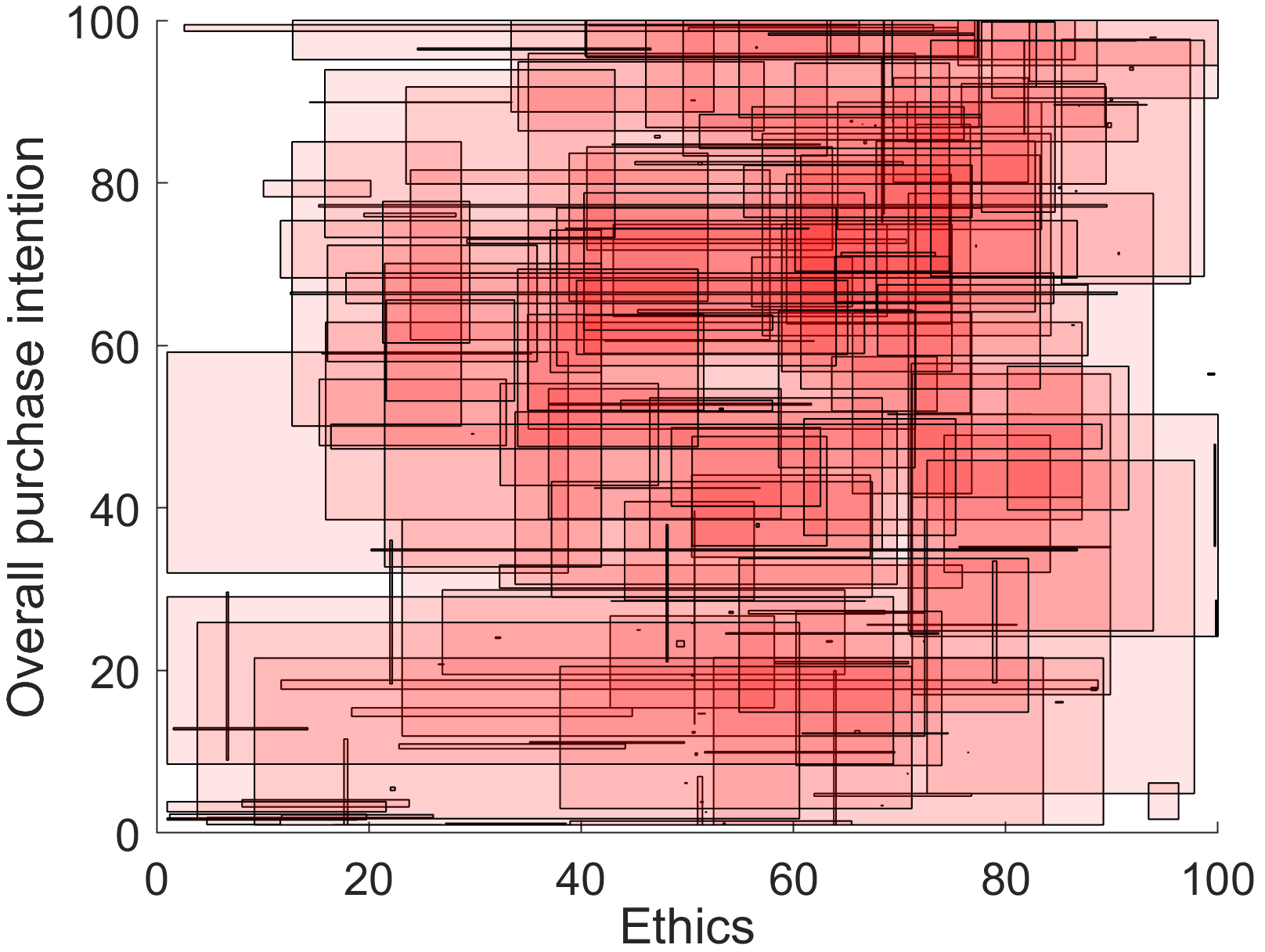}
\caption{\emph{Ethics} and \emph{overall purchase intention}}
\end{subfigure}\hfill
\begin{subfigure}{0.69\textwidth}
\centering\vspace{1.3em}
\renewcommand{\arraystretch}{1.2}
\resizebox{1\hsize}{!}{
\begin{Huge}
\begin{tabular}{ccccccc} \hline
Methods & $RMSE^-$ & $RMSE^+$ & $MAE^-$ & $MAE^+$ & $MMRE$ \\\hline\hline
CM	& 30.019 [28.708  32.576] &	30.240 [27.826  31.679] &	26.310 [24.384  28.699] &	26.592 [23.929  28.415] &	2.941 [2.082  3.792]\\
MinMax&30.491 [28.451  32.177] &	29.839 [27.750 31.569] &	26.955 [24.656  28.856] &	\textbf{26.185} [23.775  28.113] &	3.008 [2.141  3.850]\\
CRM&30.019 [28.802  31.646] &	30.240 [28.225  32.155] &	26.310 [23.925  28.277] &	26.592 [24.127  28.673] &	2.961 [2.112  3.819]\\
CCRM & 30.019 [28.802  31.646] &	30.240 [28.225  32.155] &	26.310 [23.925  28.277] &	26.592 [24.127  28.673] &	2.961 [2.112  3.819]\\
CIM & 30.624 [28.568  32.342] &	29.974 [27.906  31.758] &	26.769 [24.439  28.104] &	26.510 [24.777  28.556] &	3.014 [2.147  3.873]\\
PM & \textbf{29.631} [27.461 32.647] &	\textbf{29.825} [27.698  32.397] &	25.931 [23.389  28.190] &	26.203 [23.715  28.406] &	2.864 [1.847  3.666]\\
LM$_c$ & 29.636 [27.463  31.283] &	29.830 [27.856  31.648] &	\textbf{25.856} [23.313  27.737] &	26.250 [23.771  28.116] &	\textbf{2.859} [2.028  3.706]\\
LM$_w$ & 29.636 [27.463  31.283] &	29.830 [27.856  31.648] &	\textbf{25.856} [23.313  27.737] &	26.250 [23.771  28.116] &	\textbf{2.859} [2.028  3.706]\\\hline
\end{tabular}
\end{Huge}}
\begin{scriptsize}\textbf{Bold} = Best; Bootstrapped 95\% confidence intervals at 10000 simulations are provided.
\end{scriptsize}
\caption{Model Performance}
\end{subfigure}
\caption{Graphical presentation of \emph{ethics} and \emph{overall purchase intention}~\cite{ellerby2020insights} and performance of different regression models as to evaluation metrics using this set.}\vspace{-1.4em}
\label{fig:food_rating1}
\end{figure*}

\item \emph{Set-9 (Disjoint mixed intervals)}: 
The LM$_w$, LM$_c$ and PM methods perform best. All other methods result in relatively low fitness. In particular, the CM and CIM methods achieve the worst results. 

\item \emph{Set-10 (Some overlapping mixed intervals)}: 
The PM, LM$_c$ and LM$_w$ methods result in similar model fit for almost all metrics. The CRM and CCRM methods give better fitness results with respect to one metric ($MMRE$). 

\item \emph{Set-11 (Nested/all overlapping mixed intervals)}: 
The LM$_c$, LM$_w$, and PM methods give the best fit. The CRM and CCRM methods also closely fit the data as to $MMRE$.
\end{itemize}

\begin{figure}[t!]
\centering
\begin{subfigure}{0.24\textwidth}
\includegraphics[width=0.99\textwidth]{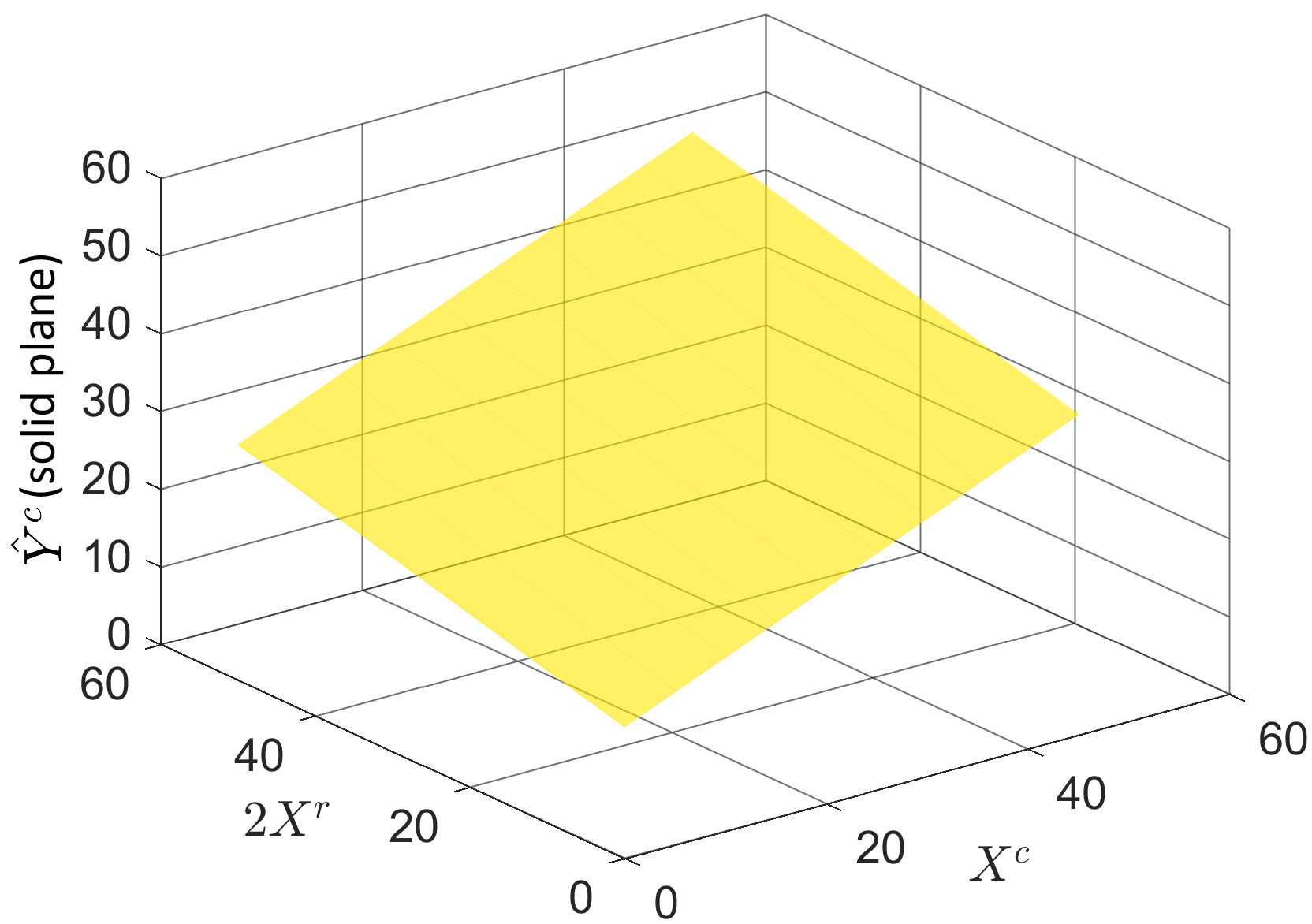}
\caption{with respect to $\hat{Y}^c$}
\end{subfigure}
\begin{subfigure}{0.24\textwidth}
\includegraphics[width=0.99\textwidth]{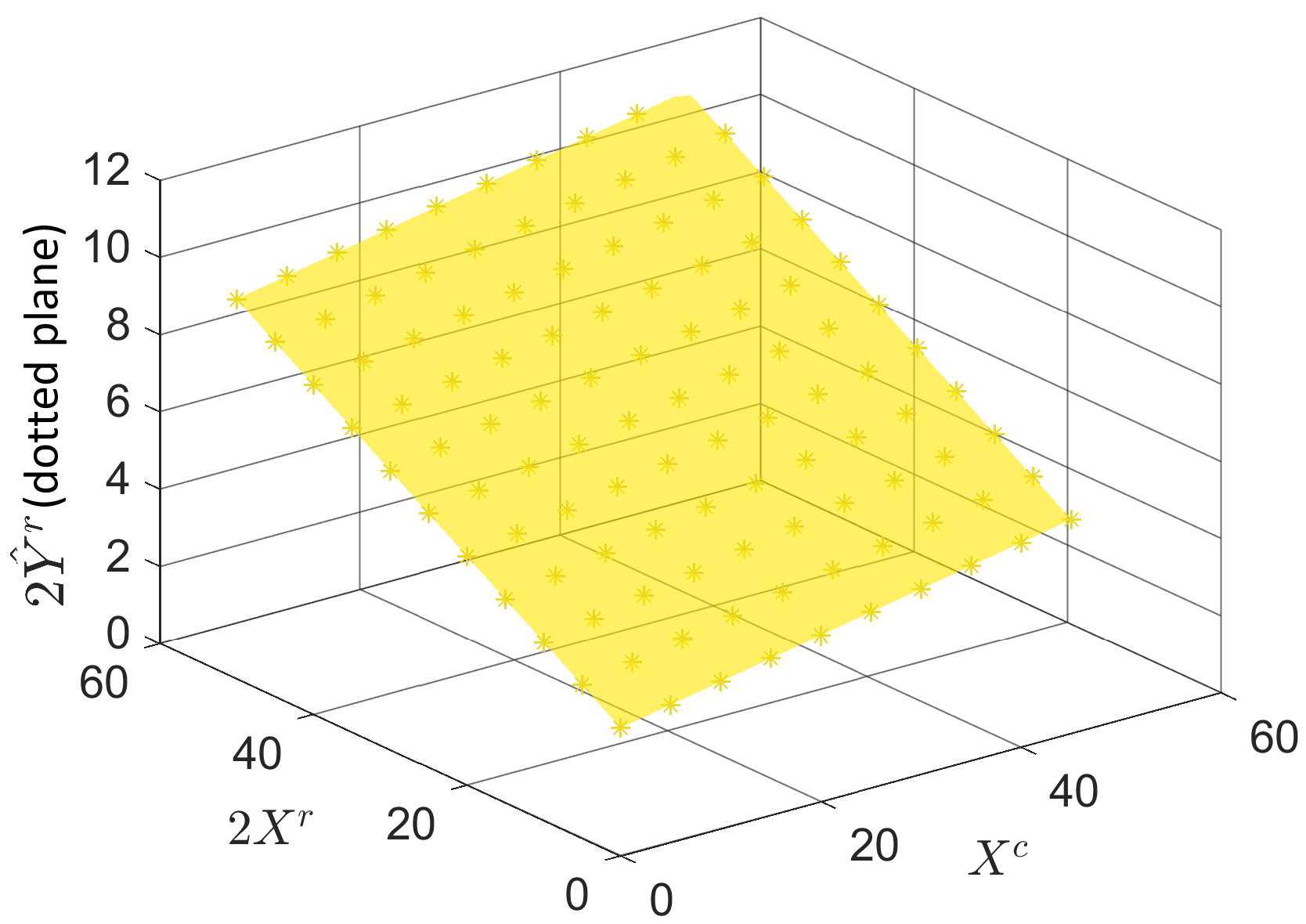}
\caption{with respect to $\hat{Y}^w$ ($\simeq 2\hat{Y}^r$)}
\end{subfigure}
\caption{\emph{IRG}s showing the relationship between \emph{ethics} and \emph{overall purchase intention}~\cite{ellerby2020insights} in terms of center and range for the PM method.}\vspace{-1.6em}
\label{fig:real-case2-IRG}
\end{figure}
\begin{figure*}[t!]
\centering \vspace{-0.8em}
\begin{subfigure}{0.29\textwidth}
\centering
\includegraphics[width=0.96\textwidth]{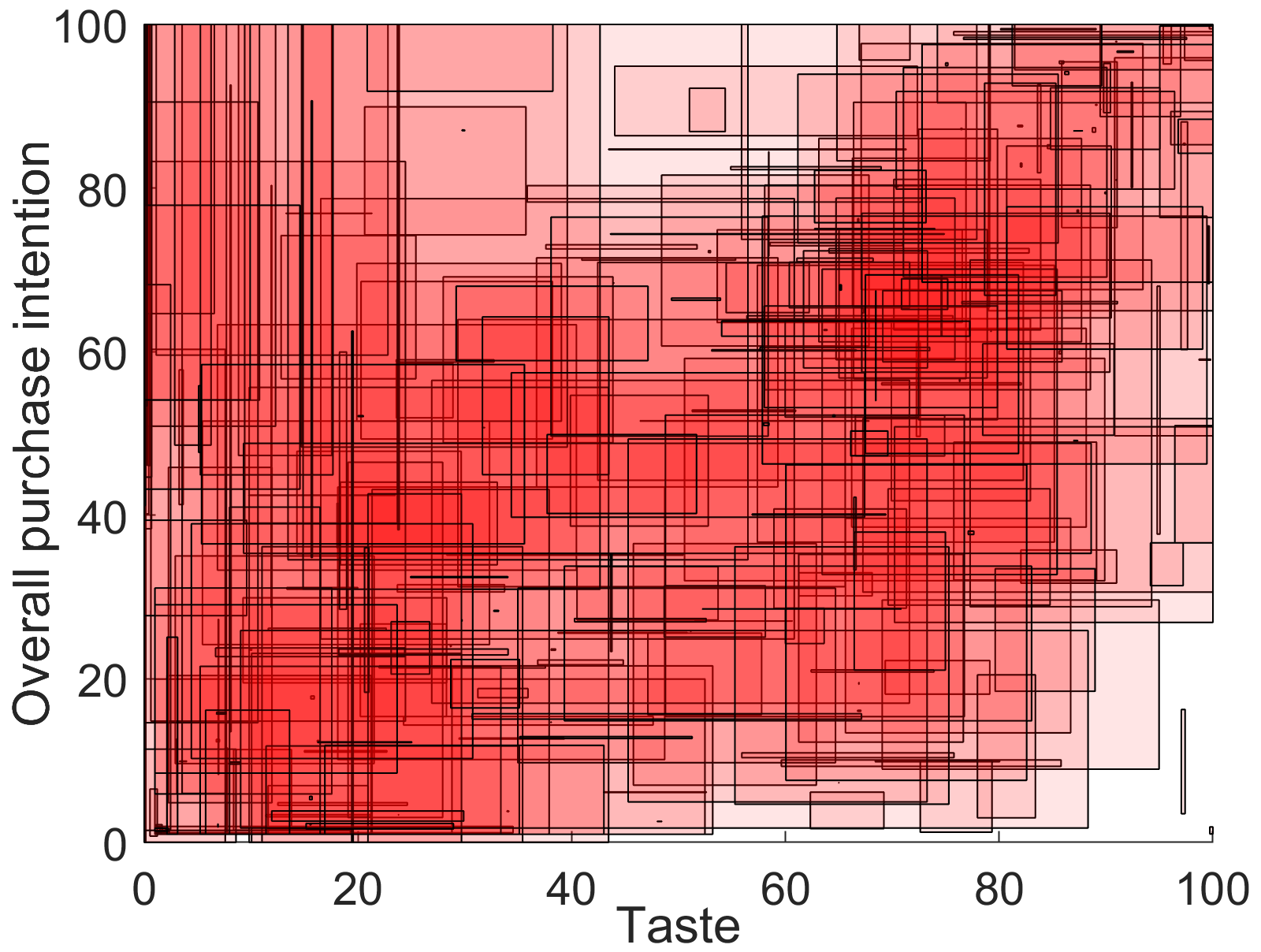}
\caption{\emph{Taste} and \emph{overall purchase intention}}
\end{subfigure}\hfill
\begin{subfigure}{0.69\textwidth}
\centering\vspace{1.3em}
\renewcommand{\arraystretch}{1.2}
\resizebox{1\hsize}{!}{
\begin{Huge}
\begin{tabular}{ccccccc} \hline
Methods & $RMSE^-$ & $RMSE^+$ & $MAE^-$ & $MAE^+$ & $MMRE$ \\\hline\hline
CM	&24.403 [22.653  26.043] &	30.492 [28.080  32.137] &	20.756 [18.979  22.375] &	24.797 [22.341  26.743] &	3.035 [2.399  3.684]\\
MinMax & 23.675 [21.965  25.223] &	29.858 [27.610  31.377] &	\textbf{19.800} [18.078  21.404] &	25.110 [22.648  26.913] &	\textbf{2.697} [2.118  3.277]\\
CRM & 24.031 [22.484  25.659] & 29.951 [27.530  31.313] &	20.552 [18.458  21.864] &	24.484 [22.349  26.677] &	2.820 [2.180  3.350]\\
CCRM & 24.031 [22.571  25.754] &	29.951 [27.631  31.478] &	20.552 [18.894  22.283] &	24.484 [22.048  26.303] &	2.820 [2.259  3.439]\\
CIM & 24.474 [22.729  26.096] & 30.457 [28.047  32.109] &	20.856 [19.086  22.446] &	24.864 [22.412  26.791] &	3.066 [2.428  3.713]\\
PM & \textbf{23.645} [21.973  51.001] &	\textbf{29.669} [27.545  65.130] &	19.806 [18.151  41.454] &	24.823 [22.466  56.996] &	2.708 [0.898  3.253] \\
LM$_c$ & 23.830 [22.088  25.393] &	29.817 [28.279  31.963] &	20.387 [18.669  21.941] &	\textbf{24.338} [22.564  26.673] &	2.821 [2.279  3.470]\\
LM$_w$ & 23.830 [22.088  25.393] &	29.817 [28.244  31.980] &	20.387 [18.669  21.941] &	\textbf{24.338} [22.516  26.637] &	2.821 [2.236  3.436]\\\hline
\end{tabular}
\end{Huge}}
\begin{scriptsize}\textbf{Bold} = Best; Bootstrapped 95\% confidence intervals at 10000 simulations are provided.\end{scriptsize}
\caption{Model Performance}\vspace{-0.4em}
\end{subfigure}
\caption{Graphical presentation of \emph{taste} and \emph{overall purchase intention}~\cite{ellerby2020insights} and performance of different regression models as to evaluation metrics using this set.}\vspace{-1.2em}
\label{fig:food_rating2}
\end{figure*}
\begin{figure}[t!]
\vspace{-0.5em}
\begin{subfigure}{0.24\textwidth}
\centering
\includegraphics[width=0.99\textwidth]{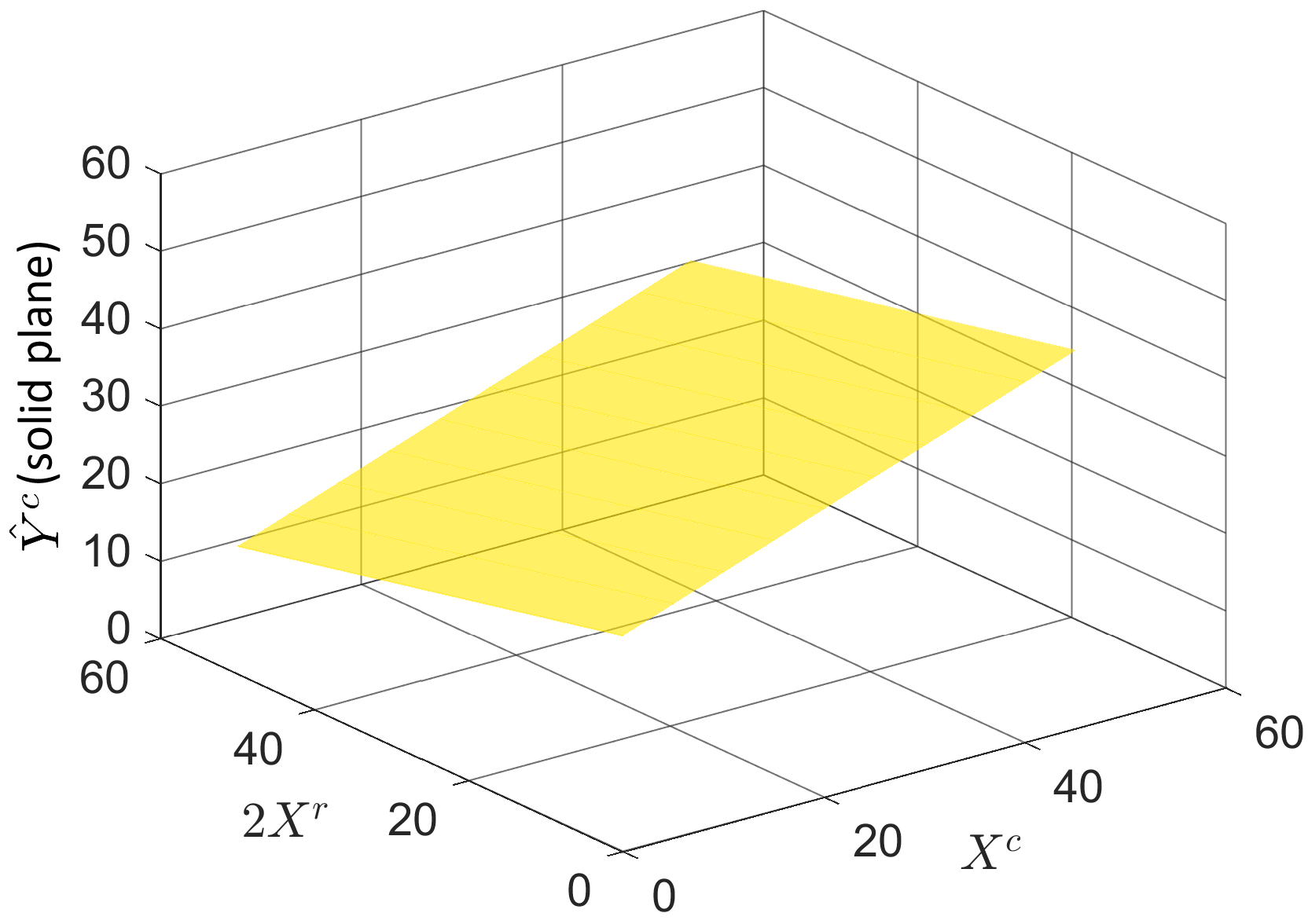}
\caption{with respect to $\hat{Y}^c$}
\end{subfigure}
\begin{subfigure}{0.24\textwidth}
\includegraphics[width=0.99\textwidth]{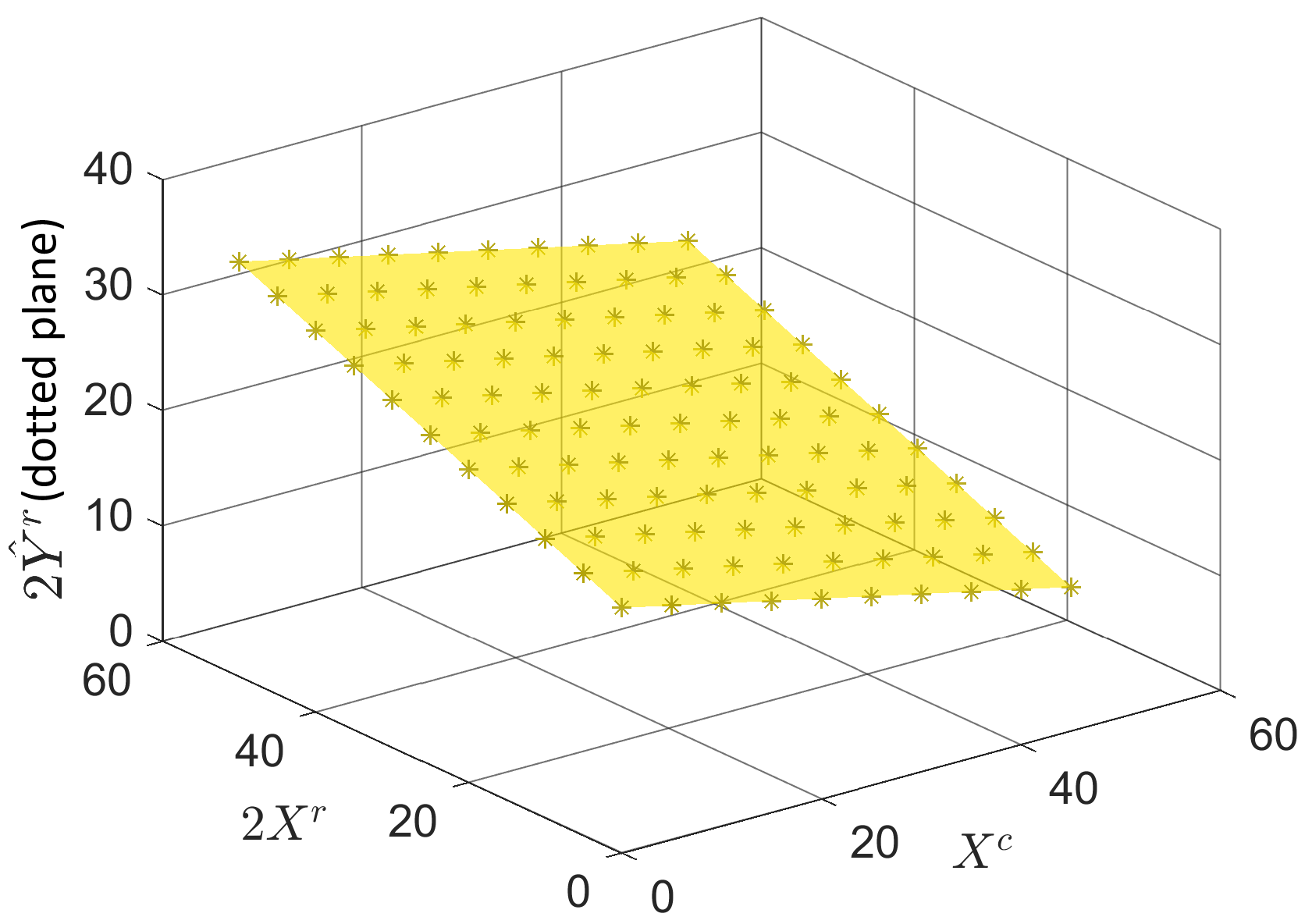}
\caption{with respect to $\hat{Y}^w$ ($\simeq 2\hat{Y}^r$)}
\end{subfigure}
\caption{\emph{IRG}s showing the relationship between \emph{taste} and \emph{overall purchase intention}~\cite{ellerby2020insights} in terms of center and range for the PM method.}\vspace{-1.4em}
\label{fig:real-case2-IRG1}
\end{figure}

\begin{figure*}[t!]
\centering \vspace{-1em}
\begin{subfigure}{0.33\textwidth}
\centering
\includegraphics[width=1\textwidth]{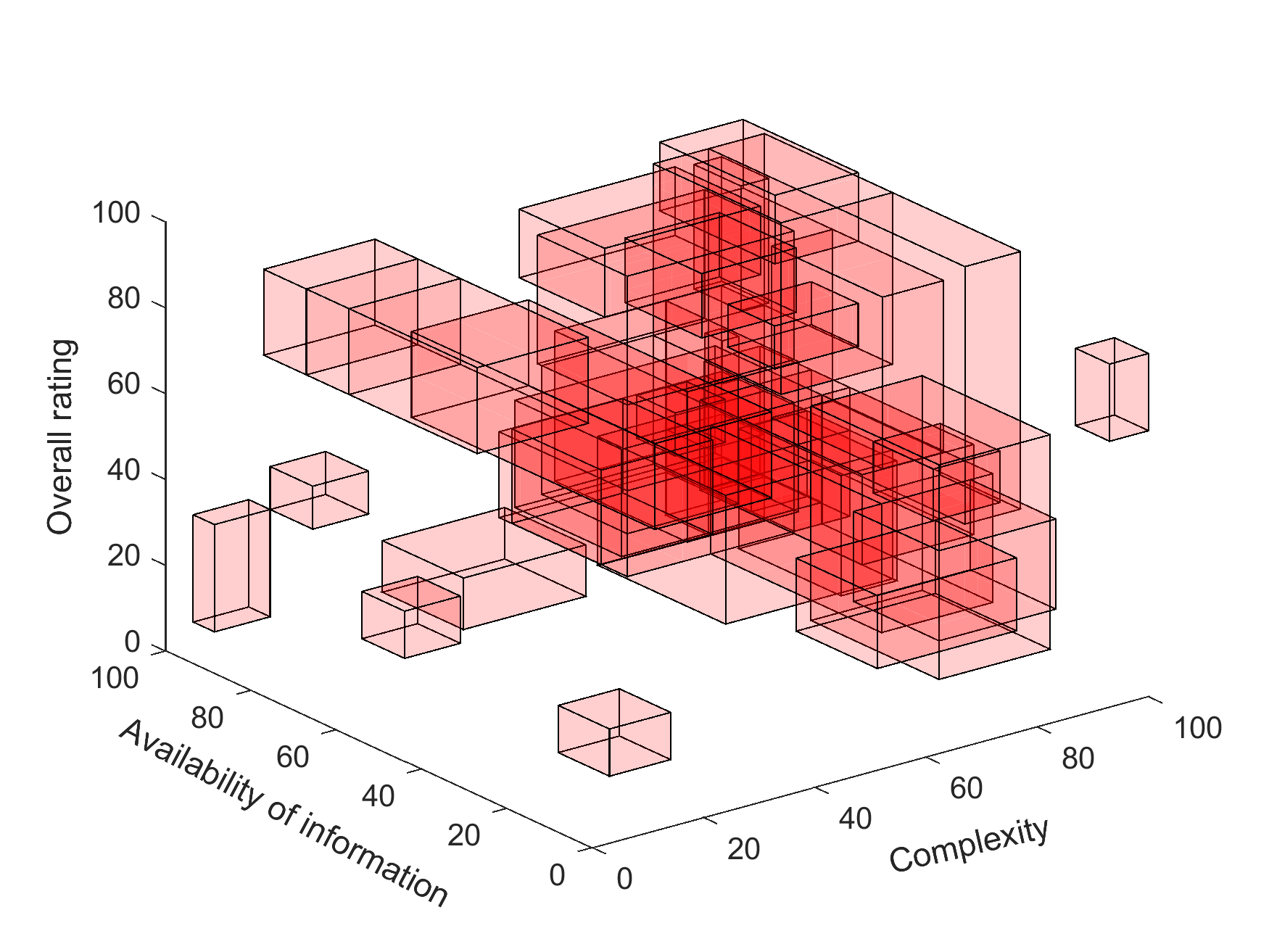}
\caption{Evade-4}
\end{subfigure}\hfill
\begin{subfigure}{0.66\textwidth}
\centering\vspace{2.5em}
\renewcommand{\arraystretch}{1.2}
\resizebox{1\hsize}{!}{
\begin{Huge}
\begin{tabular}{ccccccc} \hline
Methods & $RMSE^-$ & $RMSE^+$ & $MAE^-$ & $MAE^+$ & $MMRE$ \\\hline\hline
CM	&24.071 [21.184  29.976] & 25.124 [22.109  31.672] & 20.745	[17.909  26.513] & 20.478 [16.180  25.812] & 2.90 [1.243  4.683]	\\
MinMax	& 21.737 [16.614  24.375] & 22.856	[18.159  25.004] & 17.948	[13.028  21.189] & 19.946	[14.546  22.468] & 2.228 [0.965  3.372]	\\
CRM	&22.269	[16.612  24.434] & 21.562	[17.975  24.792] & 18.631	[13.177  21.335] & 18.641	[14.401  22.139] & 2.296 [0.949  3.294]	\\
CCRM	& 22.164 [16.670  24.470] & 21.756 [17.986  24.776] & 18.385	[13.239  21.356] & 18.820	[14.453  22.149] & 2.271 [0.950  3.279]	\\
CIM		& 22.593	[17.575  25.378] & 22.582	[18.009  25.128] & 19.297	[14.019  22.499] & 19.158	[14.037  21.830] & 2.570 [1.067  3.918]	\\
PM	& \textbf{21.681} [15.393  24.117]	&\textbf{20.984} [15.188  22.946]	& 17.852	 [11.628  20.831] & \textbf{17.541} [11.630  20.237] & \textbf{2.188}	[0.858  3.128]\\
LM$_c$	& 21.714	[15.467  23.811] & 21.194 [15.738  23.708] & \textbf{17.807} [11.759  20.374] & 17.909 [12.08  20.542]	& 2.218	[0.869  3.187]\\
LM$_w$	& 21.714	[15.467  23.811] & 21.019	[21.285  32.519] &\textbf{17.807}	[11.759  20.374] & 17.671	[17.580  27.886]  & 2.218	[0.977  3.303]\\\hline
\end{tabular}
\end{Huge}}
\begin{scriptsize}\textbf{Bold} = Best; Bootstrapped 95\% confidence intervals at 10000 simulations are provided.\end{scriptsize}
\caption{Model Performance}
\end{subfigure}
\caption{Graphical presentation of Evade-4 data set~\cite{miller2016modelling} and performance of different regression models as to evaluation metrics using this data set.}\vspace{-1.0em}
\label{fig:real_4}
\end{figure*}

Overall, the PM, and LM$_w$ (and often LM$_c$) methods perform best and result in reliably good---and almost identical---fitness for disjoint skinny or puffy intervals. With growing overlap among intervals (skinny or puffy), model fitness by the CRM, CCRM and MinMax methods generally improves. Nevertheless, they consistently fail to reach the performance levels of the PM and LM methods. For the mixed sets with puffy and skinny intervals, where these are disjoint or exhibit some overlap, both PM and both variants of the LM method result in similarly good (best) fit. Others, particularly the CRM and CCRM approaches, give nearly identical results for some metrics, for all overlapping, respectively nested, mixed cases. In all cases, the estimates by the CIM and CM methods deviate substantially from the actual values and hence they come out as poor performers. 

As well as the empirical analyses reported above, we carried out additional experiments going beyond the scope of this paper---exploring the impact of differences in mean interval size/range, differences in the standard deviation of the mean size/range, and the variation in interval centers (i.e., dispersion), on model fitness or estimation errors \cite{kabir2021interval}. Results suggest that estimation errors tend to grow for all regression methods with respect to both higher standard deviation of ranges and more distant placement of intervals (centers) \cite{kabir2021interval}. We will review this aspect with real-world data in future.

\subsection{Real-world Data Sets}
\label{subsec:real_results}
In this section, three real-world IV data sets from the application contexts described in Section~\ref{sec:methodology.data sets}, are used to explore and evaluate the behaviour of different regression methods with respect to the mixed set of features of the data (in the sense of the previous section) present in these sets. The fitness of each method is compared using the same evaluation metrics (i.e., \emph{RMSE}, \emph{MAE}, \emph{MMRE}) as before.

In addition, the larger degrees of freedom---compared to the synthetic data---of these real world data sets, permit us to empirically determine a measure of the variability associated with each performance value, utilizing a non-parametric resampling approach~\cite{efron1994introduction}. We therefore report bootstrapped (percentile) 95\% confidence intervals \cite{efron1994introduction}\cite{hesterberg2015teachers} associated with each error measure, for each method and all three real-world cases. Note here that although many of the confidence intervals overlap, this does not necessarily imply non-significance of differences between the individual regression approaches' error measures in question. As these were calculated on the same resampled data sets, the measures on each instance were non-independent, and so it is possible for variability evident in the CIs to represent common inter-simulation variance between measures, while for example one was a reliably better performer than another.

\subsubsection{Systolic and Diastolic BP Data Set}
For this data set, we regress the IV \emph{diastolic} BP (regressand) on those of the \emph{systolic} BP (regressor) to estimate the bounds of \emph{diastolic} BP. Figure~\ref{fig:systolic} presents this data set and the associated performance of the regression methods. 

Figure~\ref{fig:systolic}(b) shows that the PM, LM$_w$, and LM$_c$ methods produce comparable performance in estimating the regressand bounds. With respect to some of the evaluation metrics (\emph{RMSE$^+$} and \emph{RMSE$^-$}, and \emph{MAE$^+$}), the PM produces the best estimation, whereas the LM$_w$ and LM$_c$ methods perform better for \emph{MAE$^-$}; all these three methods  perform equally best for \emph{MMRE}. Fit for the CRM, CCRM and MinMax methods for these data is also better according to \emph{MMRE}. In this particular case, the CM and CIM methods result in the worst fit. 

Along with evaluating model fitness, we also explore the relationship between the uncertainty (and position) of the \emph{systolic} BP and that of the \emph{diastolic} BP in Fig.~\ref{fig:systolic-IRG}. Note that for space saving, we only present here the \emph{IRG}s using the PM method. Figures~\ref{fig:systolic-IRG}(a) and (b) show that the center of \emph{systolic} BP is positively correlated with the center of the \emph{diastolic} BP and the range of \emph{systolic} BP exhibits the same trend, i.e., increasing rapidly with the increasing range of the \emph{diastolic} BP. These figures also highlight that the center of \emph{systolic} BP decreases, at a lesser rate, as the range in \emph{diastolic} BP increases, and vice versa, i.e., the range of \emph{systolic} BP decreases as the center of \emph{diastolic} BP increases.

\subsubsection{Food Snacks Purchase Intention Data Set}
With this data set, we first remove outliers and then separately regress the \emph{overall purchase intention} (regressand) on each of the six attributes of snack-bars (i.e., \emph{visual appeal}, \emph{value for money}, \emph{healthiness}, \emph{taste}, \emph{branding}, and \emph{ethics}). Due to page limitations, we present regression results only with respect to two attributes---\emph{ethics} and \emph{taste} (see Figs.~\ref{fig:food_rating1} and \ref{fig:food_rating2}). The estimation results in Figs.~\ref{fig:food_rating1}(b) and \ref{fig:food_rating2}(b) show that the PM, LM$_w$, LM$_c$, and MinMax methods produce the best fit with respect to one or more indices. However, bootstrap confidence intervals indicate substantially greater proximity and variability in these estimates than for the previous BP data set. This is especially true for the PM method for the data set representing \emph{taste} and \emph{overall purchase intention}---demonstrating how in cases where transformation is required to maintain mathematical coherence, this can have a large detrimental impact on model fit, as discussed earlier.

\begin{figure*}[t!]
\centering \vspace{-0.2em}
\begin{subfigure}{0.33\textwidth}
\centering
\includegraphics[width=1\textwidth]{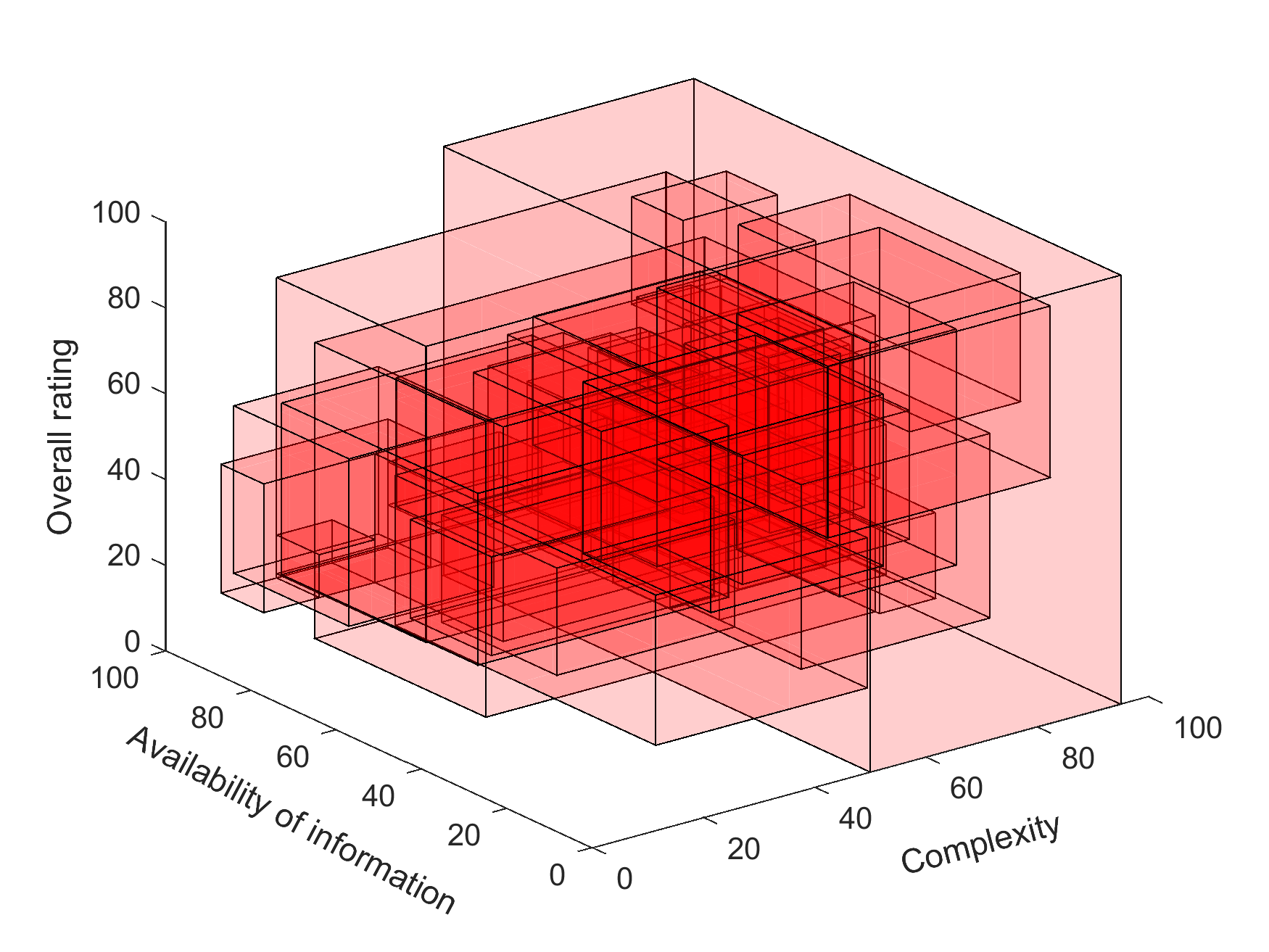}
\caption{Evade-22}
\end{subfigure}\hfill
\begin{subfigure}{0.66\textwidth}
\centering\vspace{2.5em}
\renewcommand{\arraystretch}{1.2}
\resizebox{1\hsize}{!}{
\begin{Huge}
\begin{tabular}{ccccccc} \hline
Methods & $RMSE^-$ & $RMSE^+$ & $MAE^-$ & $MAE^+$ & $MMRE$ \\\hline\hline
CM	&27.132	[23.396  39.721] & 28.312	[21.275  39.518] & 22.759	[20.573  34.154] & 23.112 [16.451  32.551] & 4.358	[2.406  6.785]\\
MinMax	&14.449	[10.331  16.469] & 17.972	[12.832  20.861] & 12.051	[8.264  13.907] & 14.591	[10.266  17.171] & 2.106	[1.035  2.765]\\
CRM	&12.491 [10.186  16.526]	&15.399	[12.333  18.632] & 10.194	[7.794  13.490] & 12.703	[9.816  16.059] & 1.492 [0.943  2.381]\\
CCRM	&12.487	 [10.305  16.552] & 15.403 [12.540  18.743] & 10.186 [8.0  13.516] & 12.714	[9.994  16.195] & 1.485  [0.886  2.357]\\
CIM		&15.636	[11.698  17.887] &18.706 [13.981  21.062] & 13.405 [9.767  15.599] & 16.011	[11.577  17.740] & 2.571	[1.315  3.530]\\
PM	&\textbf{12.333} [8.158  14.186]	& \textbf{15.220}	[9.978  16.970] & \textbf{9.862}	[6.354  11.816] & 12.756	[7.610  14.280] & \textbf{1.352} [0.647  1.807]	\\
LM$_c$	& 12.373	[8.318  14.117] & 15.255	[10.667  42.247] & 9.903	[6.503  11.798] & \textbf{12.652}	[8.284  37.102] & 1.393	[0.726  1.989]\\
LM$_w$	& 12.373 [8.318  14.117] & 15.253	[24.949  50.146] & 9.903	[6.503  11.798] & \textbf{12.652}	[8.284   44.468] & 1.392	[0.876  2.115]\\\hline
\end{tabular}
\end{Huge}}
\begin{scriptsize}\textbf{Bold} = Best; Bootstrapped 95\% confidence intervals at 10000 simulations are provided.\end{scriptsize}
\caption{Model Performance}
\end{subfigure}
\caption{Graphical presentation of Evade-22 data set~\cite{miller2016modelling} and performance of different regression models as to evaluation metrics for this data set.}\vspace{-1.4em}
\label{fig:real_22}
\end{figure*}

Beyond the evaluation of model fit, we also use the \emph{IRG}s to explore how the selected attributes of snack-bars influence overall purchase intention, as discussed in \cite{ellerby2020insights}. We construct \emph{IRG}s to show the impact of two attributes---\emph{ethics} and \emph{taste} on those of the \emph{overall purchase intention}. Again, to save space, we only show the \emph{IRG}s corresponding to the PM method. Figure~\ref{fig:real-case2-IRG} shows how the position and uncertainty of \emph{ethics} affect those of \emph{overall purchase intention}. The center of \emph{overall purchase intention}, $\hat{Y}^c$ increases with the center of \emph{ethics}, $X^c$, as well as its range, $2\hat{X}^r$. Similarly, uncertainty around \emph{overall purchase intention}, $2\hat{Y}^r$ is positively influenced to some extent by both position, $X^c$ and associated uncertainty of the \emph{ethics}, $2X^r$, though much more strongly by the latter. The \emph{IRG}s (shown in Fig.~\ref{fig:real-case2-IRG1}) relating to \emph{taste} also suggest implications for both position and width on each aspect of \emph{overall purchase intention}. However, comparing the \emph{IRG}s for \emph{taste} and \emph{ethics}, we observe that the position and uncertainty of two attributes tend to have distinctive impacts on \emph{overall purchase intention}. For example, increasing uncertainty in \emph{ethics} (together with increasing center values) increases position of \emph{overall purchase intention}, whereas this tends to decline for rising uncertainty in \emph{taste} (while also increasing for higher center values). 

These results, which are uniquely captured by the IV data, are made visible via the \emph{IRG}s. We emphasise that the regression analysis using only two variables as shown here, while a useful illustration, is limited and a broader analysis would be expected to extract real-world insights. We point the interested reader to \cite{ellerby2020insights}, where a more detailed analysis---but based on a discrete representation of the data, is conducted.
 
\subsubsection{Cyber-security Data Set}
\label{subsec:data}
Section~\ref{sec:methodology.data sets} provided an overview of this data set. Here, we only consider the evade hops as with three attributes these provide a concise basis for presenting the results within the constraints of this paper. For each evade hop, 38 experts gave ratings on following three attributes: \emph{complexity}, \emph{availability of information} and \emph{maturity}. They were also asked to provide an \emph{overall rating} on how difficult it would be \emph{overall} for an attacker to evade a given hop. These ratings were given on a scale from 0 to 100. Table~\ref{tab:evade} summarises the attributes for the evade hops. 

We explore how well the regression methods estimate the IV \emph{overall} difficulty ratings for selected evade components, based upon the three interval-valued attribute ratings, and by comparison with experts' actual overall ratings. For conciseness, we present two evade hops -- number 4 and number 22, shown in Figs.~\ref{fig:real_4} and~\ref{fig:real_22} respectively.

\begin{table}[t!]
\centering
\caption{Attributes for evade hops and associated questions}
\resizebox{1\hsize}{!}{
\begin{tabular}{l l} \hline
Attributes & Related Question\\\hline
Complexity & How complex is the job of providing this kind of defence?\\
Availability of & How likely is that there will be publicly  available  information\\
information & that could help with evading defence?\\
 Maturity & How mature is this type of technology?\\\hline
 Overall rating & Overall, how difficult would it be for an attacker to do this?\\\hline
\end{tabular}}\vspace{-1.2em}
\label{tab:evade}
\end{table} 

Figure~\ref{fig:real_4}~(b) shows that the PM method appears to be the best performer for Evade-4 with respect to four evaluation metrics (\emph{RMSE$^-$}, \emph{RMSE$^+$}, \emph{MAE$^+$}, and \emph{MMRE}) whereas the LM$_w$ method produces the joint best fit on \emph{MAE$^-$}. The LM$_c$ method also performs very close to the LM$_w$ method. Again, the CM method provides poor fit to the data. As for the Evade-4 data set, Fig.~\ref{fig:real_22}~(b) displays similar results for Evade-22. Particularly, Fig.~\ref{fig:real_22} reveals the PM method as the best approach in terms of most metrics. Further, the LM$_w$, LM$_c$, CRM, and CCRM methods come out as suitable measures as to either $RMSE^+$ or $MAE^+$. We provide all estimated coefficients for both synthetic and real-world data sets in Appendix (see Table~\ref{tab:my_label11} and \ref{tab:my_label22}). Note here again that bootstrap confidence intervals indicate substantially higher variability in model fit than for the first data set. However, in contrast to the second data set---for which the PM method often showed substantially higher upper confidence bounds than the other methods---the PM method's upper confidence bounds are now often the lowest, or nearly lowest. This may indicate relatively few cases in which transformation was required to maintain mathematical coherence for the PM method with this data set, avoiding the associated model fitness penalties.

Across the experiments, confidence bounds for the PM and the LM methods are overall fairly similar. For the BP data, Fig.~\ref{fig:systolic}, errors in model fit are fairly small in size due to densely placed intervals. Similarly, for the cyber-security data sets, Figs.~\ref{fig:real_4} and \ref{fig:real_22}, the PM and the LM methods show similar variability in confidence bounds, but the errors are larger as the intervals are relatively scattered, invariably impacting the feasibility of a linear model fit. Similarly, for the food snacks data with respect to \emph{ethics}, Fig.~\ref{fig:food_rating1}, the methods show similar variability in CI bounds, but bigger errors in terms of size---similar to the cyber-security data set.

The only exception where the PM method shows wider variability in CI bounds than other approaches is for the \emph{taste} vs \emph{overall purchase intention} data set, Fig.~\ref{fig:food_rating2}. This higher variability in CI bounds may be due to greater incidence of very wide and scattered intervals within the data set, resulting in increased application of the box-cox transformation and poorer fit across a larger subset of samples/simulations. 

Figure~\ref{fig:statistical analysis} in Appendix shows the total number of times across the 10000 simulations the regression approaches (CCRM, PM, LM$_c$, and LM$_w$) apply the positivity restrictions/regressand bound transformations to maintain coherence for the real-world data sets. Figure~\ref{fig:statistical analysis} reveals that for the \emph{taste} vs \emph{overall purchase intention} data set, for a significant proportion (22.85\% of total 10000 simulations), positivity restrictions/transformation of bounds are applied by the PM and LM$_w$ methods, compared to other sets. Exactly why the PM method is more strongly impacted than the LM$_w$ method for this case requires further research which we will address in a future publication. 

\section{Discussion on Suitability of Existing Regression Approaches}
\label{sec:discussion}
This section discusses the suitability of different regression techniques with respect to interval data sets with different properties (e.g., skinny, puffy or mixed), and based upon the results of the analyses conducted above. The experiments consistently show that the PM and LM$_w$ (and often LM$_c$) generally produce the best model fit---with reliably good performance for skinny/puffy or mixed sets of intervals, when these are disjoint or overlapping. With increasing overlap among intervals, estimation errors also decrease for the CRM, CCRM and MinMax methods, but not generally so far as to equal the performance of either PM or LM methods. In all cases, both MinMax and CRM require checking as to whether they maintain consistency of interval bounds. The real-world data experiments largely substantiate the synthetic experiments in terms of relative model performance. 

Results reveal that the LM$_w$ method produces better fitness than its permanently constrained sibling (LM$_c$) in most cases, as the weaker constraining approach lets model parameters vary freely unless mathematical coherence is being compromised. Both approaches generate the same outcome when the coherence of the bounds does break down and LM$_w$ effectively reverts to LM$_c$. Similarly, the PM approach may produce much poorer fit in cases where it requires the Box-Cox transformation~\cite{box1964analysis,souza2017parametrized} of the regressand to restore coherence of the bounds. We provide such an example in Appendix (see Fig.~\ref{fig:box_cox_PM}), and we also hypothesise that this is the cause of the high upper 95\% bootstrap confidence bounds in relation to this model's performance on our second real-world data set. Generally, both CM and CIM methods yield poor model-fit. 

\begin{figure}
\vspace{-0.2em}
\centering
\includegraphics[width=1\columnwidth, height=4.8cm]{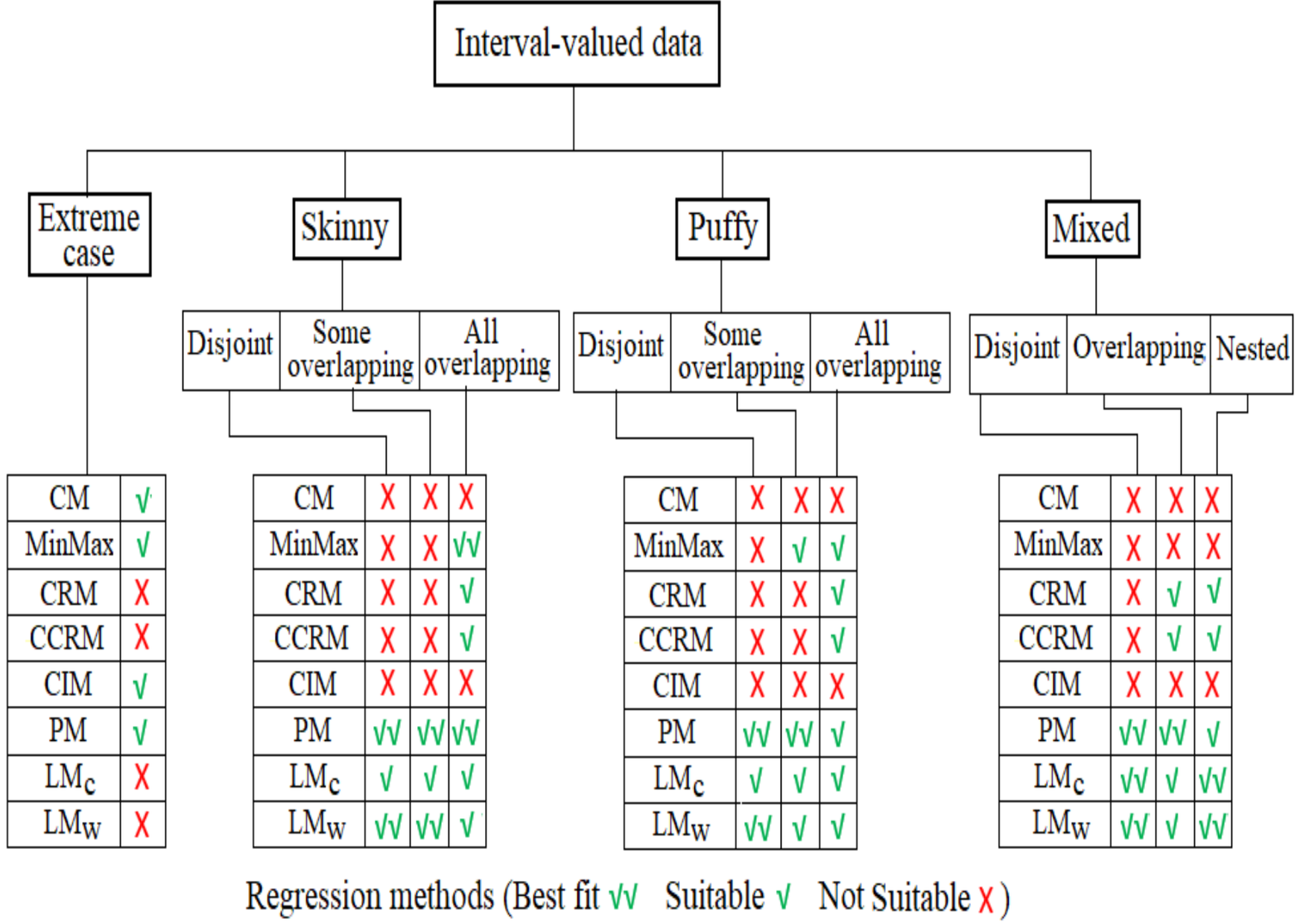}
\caption{A flowchart presenting recommendations on linear regression methods with respect to different IV data sets.}\vspace{-1.6em}
\label{fig:flowchart}
\end{figure}

Having said that, the PM and LM models (particularly its weakly-constrained form, LM$_w$) were overall the most widely suitable regression approaches. Nevertheless, as we discuss throughout, better performance can be found for data sets following specific properties. To help researchers choose the most suitable method for their data, Fig.~\ref{fig:flowchart} provides a flowchart to summarise the findings presented. The figure provides a recommendation on suitability of regression approaches as to different features of the intervals in given data sets. 

In addition to comparing model fitness among regression approaches, we have also visualized the relationship between IV variables with respect to their key features (i.e., center and range) by introducing the \emph{IRG}. The \emph{IRG} provides a powerful visual tool, akin to the `regression line' of traditional numeric regression, offering rapid, intuitive insight into the data. While the \emph{IRG} can be generated for all regression models, irrespective of the regression method, we generally favour the PM method as consistently producing strong model fit.

Lastly, as for computational efficiency, the LM$_c$ generally requires more execution time than LM$_w$ as it generically restricts parameters. Similarly, the PM approach involves extra execution time when it applies the Box-Cox transformation to restore consistency, in other words, its execution time is dependent on the data set. Overall, in this paper we do not explore execution time in more detail, instead focusing on performance as this is the more common driver of data analysis choices in the domains considered.

\section{Conclusions}
\label{sec:conclusion}
Inferential statistics and in particular (linear) regression represent a key building block of AI. For numeric variables, linear regression does not only afford a model for estimation, it provides powerful insight into the relationships between variables. Indeed, the regression-line or graph is perhaps the most widely understood visualisation in data analysis. 

While well understood for the numeric case, linear regression for `fully' IV data sets, i.e. data sets where both regressor and regressand are IV, remain challenging and an active research topic. At the same time, the potential for insights from such data sets is extremely promising, not only for regression, but for AI in general.

The complexity of IV data in terms of how intervals are distributed, how they vary in size, how they overlap--or not, makes it notoriously challenging to establish universal superiority of a single technique. Acknowledging this, in this paper, we have carried out an in-depth review and analysis of linear regression models, exploring their behaviour with respect to systematically designed synthetic and real-world, fully IV data sets, with a view to support researchers and practitioners in the choice of techniques suitable to their problem and data.

Building on existing literature, we have articulated key features of IV data, such as whether intervals are \emph{puffy, skinny, scattered, densely placed} or \emph{overlapping}. For all methods, we have investigated how they do, or do not, maintain mathematical coherence, and the impact of ensuring coherence on model-fitness. As part of this, we also extended current methods, proposing two variants of the LM method~\cite{sun2015linear} to enhance its general suitability for real-world applications by ensuring mathematical coherence of bounds.  

Going beyond the review of the state of the art and refinement of regression approaches, the paper discusses the potential of IV data in regression and AI more broadly, in particular, highlighting the value of mapping not only variable \emph{position} (as in the numeric case), but also uncertainty/range---for example to infer how uncertainty about a regressor, such as the ethical origin of a snack-food product's ingredients, impacts uncertainty---and/or position of a regressand, such as a consumer's purchase-intention. 

Acknowledging the powerful aspect of visualization in terms of the popularity of traditional linear regression, we have introduced the Interval Regression Graph (\emph{IRG}), facilitating the interpretation of IV regression models, and by extension, IV data. \emph{IRG}s succinctly visualize and articulate the relationship between IV regressor and regressand as to both position (center) and uncertainty (range). As \emph{IRG}s use all three visualizable dimensions, they are limited to the visualization of the relationship between two variables at a time. Naturally, this can be applied to multivariate cases by presenting the relationship between two variables while keeping other variables constant. An alternative to this approach could be to leverage techniques such as Principal Component Analysis (PCA) for intervals \cite{gioia2006principal}, and visualizing the relationships between resulting components. However, here, the intuitive interpretation of said intervals will be complex, potentially limiting one of the primary assets of leveraging intervals in real-world settings as discussed in this paper.  We will explore these aspects in future work.

Finally, as alluded to above, we have leveraged the empirical results to provide recommendations on which regression models are best suited to which type of IV data sets, based on the data's features, such as whether intervals strongly overlap, are narrow or wide, etc. This is supported by complementary insights, such as generally better model prediction observed for IV sets with narrow range or condensed interval placement. Further, in this paper, we have provided open-source software which--for the first time--offers access to the suite of IV regression approaches for facilitating adoption and replication.

In summary, analyses have shown that the PM~\cite{souza2017parametrized} and the LM~\cite{sun2015linear} methods (with its weaker set of restrictions, i.e., LM$_w$) are the most suitable regression approaches in many cases including where intervals are skinny, puffy or a mix thereof. The constrained variant of the LM model~\cite{sun2015linear}, LM$_c$ also performs equally well in many cases. However, when intervals are densely placed or some overlapping/mixed, other existing methods, such as CRM~\cite{neto2008centre}, CCRM~\cite{neto2010constrained}, and MinMax~\cite{billard2002symbolic} may turn into suitable estimation approaches, offering both relatively simple implementation and efficient computation. In such instances, it is desirable to verify if the CRM~\cite{neto2008centre} and MinMax~\cite{billard2002symbolic} methods maintain consistency of the bounds. Certainly, maintaining this coherence automatically, such as by the PM, LM, and CCRM methods, tends to require more execution time. The latter may be relevant as the number of variables increases. 

With increasing popularity of IV data, being able to analyse and derive the important insights available from these data using AI becomes an increasingly important topic. The substantial body of work advancing IV regression over the last ten years, and advances such as those put forward in this paper, provide the foundation for leveraging intervals more generally in AI. Substantial work remains, from further refining our capacity for communicating resulting insights to advancing our modelling techniques, such as to non-linear approaches.   

\begin{table*}[b]
\centering \vspace{-1.2em}
\caption{Synthetic IV data sets used for sensitivity analysis (Set-1 to Set-11)}
\begin{subtable}{.24\linewidth}
\centering
\caption{Set-1}
\resizebox{1\columnwidth}{!}{
\begin{Huge}
\begin{tabular}{c|cc|cc} \hline
Entry&$X^-$	&$X^+$&$Y^-$&$Y^+$ \\\hline
1& 0.6	&0.95 &1.0	&2.0\\
2& 1.4	&1.92 &1.01	&3.01\\
3& 2.4	&3.13 &1.02	&4.02\\
4& 3.6	&4.45 &1.03	&5.03\\
5& 5.0  &5.95 &1.04 &6.04\\\hline
&$\mu_{X^c}$=2.94 & $\mu_{X^w}$=0.68 & $\mu_{Y^c}$=2.52  & $\mu_{Y^w}$=3.0 \\
& $\sigma_{X^c}$=1.87 & $\sigma_{X^w}$=0.24 & $\sigma_{Y^c}$=0.81 &  $\sigma_{Y^w}$=1.58\\\hline
\end{tabular}
\end{Huge}}
\end{subtable}
\begin{subtable}{.24\linewidth}
\centering
\caption{Set-2}
\resizebox{1\columnwidth}{!}{
\begin{Huge}
\begin{tabular}{c|cc|cc} \hline
Entry&$X^-$	&$X^+$&$Y^-$&$Y^+$ \\\hline
1& 0.6	&0.65 &1.0	&2.0\\
2& 1.4	&1.44 &1.01	&3.01\\
3& 2.4	&2.46 &1.02	&4.02\\
4& 3.6	&3.61 &1.03	&5.03\\
5& 5.0  &5.03 &1.04 &6.04\\\hline
& $\mu_{X^c}$=2.62 & $\mu_{X^w}$=0.04 & $\mu_{Y^c}$=2.52 &$\mu_{Y^w}$=3.0 \\
& $\sigma_{X^c}$=1.74 & $\sigma_{X^w}$=0.02 & $\sigma_{Y^c}$=0.81 &  $\sigma_{Y^w}$=1.58\\\hline
\end{tabular}
\end{Huge}}
\end{subtable}
\begin{subtable}{.24\linewidth}
\centering
\caption{Set-3}
\resizebox{1\columnwidth}{!}{
\begin{Huge}
\begin{tabular}{c|cc|cc} \hline
Entry&$X^-$	&$X^+$&$Y^-$&$Y^+$ \\\hline
1& 4.4	&4.7 &3.8	&4.2\\
2& 3.4	&3.8 &5.8	&6.3\\
3& 4.8	&5.3 &7.5	&7.8\\
4& 6.4	&6.7 &5.4	&5.8\\
5& 8.1  &8.5 &6.8   &7.3\\\hline
& $\mu_{X^c}$=5.61 & $\mu_{X^w}$=0.38& $\mu_{Y^c}$=6.07 &$\mu_{Y^w}$=0.42 \\
& $\sigma_{X^c}$=1.84 &$\sigma_{X^w}$=0.08& $\sigma_{Y^c}$=1.41&  $\sigma_{Y^w}$=0.08\\\hline
\end{tabular}
\end{Huge}}
\end{subtable}
\begin{subtable}{.24\linewidth}
\centering
\caption{Set-4}
\resizebox{1\columnwidth}{!}{
\begin{Huge}
\begin{tabular}{c|ll|ll} \hline
Entry&$X^-$	&$X^+$&$Y^-$&$Y^+$ \\\hline
1& 3.7&	4.0 & 5.5&	5.9\\
2& 3.4&	3.8 & 5.8&	6.3\\
3& 4.8&	5.3 & 7.5&	7.8\\
4& 7.9&	8.2 & 6.5&	6.9\\
5& 8.1&	8.5 & 6.8&	7.3\\\hline
& $\mu_{X^c}$=5.77 & $\mu_{X^w}$=0.38& $\mu_{Y^c}$=6.63 &$\mu_{Y^w}$=0.42 \\
& $\sigma_{X^c}$=2.26 &$\sigma_{X^w}$=0.08& $\sigma_{Y^c}$=0.78 &  $\sigma_{Y^w}$=0.08\\\hline
\end{tabular}
\end{Huge}}
\end{subtable}
\\%
\begin{subtable}{.24\linewidth}
\centering
\caption{Set-5}
\resizebox{1\columnwidth}{!}{
\begin{Huge}
\begin{tabular}{c|ll|ll} \hline
Entry&$X^-$	&$X^+$&$Y^-$&$Y^+$ \\\hline
1& 5.4&	5.7&	6.4&	6.8\\
2& 5.6&	6.0&	6.1&	6.6\\
3& 5.3&	5.8&	6.6&	6.9\\
4& 5.5&	5.8&	6.6&	7.0\\
5& 5.7&	6.1&	6.3&	6.8\\\hline
& $\mu_{X^c}$=5.69 & $\mu_{X^w}$=0.38& $\mu_{Y^c}$=6.61 &$\mu_{Y^w}$=0.42 \\
& $\sigma_{X^c}$=0.16 &$\sigma_{X^w}$=0.08& $\sigma_{Y^c}$=0.18 &  $\sigma_{Y^w}$=0.08\\\hline
\end{tabular}
\end{Huge}}
\end{subtable}
\begin{subtable}{.24\linewidth}
\centering
\caption{Set-6}
\resizebox{1\columnwidth}{!}{
\begin{Huge}
\begin{tabular}{c|ll|ll} \hline
Entry&$X^-$	&$X^+$&$Y^-$&$Y^+$ \\\hline
1& 1.2	&2.4	&4.8	&6.8\\
2& 2.6	&4.2	&3.6	&5.8\\
3& 3.8	&5.6	&7.6	&9.5\\
4& 5.0	&7.8	&2.6	&4.5\\
5& 7.6	&9.4	&5.0	&7.6\\\hline
& $\mu_{X^c}$=4.96 & $\mu_{X^w}$=1.84& $\mu_{Y^c}$=5.78 &$\mu_{Y^w}$=2.12 \\
& $\sigma_{X^c}$=2.60 &$\sigma_{X^w}$=0.59& $\sigma_{Y^c}$=1.88 &  $\sigma_{Y^w}$=0.29\\\hline
\end{tabular}
\end{Huge}}
\end{subtable}
\begin{subtable}{.24\linewidth}
\centering
\caption{Set-7}
\resizebox{1\columnwidth}{!}{
\begin{Huge}
\begin{tabular}{c|ll|ll} \hline
Entry&$X^-$	&$X^+$&$Y^-$&$Y^+$ \\\hline
1& 1.2&	2.4&	4.8&	6.8\\
2& 2.0&	3.6&	3.3&	5.5\\
3& 3.8&	5.6&	7.6&	9.5\\
4& 5.8&	8.6&	4.2&	6.1\\
5& 7.4&	9.2&	4.8&	7.4\\\hline
& $\mu_{X^c}$=4.96 & $\mu_{X^w}$=1.84& $\mu_{Y^c}$=6.00 &$\mu_{Y^w}$=2.12 \\
& $\sigma_{X^c}$=2.78 &$\sigma_{X^w}$=0.59& $\sigma_{Y^c}$=1.57 &  $\sigma_{Y^w}$=0.29\\\hline
\end{tabular}
\end{Huge}}
\end{subtable}%
\begin{subtable}{.24\linewidth}
\centering
\caption{Set-8}
\resizebox{1\columnwidth}{!}{
\begin{Huge}
\begin{tabular}{c|cc|cc} \hline
Entry&$X^-$	&$X^+$&$Y^-$&$Y^+$ \\\hline
1& 5.3&	6.5&	6.2&	8.2\\
2& 5.5&	7.1&	5.2&	7.4\\
3& 4.9&	6.7&	6.5&	8.4\\
4& 5.1&	7.9&	5.7&	7.6\\
5& 6.2&	8.0&	5.9&	8.5\\\hline
& $\mu_{X^c}$=6.32 & $\mu_{X^w}$=1.84& $\mu_{Y^c}$=6.96 &$\mu_{Y^w}$=2.12 \\
& $\sigma_{X^c}$=0.52 &$\sigma_{X^w}$=0.59& $\sigma_{Y^c}$=0.47 &  $\sigma_{Y^w}$=0.29\\\hline
\end{tabular}
\end{Huge}}
\end{subtable}
\\
\begin{subtable}{.24\linewidth}
\centering
\caption{Set-9}
\resizebox{1\columnwidth}{!}{
\begin{Huge}
\begin{tabular}{c|ll|ll} \hline
Entry&$X^-$	&$X^+$&$Y^-$&$Y^+$ \\\hline
1& 2.8&	5.4&	6.4&	9.2\\
2& 3.9&	5.1&	4.4&	5.8\\
3& 1.1&	3.2&	2.4&	4.6\\
4& 5.3&	5.6&	1.8&	2.3\\
5& 7.7&	9.3&	3.2&	5.0\\\hline
& $\mu_{X^c}$=4.94 & $\mu_{X^w}$=1.56 & $\mu_{Y^c}$=4.51 & $\mu_{Y^w}$=1.74 \\
& $\sigma_{X^c}$=2.32 & $\sigma_{X^w}$=0.88 & $\sigma_{Y^c}$=2.15&$\sigma_{Y^w}$=0.86\\\hline
\end{tabular}
\end{Huge}}
\end{subtable}
\begin{subtable}{.24\linewidth}
\centering
\caption{Set-10}
\resizebox{1\columnwidth}{!}{
\begin{Huge}
\begin{tabular}{c|ll|ll} \hline
Entry&$X^-$	&$X^+$&$Y^-$&$Y^+$ \\\hline
1& 4.8&	6.4&	5.4&	7.2\\
2& 4.8&	5.1&	4.0&	4.5\\
3& 4.2&	5.4&	4.4&	5.8\\
4& 5.0&	7.6&	3.6&	6.4\\
5& 6.1&	8.2&	4.8&	7.0\\\hline
& $\mu_{X^c}$=5.76 & $\mu_{X^w}$=1.56& $\mu_{Y^c}$=5.31 &$\mu_{Y^w}$=1.74 \\
& $\sigma_{X^c}$=0.98 &$\sigma_{X^w}$=0.88& $\sigma_{Y^c}$=0.80 &  $\sigma_{Y^w}$=0.86\\\hline
\end{tabular}
\end{Huge}}
\end{subtable}%
\begin{subtable}{.24\linewidth}
\centering
\caption{Set-11}
\resizebox{1\columnwidth}{!}{
\begin{Huge}
\begin{tabular}{c|ll|ll} \hline
Entry&$X^-$	&$X^+$&$Y^-$&$Y^+$ \\\hline
1& 6.1&	6.4&	4.7&	5.2\\
2& 5.6&	6.8&	4.3&	5.7\\
3& 5.4&	7.0&	4.1&	5.9\\
4& 5.2&	7.3&	3.9&	6.1\\
5& 5.0&	7.6&	3.6&	6.4\\\hline
& $\mu_{X^c}$=6.24 & $\mu_{X^w}$=1.56& $\mu_{Y^c}$=4.99 &$\mu_{Y^w}$=1.74 \\
& $\sigma_{X^c}$=0.04 & $\sigma_{X^w}$=0.88 & $\sigma_{Y^c}$=0.02 &  $\sigma_{Y^w}$=0.86\\\hline
\end{tabular}
\end{Huge}}
\end{subtable}%
\label{tab:data set2}
\end{table*}

\begin{figure*}[b!]
\centering 
\noindent\par{Example-1: Relatively low fit of the PM approach when the Box-Cox transformation is applied. }\vspace{0.5mm}
\begin{subfigure}{.24\textwidth}
\centering
\resizebox{1\columnwidth}{!}{
\begin{Huge}
\begin{tabular}{c|ll|ll} \hline
Entry&$X^-$	&$X^+$&$Y^-$&$Y^+$ \\\hline
1& 0.5 &	0.6	&5.5&	5.8\\
2& 0.2&	0.5	&0.8&	10\\
3& 0.8&	1.0	&4.8&	6.8\\
4& 0.6&	0.9&	2.2	&10\\
5& 2.0&	2.3	&1.8&	9.8\\\hline
& $\mu_{X^c}$=0.94 & $\mu_{X^w}$=0.24& $\mu_{Y^c}$=5.75 &$\mu_{Y^w}$=5.46 \\
& $\sigma_{X^c}$=0.71 &$\sigma_{X^w}$=0.09& $\sigma_{Y^c}$=0.25 & $\sigma_{Y^w}$=4.02\\\hline
\end{tabular}
\end{Huge}}
\caption{Set-12}
\end{subfigure} 
\begin{subfigure}{0.24\textwidth}
\centering
\includegraphics[width=0.7\columnwidth]{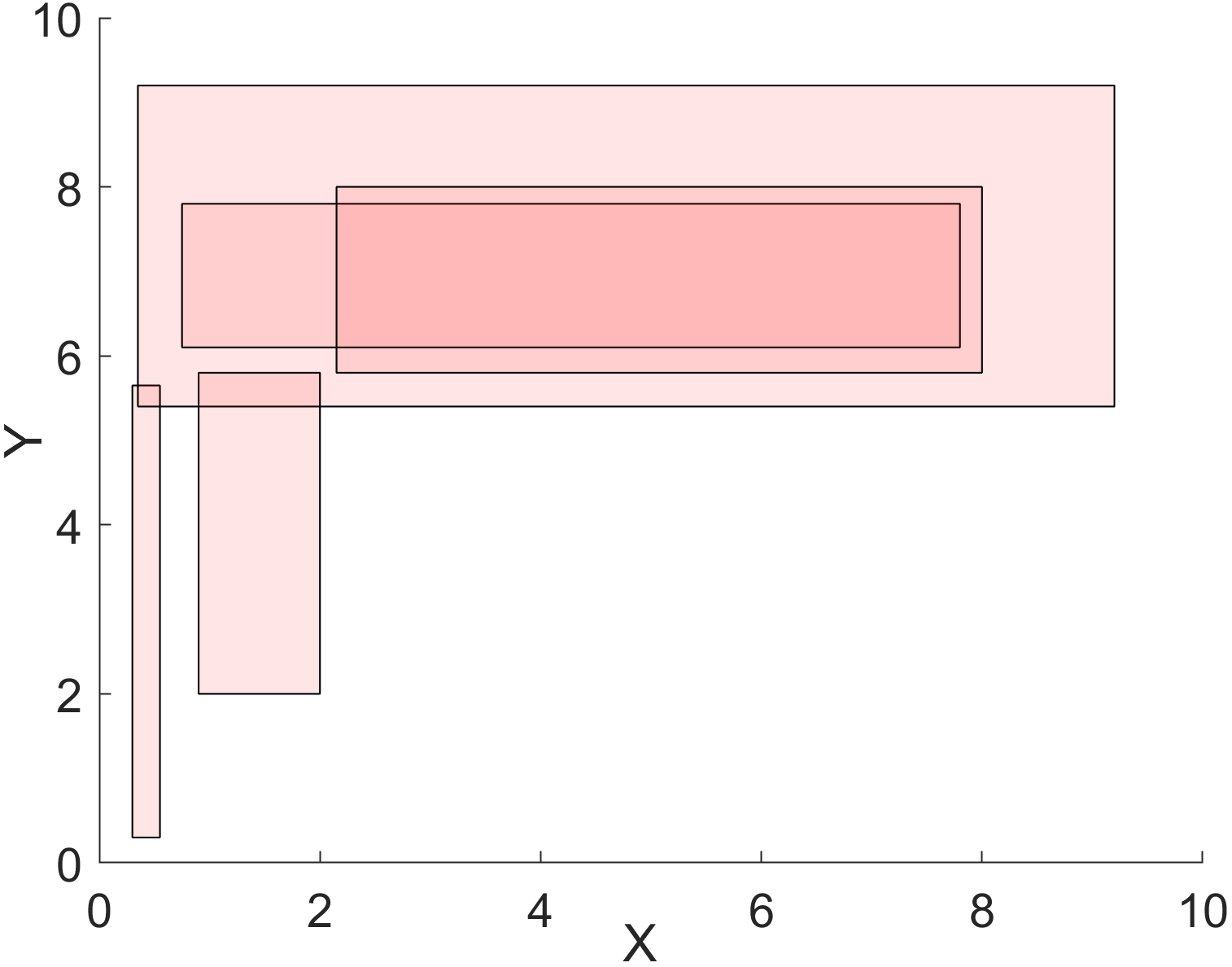}
\caption{2D visualization}
\end{subfigure} 
\begin{subfigure}{0.24\textwidth}
\centering
\resizebox{0.9\hsize}{!}{
\begin{Huge}
\begin{tabular}{ccccccc} \hline
Methods & $RMSE^-$ & $RMSE^+$ & $MAE^-$ & $MAE^+$ & $MMRE$\\\hline
CM &3.271	&3.251	&2.716	&2.716	&1.159\\
MinMax &1.797	&1.714	&1.654	&1.558	&0.573\\
CRM &0.645	&0.441	&0.567	&0.372	&0.187\\
CCRM &0.995	&0.899	&0.859	&0.783	&0.280\\
CIM &3.271	&3.251	&2.716	&2.716	&1.159\\
\textcolor{blue}{PM}  &\textcolor{blue}{2.636}	&\textcolor{blue}{7.594}	&\textcolor{blue}{2.037}	&\textcolor{blue}{7.375}	&\textcolor{blue}{0.715}\\
LM$_c$ &0.646	&1.603	&0.565	&1.380	&0.261\\
LM$_w$ &0.646	&1.603	&0.565	&1.380	&0.261\\\hline
\end{tabular}
\end{Huge}}
\caption{Model Performance}
\end{subfigure}
\caption{Comparatively low fit of the PM approach.}
\vspace{-1.0em}
\label{fig:box_cox_PM}
\end{figure*}

Beyond advancing techniques, in the future, we aim to explore additional real-world cases to demonstrate and advance the understanding of the value of using intervals, rather than `crisp', numeric data. Specifically, we will revisit the nature of the information captured by intervals, whether disjunctive (such as in confidence intervals), or conjunctive (true ranges, such as the real numbers between 2 and 6) and explore how different analysis techniques can, and whether they should, deal with such data differently.  \vspace{-0.8em}

\section*{Acknowledgment}
This work was supported by the UK EPSRC’s Leveraging the Multi-Stakeholder Nature of Cyber Security EP/P011918/1 and Horizon: Trusted Data-Driven Products EP/T022493/1 grants.\vspace{-0.8em}

\bibliographystyle{IEEEtran}
\balance
\bibliography{IEEEabrv,reference}

\section*{\begin{large}Appendix\end{large}}
\begin{figure*}[t!]
\begin{subfigure}{0.24\textwidth}
\centering
\includegraphics[width=1\textwidth]{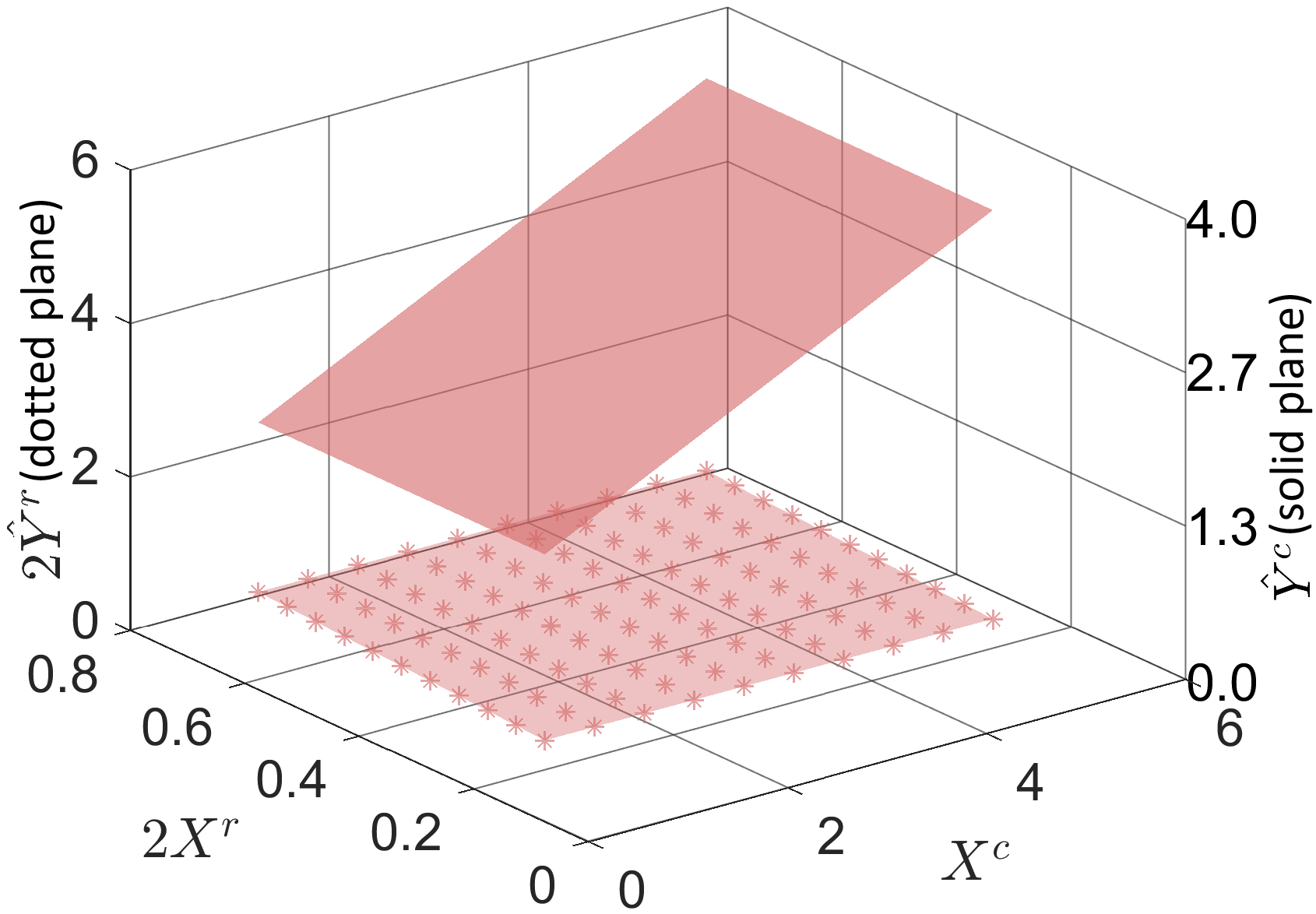}
\caption{\footnotesize CM Method}
\end{subfigure}
\begin{subfigure}{0.24\textwidth}
\centering
\includegraphics[width=1\textwidth]{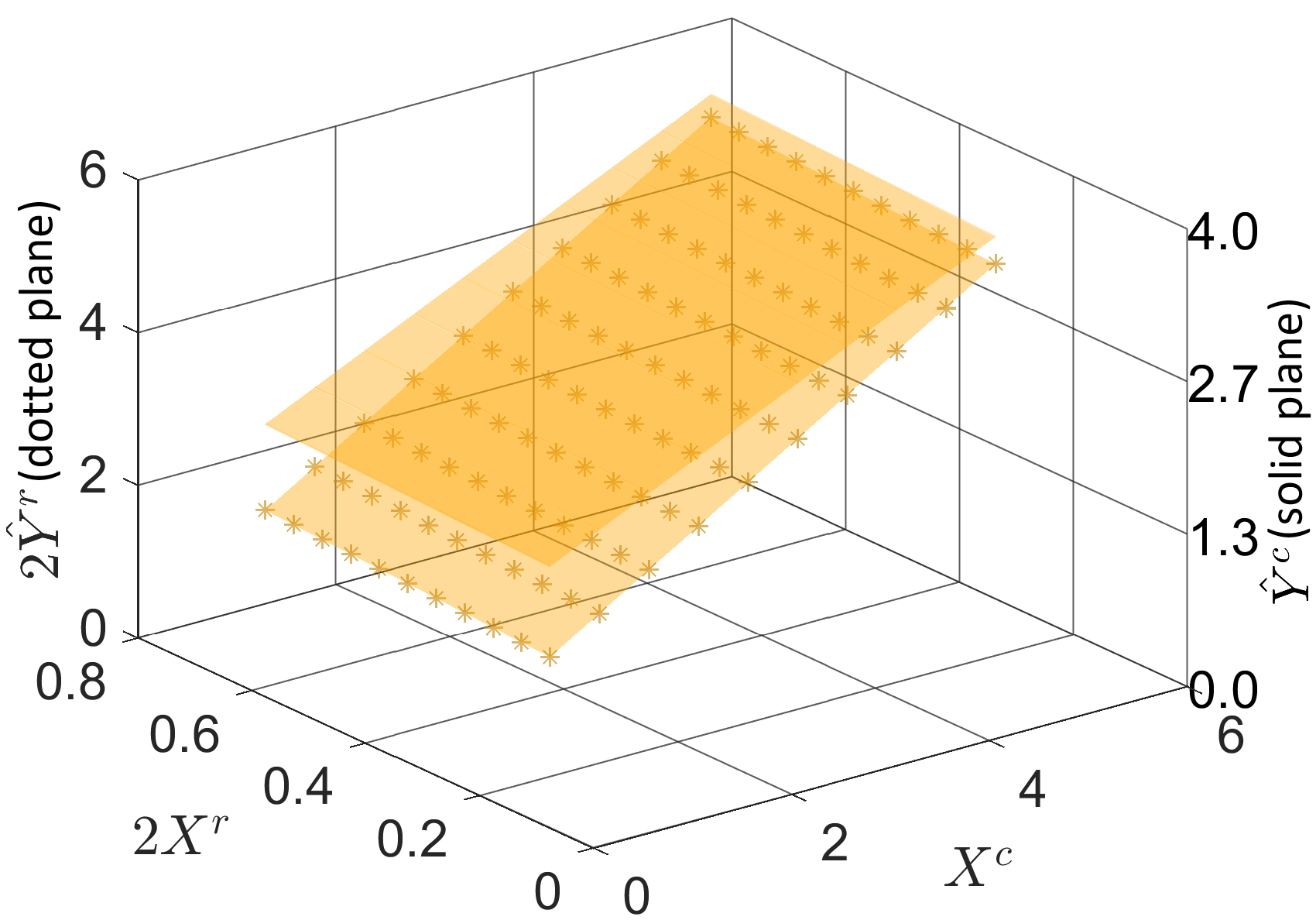}
\caption{\footnotesize MinMax Method}
\end{subfigure}
\begin{subfigure}{0.24\textwidth}
\centering
\includegraphics[width=1\textwidth]{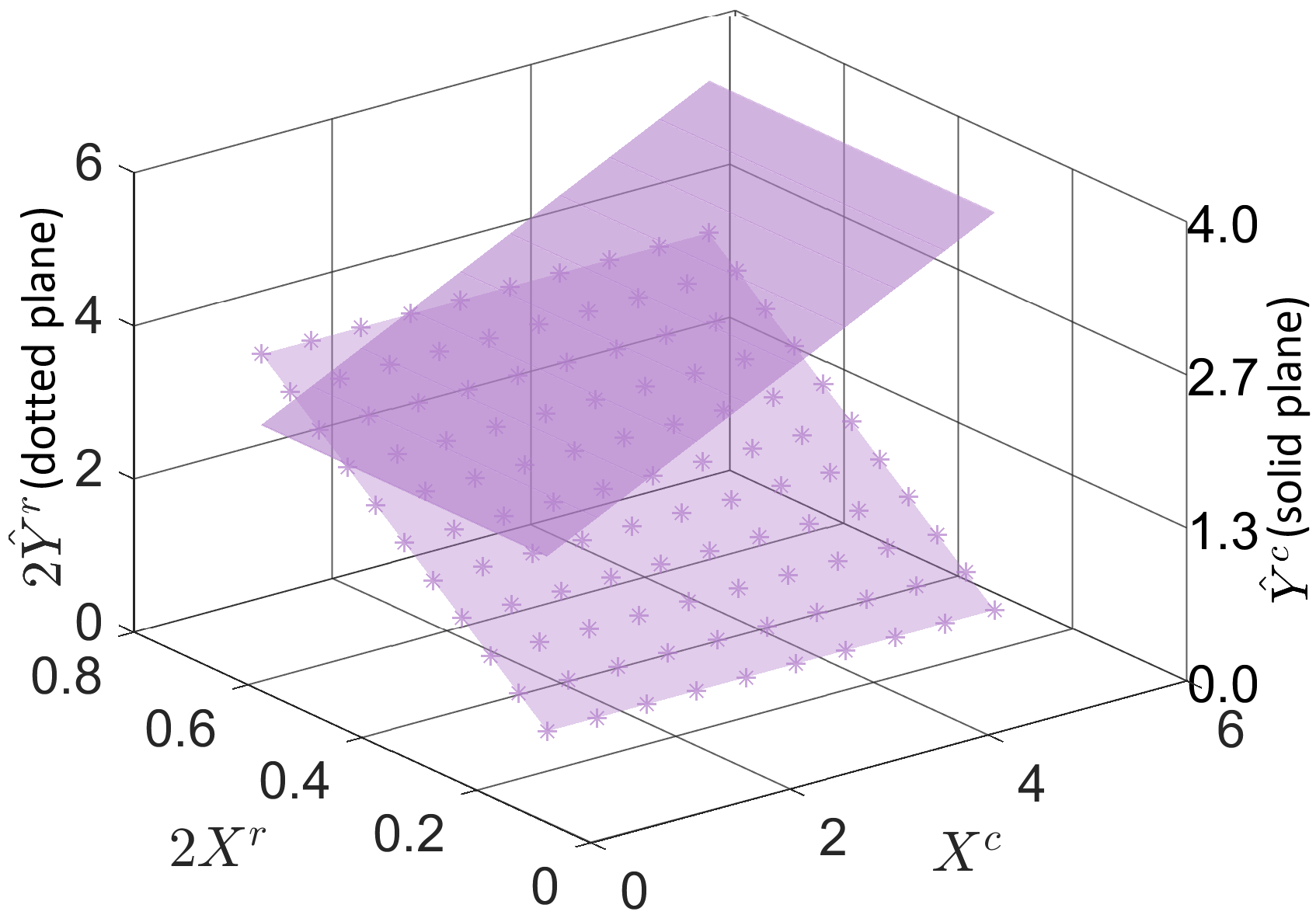}
\caption{\footnotesize CRM Method}
\end{subfigure}
\begin{subfigure}{0.24\textwidth}
\centering
\includegraphics[width=1\textwidth]{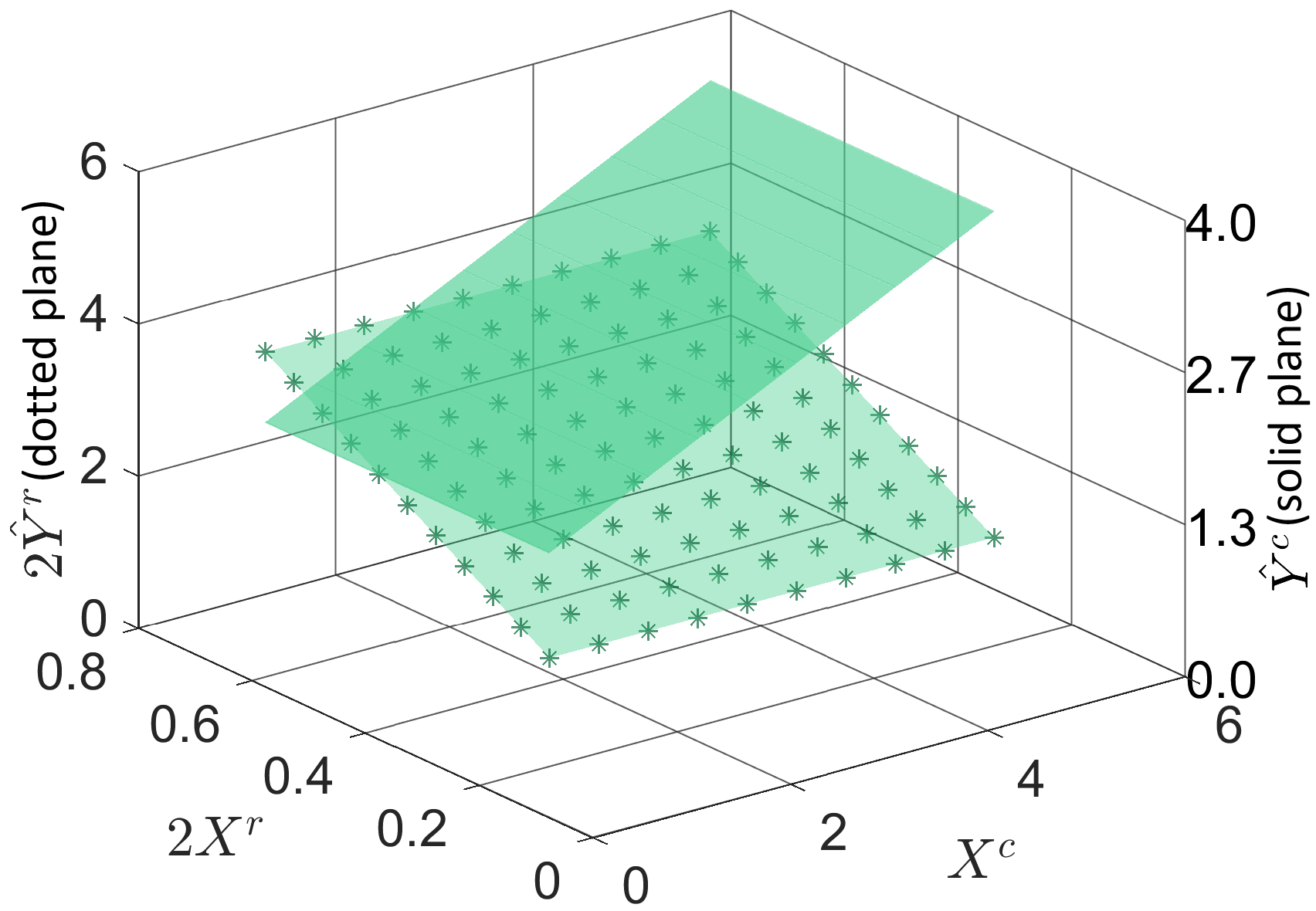}
\caption{\footnotesize CCRM Method}
\end{subfigure}\vspace{0.2cm}\\
\begin{subfigure}{0.24\textwidth}
\centering
\includegraphics[width=1\textwidth]{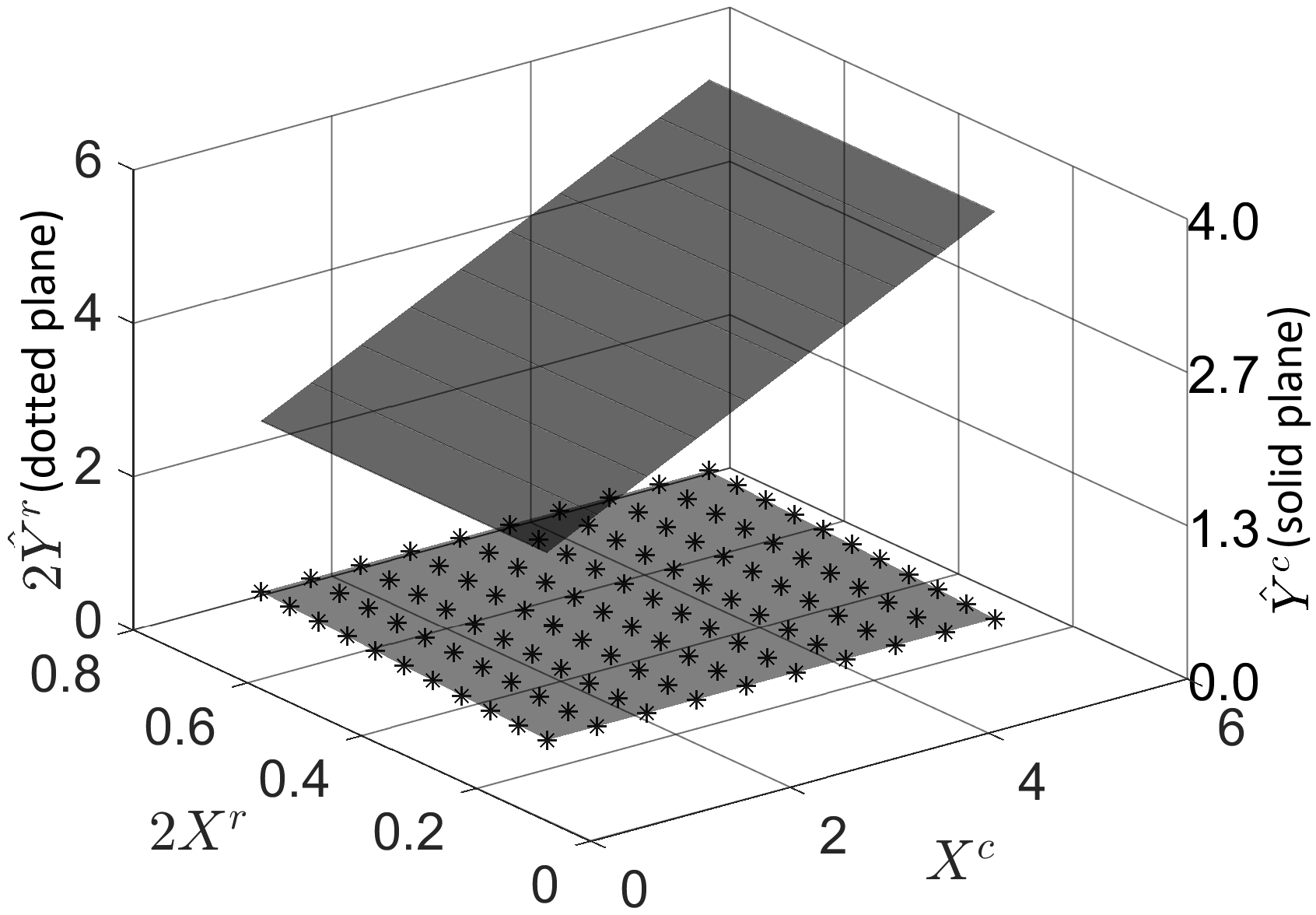}
\caption{\footnotesize CIM Method}
\end{subfigure}
\begin{subfigure}{0.24\textwidth}
\centering
\includegraphics[width=1\textwidth]{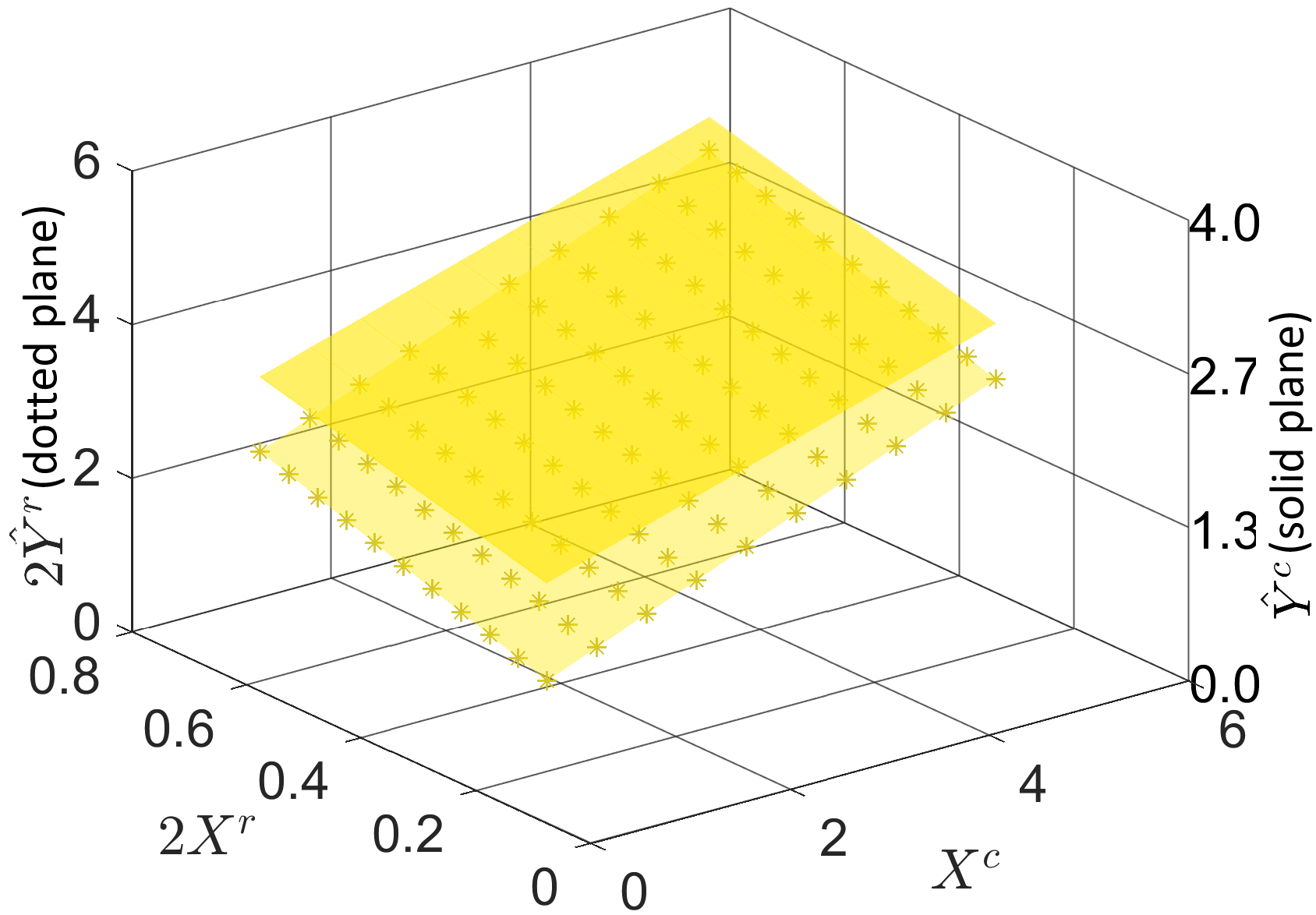}
\caption{\footnotesize PM Method}
\end{subfigure}
\begin{subfigure}{0.24\textwidth}
\centering
\includegraphics[width=1\textwidth]{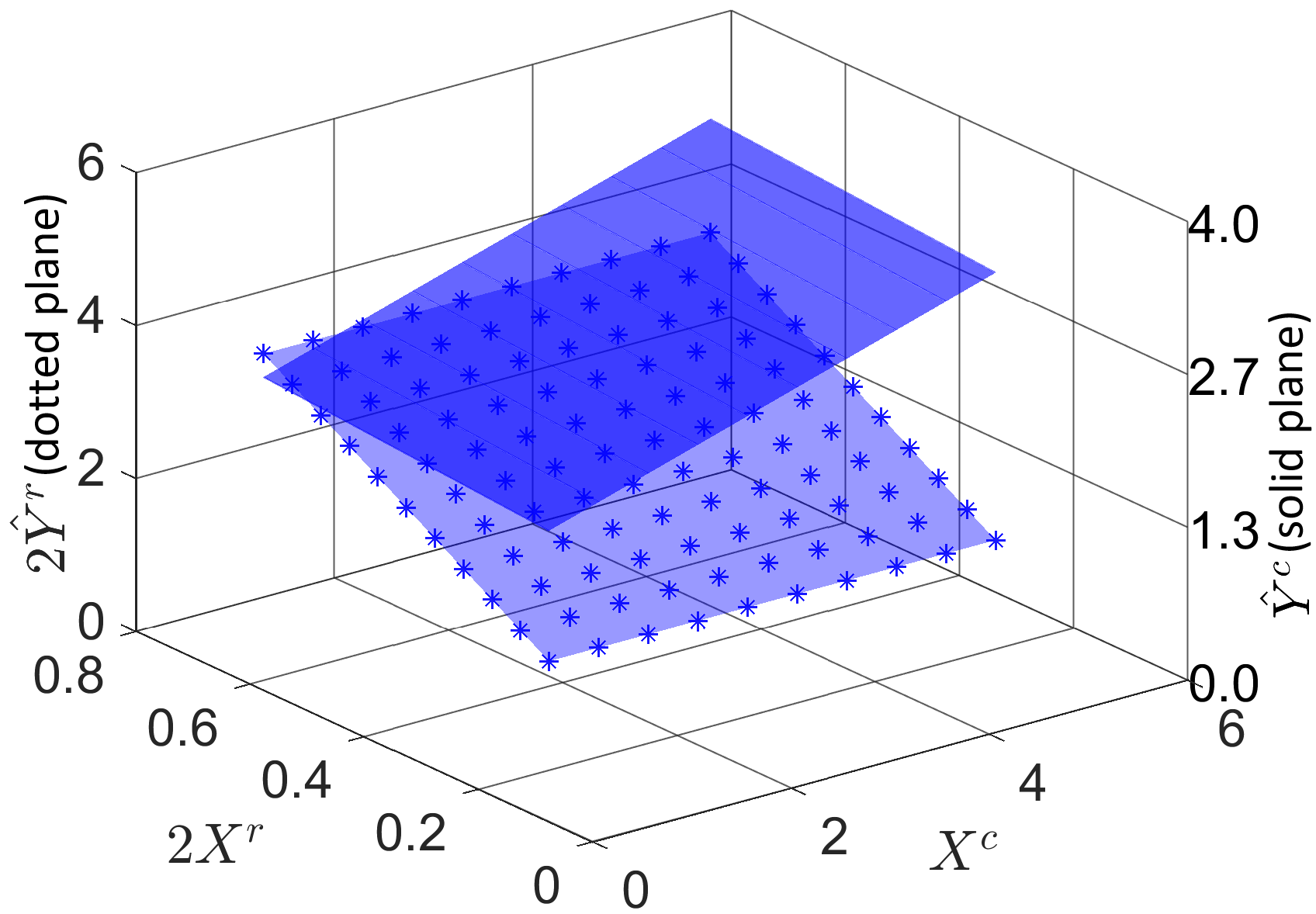}
\caption{\footnotesize LM$_c$ Method}
\end{subfigure}
\begin{subfigure}{0.24\textwidth}
\centering
\includegraphics[width=1\textwidth]{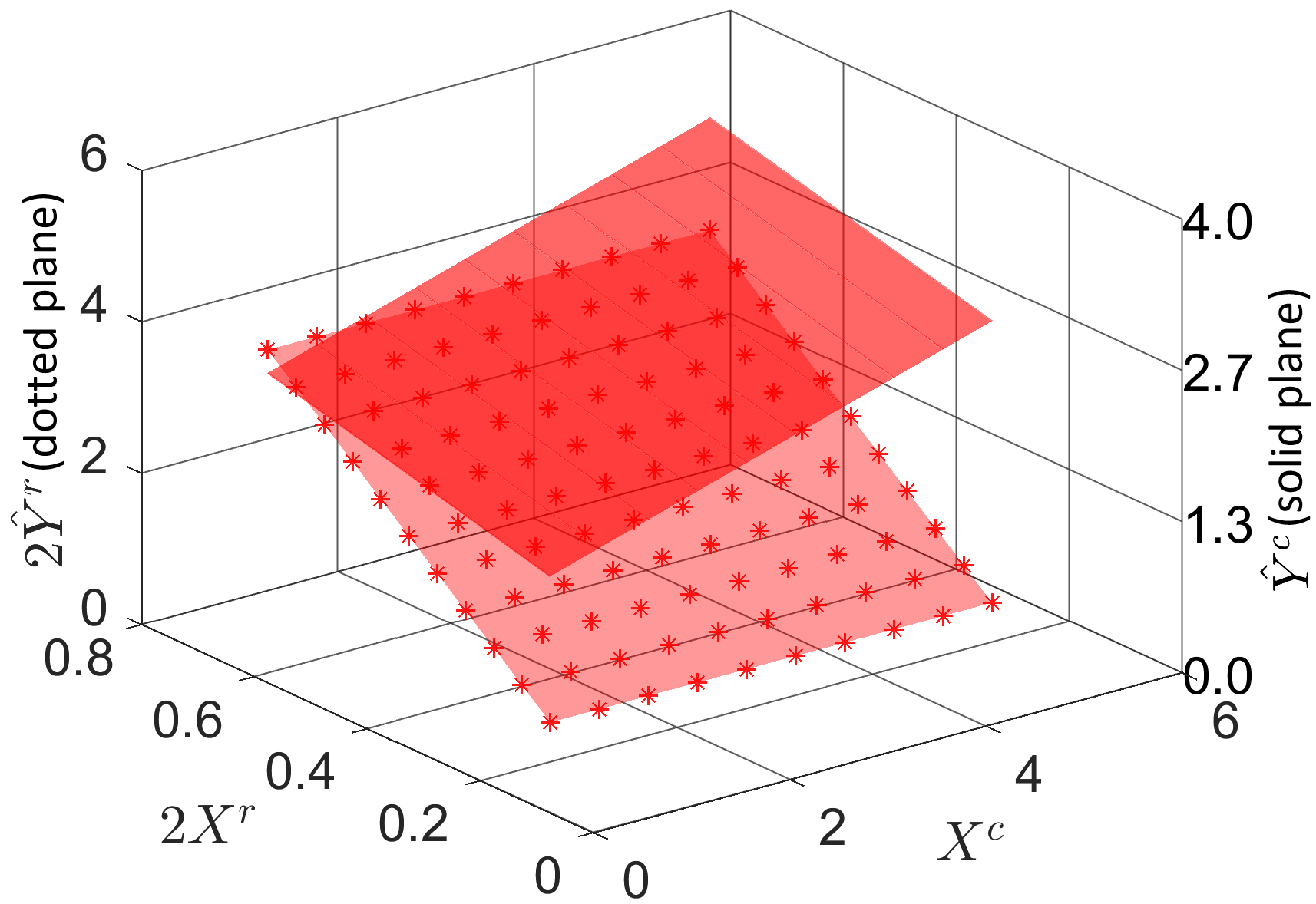}
\caption{\footnotesize LM$_w$ Method}
\end{subfigure}
\caption{\emph{IRG}s showing the relationship between regressor and regressand in terms of their center and range using different linear regression models for Set-1 (Fig.~\ref{fig1}(a)).}
\vspace{-0.8em}
\label{fig:case-1}
\end{figure*}
\begin{figure*}[htbp]
\begin{subfigure}{0.24\textwidth}
\centering
\includegraphics[width=1\textwidth]{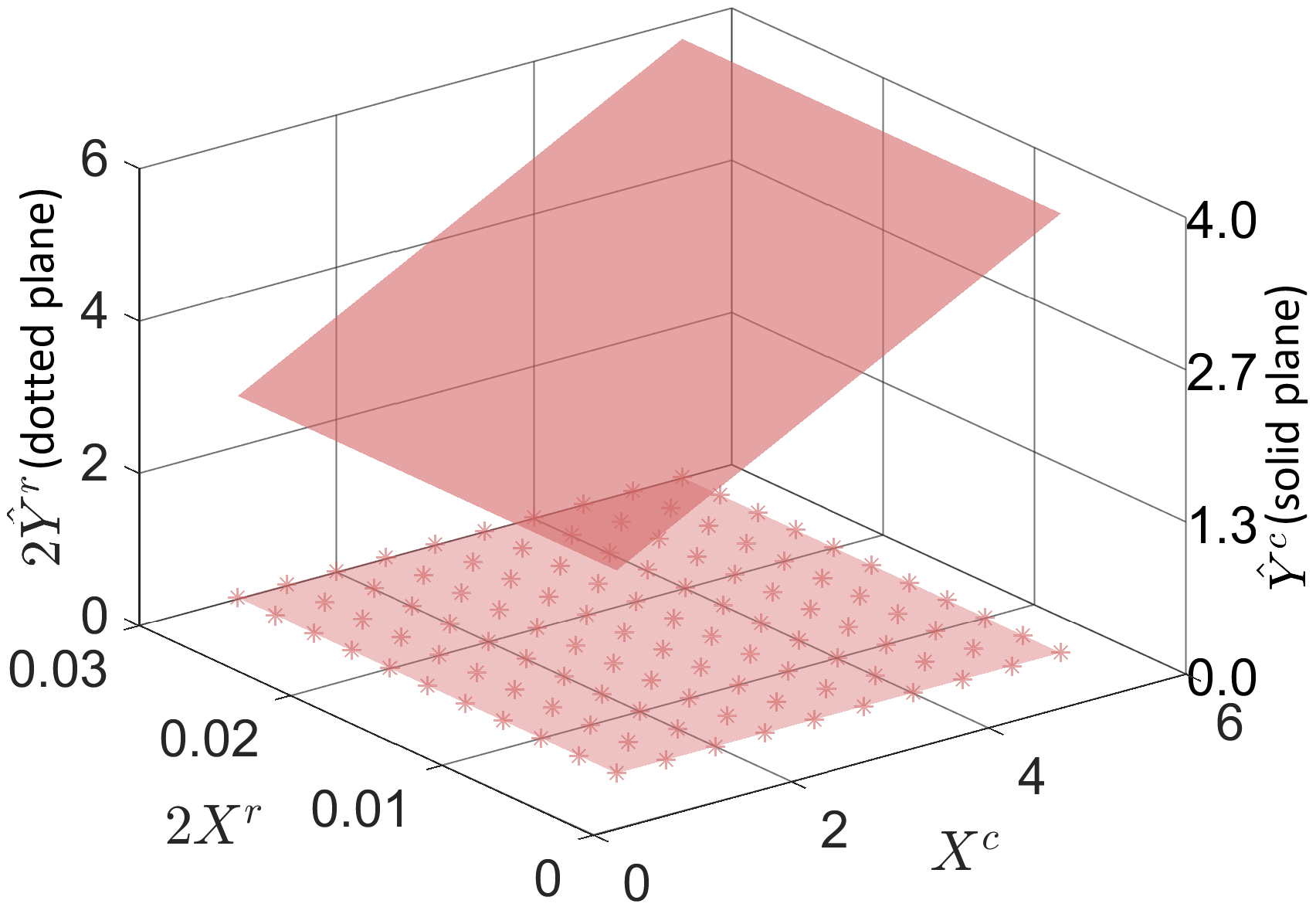}
\caption{\footnotesize CM Method}
\end{subfigure}
\begin{subfigure}{0.24\textwidth}
\centering
\includegraphics[width=1\textwidth]{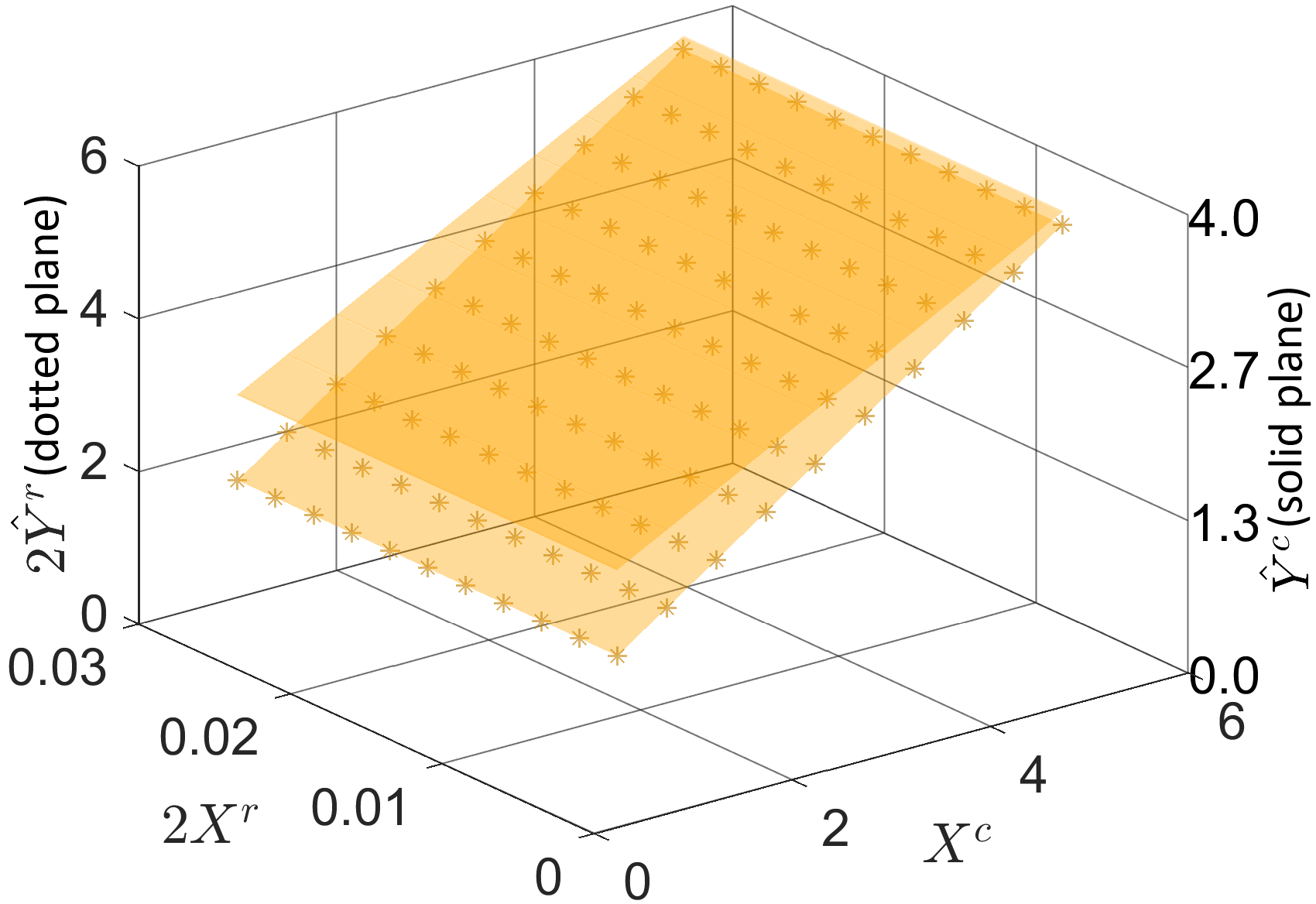}
\caption{\footnotesize MinMax Method}
\end{subfigure}
\begin{subfigure}{0.24\textwidth}
\centering
\includegraphics[width=1\textwidth]{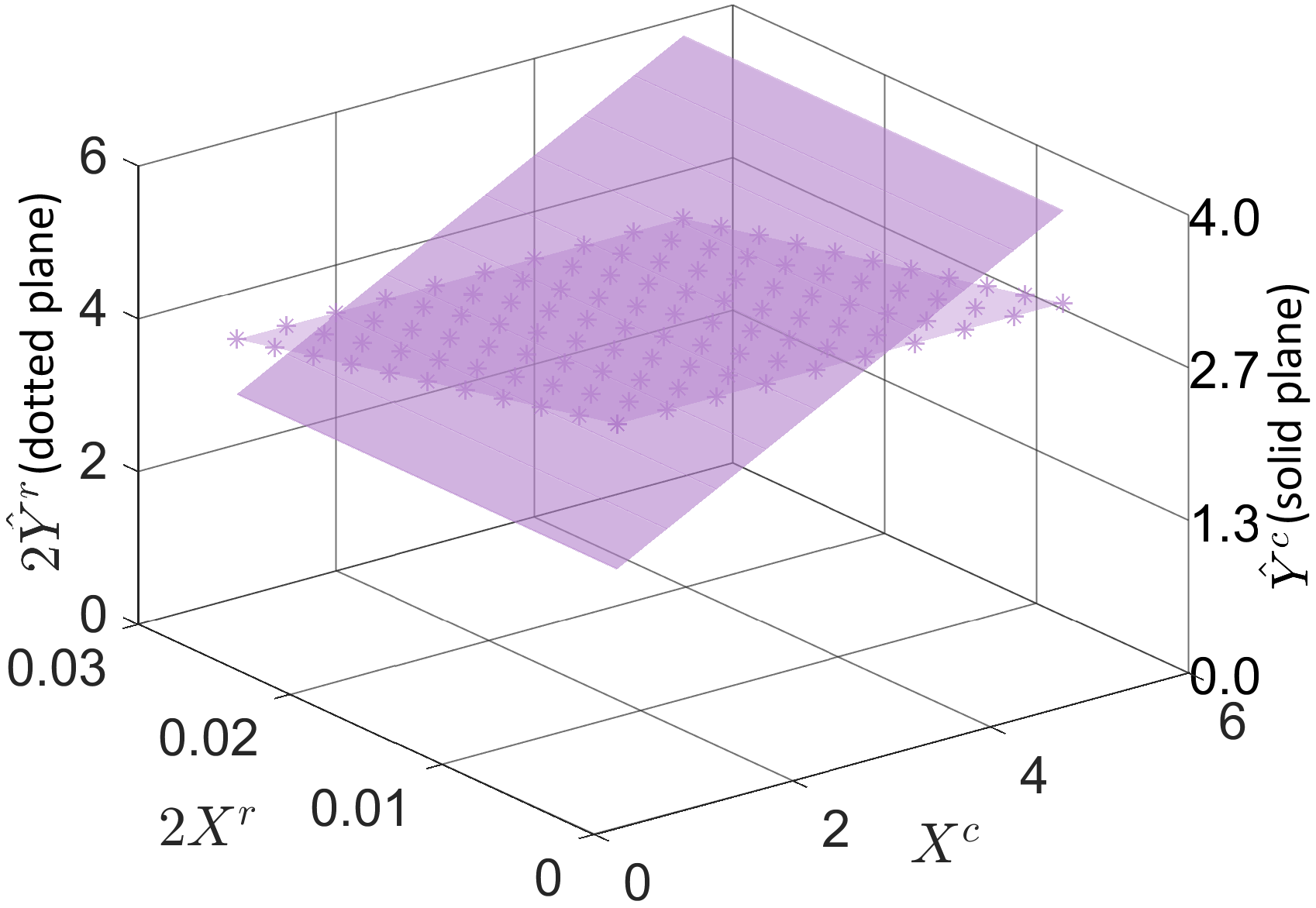}
\caption{\footnotesize CRM Method}
\end{subfigure}
\begin{subfigure}{0.24\textwidth}
\centering
\includegraphics[width=1\textwidth]{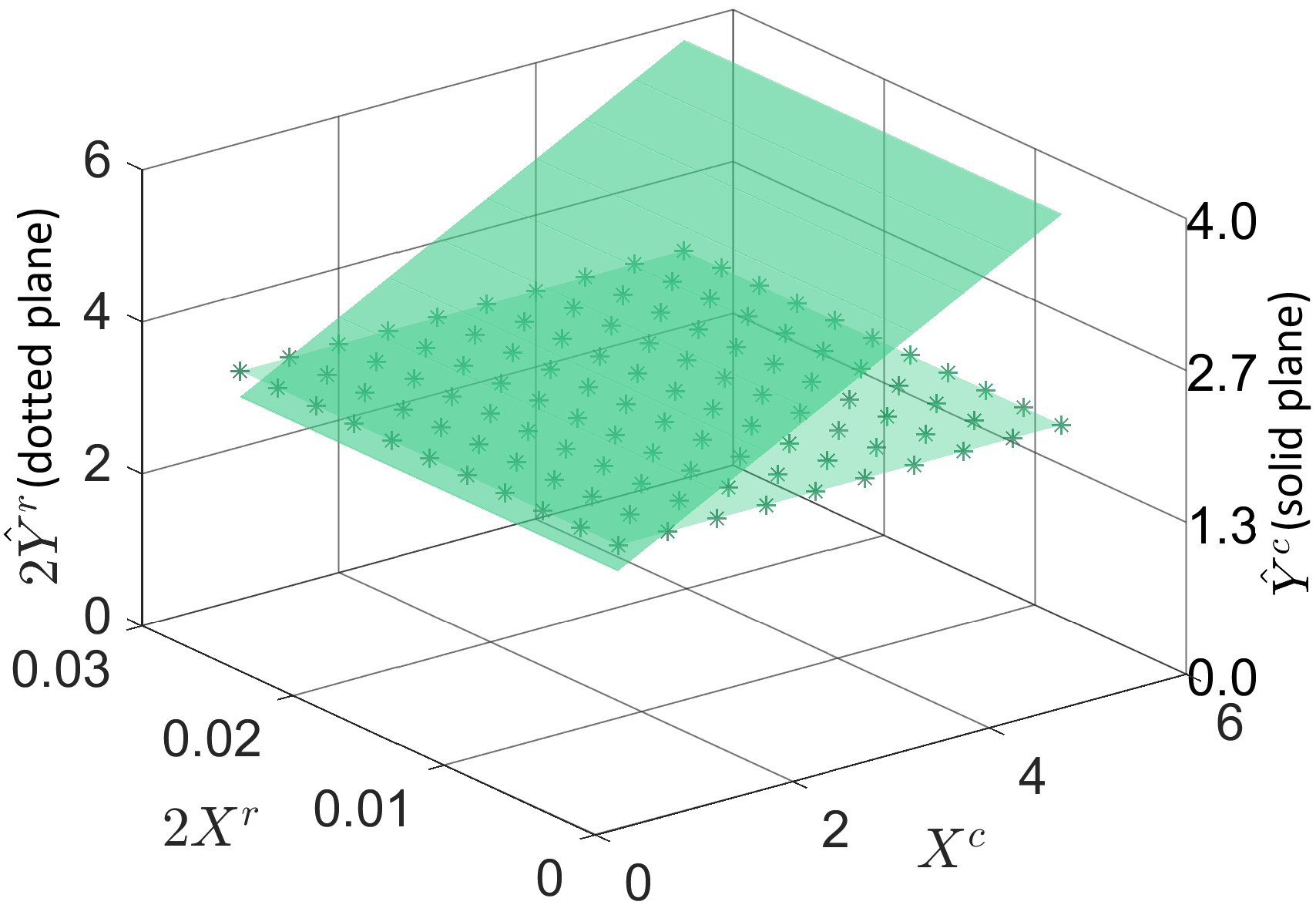}
\caption{\footnotesize CCRM Method}
\end{subfigure}\vspace{0.2cm}\\
\begin{subfigure}{0.24\textwidth}
\centering
\includegraphics[width=1\textwidth]{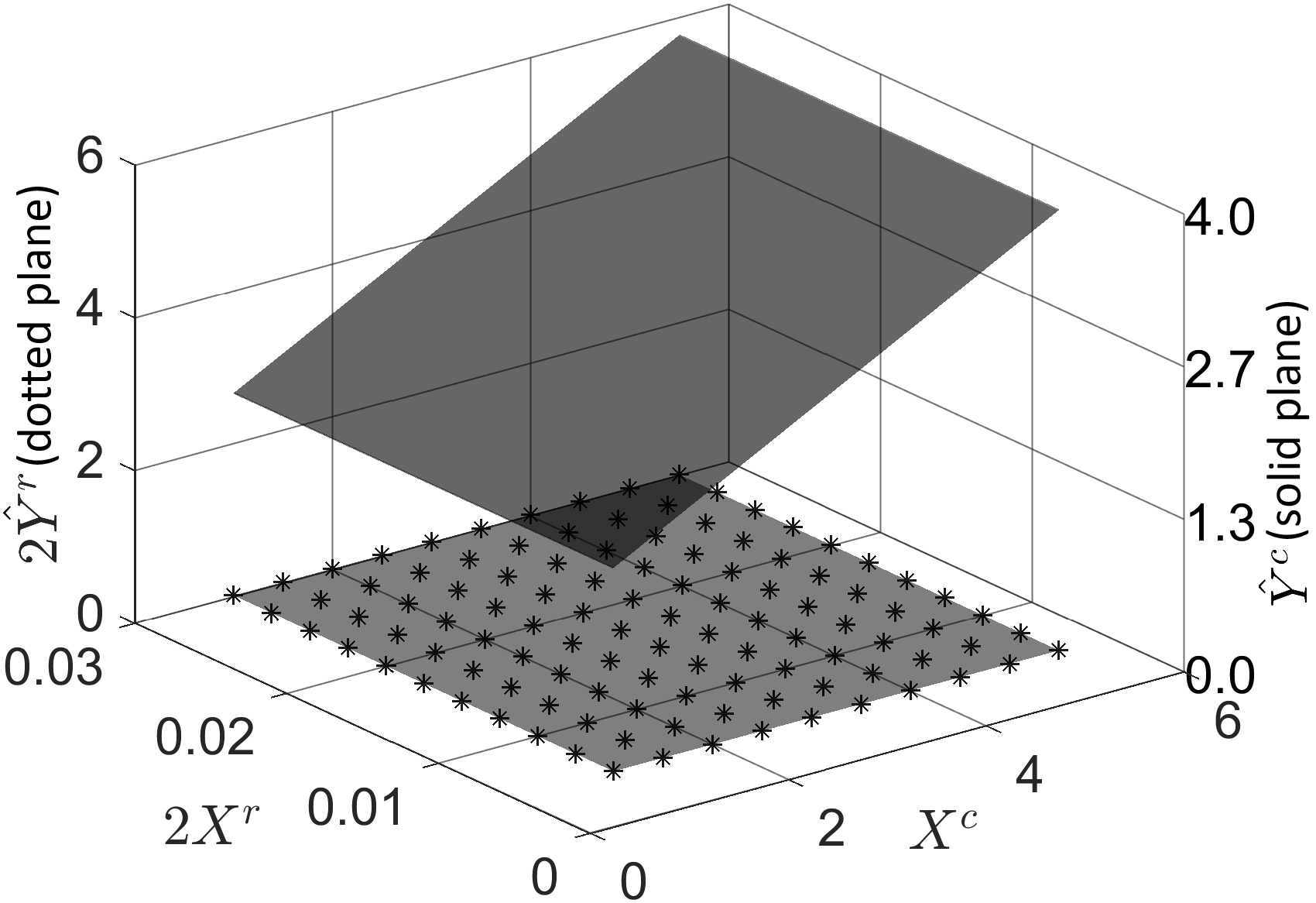}
\caption{\footnotesize CIM Method}
\end{subfigure}
\begin{subfigure}{0.24\textwidth}
\centering
\includegraphics[width=1\textwidth]{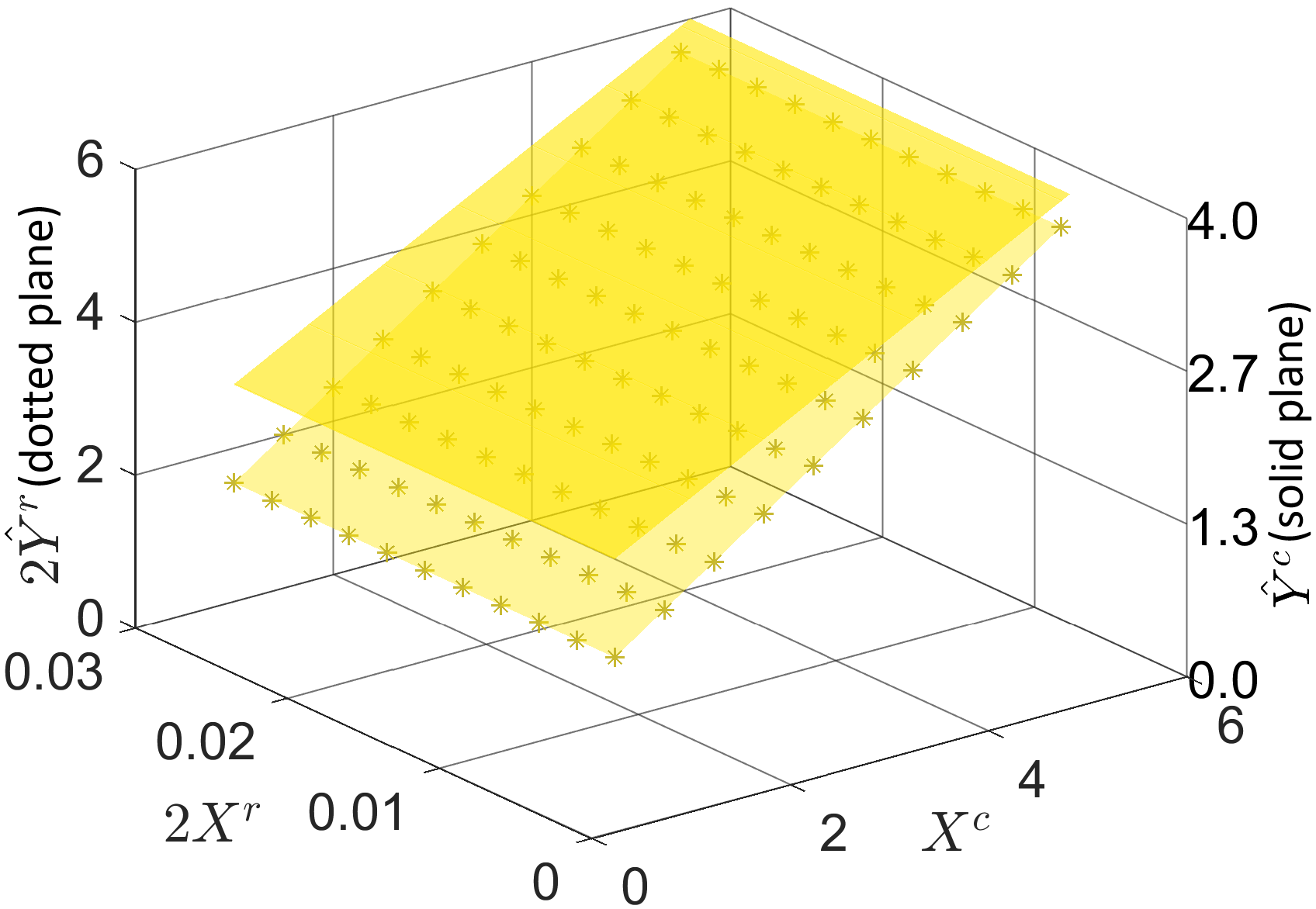}
\caption{\footnotesize PM Method}
\end{subfigure}
\begin{subfigure}{0.24\textwidth}
\centering
\includegraphics[width=1\textwidth]{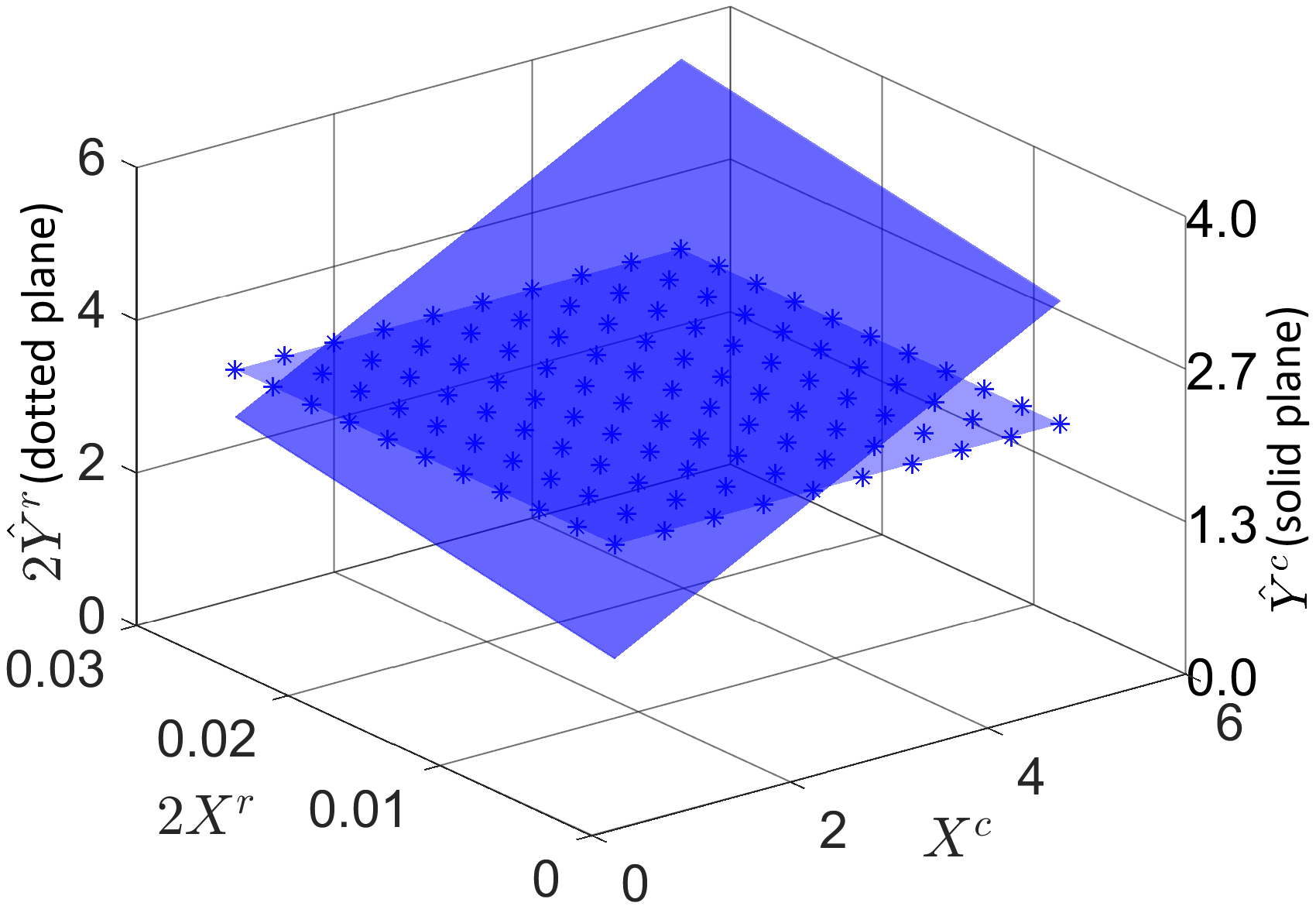}
\caption{\footnotesize LM$_c$ Method}
\end{subfigure}
\begin{subfigure}{0.24\textwidth}
\centering
\includegraphics[width=1\textwidth]{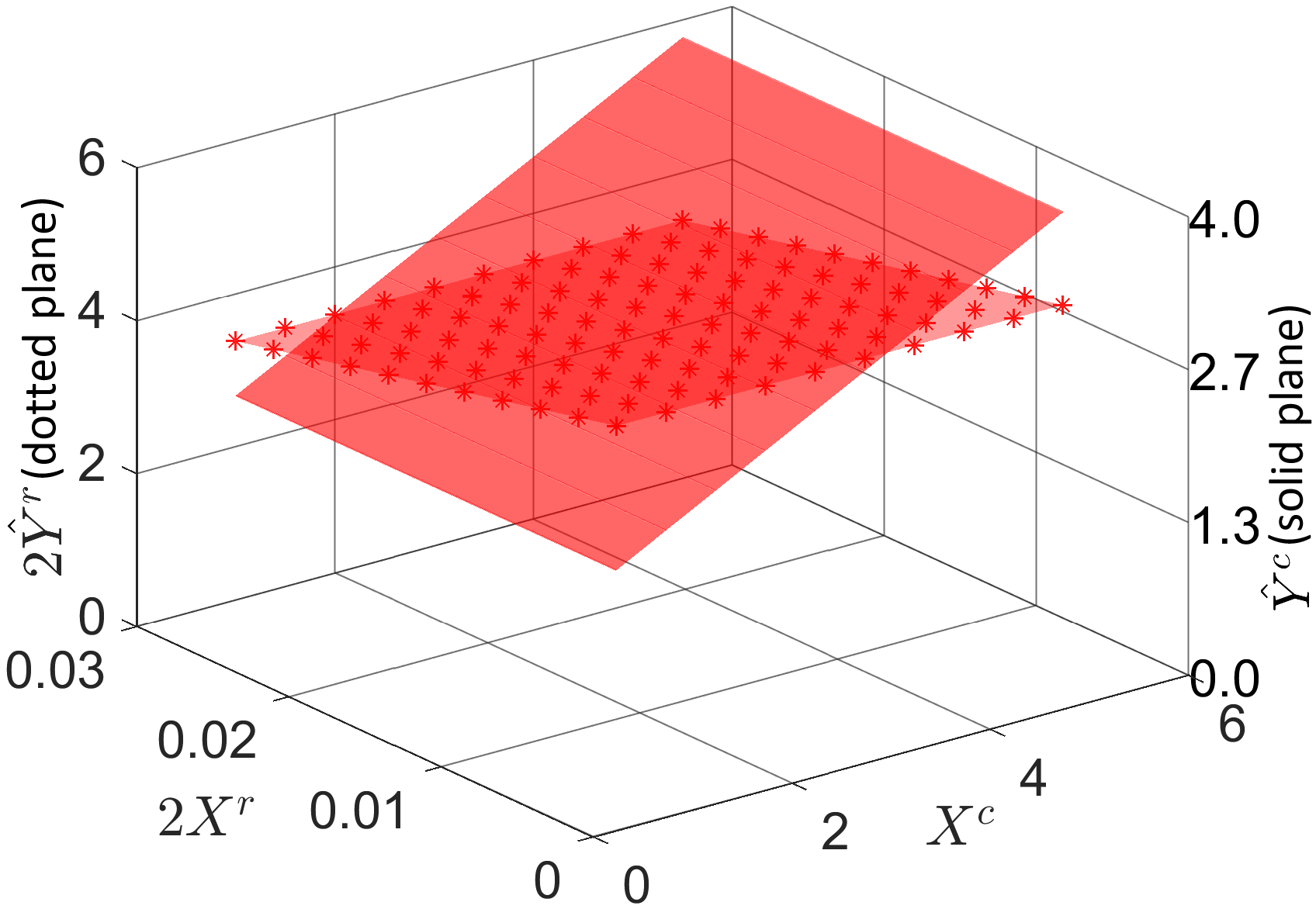}
\caption{\footnotesize LM$_w$ Method}
\end{subfigure}
\caption{\emph{IRG}s showing the relationship between regressor and regressand in terms of their center and range using different linear regression models for Set-2 (Fig.~\ref{fig11}(a)).}
\label{fig:case-2}
\end{figure*}
\begin{table*}
\centering
\caption{Estimated coefficients of different regression methods for synthetic data sets, Set-1 to set-11}
    \begin{subtable}{1\textwidth}
    \resizebox{1\textwidth}{!}{      
    \begin{Huge}
     \begin{tabular}{c|c|c|c|c|c|c}\hline
&   \multicolumn{6}{c}{Synthetic data sets} \\\cline{2-7}
Methods  & Set-1 & Set-2 & Set-3 & Set-4& Set-5 & Set-6 \\\hline
CM & $\hat{\beta}^c_0=1.256,\hat{\beta}^c_1=0.43$   & $\hat{\beta}^c_0=1.316,\hat{\beta}^c_1=0.46$ & $\hat{\beta}^c_0=4.608,\hat{\beta}^c_1=0.261$& $\hat{\beta}^c_0=5.637,\hat{\beta}^c_1=0.172$ & $\hat{\beta}^c_0=10.54,\hat{\beta}^c_1=-0.691$ & $\hat{\beta}^c_0=5.866,\hat{\beta}^c_1=-0.017$   \\\hline

MinMax & $\hat{\beta}^-_0=0.997,\hat{\beta}^-_1=0.009$& $\hat{\beta}^-_0=0.997,\hat{\beta}^-_1=0.009$ & $\hat{\beta}^-_0=4.56,\hat{\beta}^-_1=0.24$ & $\hat{\beta}^-_0=5.505,\hat{\beta}^-_1=0.164$ & $\hat{\beta}^-_0=11.35,\hat{\beta}^-_1=-0.90$ & $\hat{\beta}^-_0=4.778,\hat{\beta}^-_1=-0.015$   \\

        & $\hat{\beta}^+_0=1.393,\hat{\beta}^+_1=0.801$   &
        $\hat{\beta}^+_0=1.611,\hat{\beta}^+_1=0.913$ & 
        $\hat{\beta}^+_0=4.65,\hat{\beta}^+_1=0.281$ & $\hat{\beta}^+_0=5.766,\hat{\beta}^+_1=0.18$ & $\hat{\beta}^+_0=9.433,\hat{\beta}^+_1=-0.444$ & $\hat{\beta}^+_0=6.967,\hat{\beta}^+_1=-0.022$  \\\hline

CRM & $\hat{\beta}^c_0=1.256,\hat{\beta}^c_1=0.43$   & $\hat{\beta}^c_0=1.316,\hat{\beta}^c_1=0.46$ & $\hat{\beta}^c_0=4.608,\hat{\beta}^c_1=0.261$& $\hat{\beta}^c_0=5.636,\hat{\beta}^c_1=0.172$ & $\hat{\beta}^c_0=10.54,\hat{\beta}^c_1=-0.691$ & $\hat{\beta}^c_0=5.867,\hat{\beta}^c_1=-0.017$ \\
        &$\hat{\beta}^r_0=-0.678,\hat{\beta}^r_1=6.407$ & $\hat{\beta}^r_0=2.398,\hat{\beta}^r_1=-47.297$ & $\hat{\beta}^r_0=0.264,\hat{\beta}^r_1=-0.286$ & $\hat{\beta}^r_0=0.264,\hat{\beta}^r_1=-0.286$ &
        $\hat{\beta}^r_0=0.264,\hat{\beta}^r_1=-0.286$ & $\hat{\beta}^r_0=1.168,\hat{\beta}^r_1=-0.117$   \\\hline

CCRM &  $\hat{\beta}^c_0=1.256,\hat{\beta}^c_1=0.43$  & $\hat{\beta}^c_0=1.316,\hat{\beta}^c_1=0.46$ & $\hat{\beta}^c_0=4.608,\hat{\beta}^c_1=0.261$& $\hat{\beta}^c_0=5.636,\hat{\beta}^c_1=0.172$ & $\hat{\beta}^c_0=10.54,\hat{\beta}^c_1=-0.691$ & $\hat{\beta}^c_0=5.866,\hat{\beta}^c_1=-0.017$ \\
        &$\hat{\beta}^r_0=0.0,\hat{\beta}^r_1=4.60$ & $\hat{\beta}^r_0=1.5,\hat{\beta}^r_1=0.0$ & $\hat{\beta}^r_0=0.21,\hat{\beta}^r_1=0.0$ & $\hat{\beta}^r_0=0.21,\hat{\beta}^r_1=0.0$ &
        $\hat{\beta}^r_0=0.21,\hat{\beta}^r_1=0.0$ & $\hat{\beta}^r_0=1.06,\hat{\beta}^r_1=0.0$      \\\hline

CIM  & $\hat{\beta}_0=1.275,\hat{\beta}_1=0.423$   & $\hat{\beta}_0=1.316,\hat{\beta}_1=0.46$ & $\hat{\beta}_0=4.615,\hat{\beta}_1=0.26$ & $\hat{\beta}_0=5.639,\hat{\beta}_1=0.172$ & $\hat{\beta}_0=9.0,\hat{\beta}_1=-0.42$ & $\hat{\beta}_0=5.861,\hat{\beta}_1=-0.016$   \\\hline    

PM & $\hat{\beta}^-_0=0.987,\hat{\alpha}^-_1=-0.023$ &
$\hat{\beta}^-_0=0.997,\hat{\alpha}^-_1=0.008$   & $\hat{\beta}^-_0=-1.683,\hat{\alpha}^-_1=-15.185$ & $\hat{\beta}^-_0=2.295,\hat{\alpha}^-_1=-7.562$& $\hat{\beta}^-_0=11.66,\hat{\alpha}^-_1=-0.60$& $\hat{\beta}^-_0=7.26,\hat{\alpha}^-_1=2.069$      \\
        &$\hat{\omega}^-_1=0.028,\hat{\beta}^+_0=0.737 $ &$\hat{\omega}^-_1=0.001,\hat{\beta}^+_0=1.655$ & $\hat{\omega}^-_1=15.49,\hat{\beta}^+_0=-1.206$ &$\hat{\omega}^-_1=7.772,\hat{\beta}^+_0=2.796$ & $\hat{\omega}^-_1=-0.333,\hat{\beta}^+_0=9.433$ & $\hat{\omega}^-_1=-1.854,\hat{\beta}^+_0=9.58$  \\
        &$\hat{\alpha}^+_1=-2.295,\hat{\omega}^+_1=2.82$ & $\hat{\alpha}^+_1=0.806,\hat{\omega}^+_1=0.101$   & $\hat{\alpha}^+_1=-14.908,\hat{\omega}^+_1=15.222$
        &$\hat{\alpha}^+_1=-7.289,\hat{\omega}^+_1=7.503$ &  $\hat{\alpha}^+_1=0.0,\hat{\omega}^+_1=-0.444$& $\hat{\alpha}^+_1=2.565,\hat{\omega}^+_1=-2.228  $   \\\hline

LM$_c$ &$\hat{\eta}=1.541,\hat{\alpha}_1=2.045$   & $\hat{\eta}=-1.073,\hat{\alpha}_1=-23.241$ & $\hat{\eta}=-1.709,\hat{\alpha}_1=-15.189$&  $\hat{\eta}=2.282,\hat{\alpha}_1=-7.568$&   $\hat{\eta}=10.282,\hat{\alpha}_1=-0.443$ & $\hat{\eta}=7.252,\hat{\alpha}_1=2.258$  \\
        &  $\hat{\beta}_1=-1.779,\hat{\theta}=0.0$ & $\hat{\beta}_1=23.7,\hat{\theta}=3.0$ & $\hat{\beta}_1=15.499,\hat{\theta}=0.42$&  $\hat{\beta}_1=7.78,\hat{\theta}=0.42$&    $\hat{\beta}_1=-0.246,\hat{\theta}=0.42$& $\hat{\beta}_1=-1.982,\hat{\theta}=2.12$  \\
        &$\hat{\gamma}_1=4.60$&  $\hat{\gamma}_1=0.0$ & $\hat{\gamma}_1=0.0$ & $\hat{\gamma}_1=0.0$ & $\hat{\gamma}_1=0.0$& $\hat{\gamma}_1=0.0$\\\hline

LM$_w$ & $\hat{\eta}=1.541,\hat{\alpha}_1=2.045$   & $\hat{\eta}=-1.073,\hat{\alpha}_1=-23.241$ & $\hat{\eta}=-1.709,\hat{\alpha}_1=-15.189$&  $\hat{\eta}=2.282,\hat{\alpha}_1=-7.568$&   $\hat{\eta}=10.282,\hat{\alpha}_1=-0.443$ &$\hat{\eta}=7.252,\hat{\alpha}_1=2.258$   \\
        &  $\hat{\beta}_1=-1.779,\hat{\theta}=-1.357$ & $\hat{\beta}_1=23.7,\hat{\theta}=4.797$ & $\hat{\beta}_1=15.499,\hat{\theta}=0.529$&  $\hat{\beta}_1=7.78,\hat{\theta}=0.529$&   $\hat{\beta}_1=-0.246,\hat{\theta}=0.529$& $\hat{\beta}_1=-1.982,\hat{\theta}=2.337$   \\
        & $\hat{\gamma}_1=6.407$ &$\hat{\gamma}_1=-47.297$  & $\hat{\gamma}_1=-0.286$ & $\hat{\gamma}_1=-0.286$  &  $\hat{\gamma}_1=-0.286$ & $\hat{\gamma}_1=-0.118$   \\\hline\hline
\end{tabular} \end{Huge}  }
\end{subtable}\vspace{1mm}\\
 \begin{subtable}{.8\textwidth}
       \resizebox{1\textwidth}{!}{      
       \begin{Huge}
     \begin{tabular}{c|c|c|c|c|c}\hline
&   \multicolumn{5}{c}{Synthetic Data Sets} \\\cline{2-6}
Methods  & Set-7 & set-8 & Set-9 & Set-10 & Set-11 \\\hline
CM & $\hat{\beta}^c_0=5.702,\hat{\beta}^c_1=0.06$ & 
$\hat{\beta}^c_0=8.186,\hat{\beta}^c_1=-0.194$ & 
$\hat{\beta}^c_0=5.177,\hat{\beta}^c_1=-0.135$   & 
$\hat{\beta}^c_0=2.97,\hat{\beta}^c_1=0.406$ & 
$\hat{\beta}^c_0=5.436,\hat{\beta}^c_1=-0.071$ \\\hline

MinMax & $\hat{\beta}^-_0=4.738,\hat{\beta}^-_1=0.049$& 
$\hat{\beta}^-_0=7.736,\hat{\beta}^-_1=-0.34$ & 
$\hat{\beta}^-_0=4.272,\hat{\beta}^-_1=-0.152$   & 
$\hat{\beta}^-_0=3.603,\hat{\beta}^-_1=0.168$ & 
$\hat{\beta}^-_0=-1.202,\hat{\beta}^-_1=0.975$ \\
      & $\hat{\beta}^+_0=6.683,\hat{\beta}^+_1=0.064$ &  
      $\hat{\beta}^+_0=8.654,\hat{\beta}^+_1=-0.088$  & 
      $\hat{\beta}^+_0=5.528,\hat{\beta}^+_1=-0.026$  & 
      $\hat{\beta}^+_0=2.309,\hat{\beta}^+_1=0.592$ & 
      $\hat{\beta}^+_0=-0.961,\hat{\beta}^+_1=0.972$ \\\hline

CRM & $\hat{\beta}^c_0=5.702,\hat{\beta}^c_1=0.06$ &
$\hat{\beta}^c_0=8.186,\hat{\beta}^c_1=-0.194$ & 
$\hat{\beta}^c_0=5.177,\hat{\beta}^c_1=-0.135$   &
$\hat{\beta}^c_0=2.97,\hat{\beta}^c_1=0.406$ & 
$\hat{\beta}^c_0=5.436,\hat{\beta}^c_1=-0.071$ \\
      & $\hat{\beta}^r_0=1.168,\hat{\beta}^r_1=-0.118$ &  
      $\hat{\beta}^r_0=1.168,\hat{\beta}^r_1=-0.118$ & 
      $\hat{\beta}^r_0=0.104,\hat{\beta}^r_1=0.983$   & 
      $\hat{\beta}^r_0=0.104,\hat{\beta}^r_1=0.983$ & 
      $\hat{\beta}^r_0=0.104,\hat{\beta}^r_1=0.983$ \\\hline

CCRM & $\hat{\beta}^c_0=5.702,\hat{\beta}^c_1=0.06$ &
$\hat{\beta}^c_0=8.186,\hat{\beta}^c_1=-0.194$ & 
$\hat{\beta}^c_0=5.177,\hat{\beta}^c_1=-0.135$   &
$\hat{\beta}^c_0=2.97,\hat{\beta}^c_1=0.406$ & 
$\hat{\beta}^c_0=5.436,\hat{\beta}^c_1=-0.071$ \\
      & $\hat{\beta}^r_0=1.06,\hat{\beta}^r_1=0.0$ &  
      $\hat{\beta}^r_0=1.06,\hat{\beta}^r_1=0.0$ & 
      $\hat{\beta}^r_0=0.104,\hat{\beta}^r_1=0.983$   & 
      $\hat{\beta}^r_0=0.104,\hat{\beta}^r_1=0.983$ & 
      $\hat{\beta}^r_0=0.104,\hat{\beta}^r_1=0.983$ \\\hline

CIM  & $\hat{\beta}_0=5.716,\hat{\beta}_1=0.057$& 
$\hat{\beta}_0=7.47,\hat{\beta}_1=-0.081$ & 
$\hat{\beta}_0=5.14,\hat{\beta}_1=-0.128$   & 
$\hat{\beta}_0=3.554,\hat{\beta}_1=0.305$ & 
$\hat{\beta}_0=4.992,\hat{\beta}_1=0.001$ \\\hline    

PM & $\hat{\beta}^+_0=5.526,\hat{\alpha}^-_1=0.769$ &  
     $\hat{\beta}^+_0=8.54,\hat{\alpha}^-_1=-0.157$  &
     $\hat{\beta}^+_0=0.842,\hat{\alpha}^-_1=-1.408$ & $\hat{\beta}^+_0=3.458,\hat{\alpha}^-_1=0.338$   & $\hat{\beta}^+_0=7.086,\hat{\alpha}^-_1=0.283$  \\
       &$\hat{\omega}^-_1=-0.628,\hat{\beta}^+_0=7.992 $& $\hat{\omega}^-_1=-0.248,\hat{\beta}^+_0=7.561 $&
       $\hat{\omega}^-_1=1.513, \hat{\beta}^+_0=0.973 $ &
       $\hat{\omega}^-_1=-0.107, \hat{\beta}^+_0=3.934$ & 
       $\hat{\omega}^-_1=-0.643, \hat{\beta}^+_0=7.082$   \\
      & $\hat{\alpha}^+_1=1.342, \hat{\omega}^+_1=-1.081$ 
      & $\hat{\alpha}^+_1=0.431, \hat{\omega}^+_1=-0.258$ 
      & $\hat{\alpha}^+_1=-2.395,\hat{\omega}^+_1=2.512$ 
      & $\hat{\alpha}^+_1=-0.728,\hat{\omega}^+_1=0.897$   
      & $\hat{\alpha}^+_1=-0.682,\hat{\omega}^+_1=0.356 $ \\\hline

LM$_c$ & $\hat{\eta}= 5.591,\hat{\alpha}_1=0.997$ &
         $\hat{\eta}= 6.882,\hat{\alpha}_1=0.078$ &
         $\hat{\eta}= 0.804,\hat{\alpha}_1=-1.408$   & 
         $\hat{\eta}=3.593,\hat{\alpha}_1=0.297$ &
         $\hat{\eta}=6.98,\hat{\alpha}_1=0.292$  \\
         
       &$\hat{\beta}_1=-0.796,\hat{\theta}=2.12$ & 
       $\hat{\beta}_1=-0.194,\hat{\theta}=2.12$ &  
       $\hat{\beta}_1=1.521,\hat{\theta}=0.207$ & 
       $\hat{\beta}_1=-0.096,\hat{\theta}=0.207$ & 
       $\hat{\beta}_1=-0.634,\hat{\theta}=0.207$  \\
       
       & $\hat{\gamma}_1=0.0$ &  $\hat{\gamma}_1=0.0$ & 
       $\hat{\gamma}_1=0.983$   &  $\hat{\gamma}_1=0.983$ &
       $\hat{\gamma}_1=0.983$ \\\hline

LM$_w$ & $\hat{\eta}= 5.591,\hat{\alpha}_1=0.997$ &
         $\hat{\eta}= 6.882,\hat{\alpha}_1=0.078$ &
         $\hat{\eta}= 0.804,\hat{\alpha}_1=-1.408$   & 
         $\hat{\eta}=3.593,\hat{\alpha}_1=0.297$ &
         $\hat{\eta}=6.98,\hat{\alpha}_1=0.292$  \\
         
       &$\hat{\beta}_1=-0.796,\hat{\theta}=2.337$ & 
       $\hat{\beta}_1=-0.194,\hat{\theta}=2.337$ &  
       $\hat{\beta}_1=1.521,\hat{\theta}=0.207$ & 
       $\hat{\beta}_1=-0.096,\hat{\theta}=0.207$ & 
       $\hat{\beta}_1=-0.635,\hat{\theta}=0.207$  \\

       & $\hat{\gamma}_1=-0.118$ &  $\hat{\gamma}_1=-0.118$ &
       $\hat{\gamma}_1=0.983$   &  $\hat{\gamma}_1=0.983$ &
       $\hat{\gamma}_1=0.983$ \\\hline
\end{tabular}
\end{Huge} }
\end{subtable}
\label{tab:my_label11}
\end{table*}

\begin{table*}
\centering
\caption{Estimated coefficients of different regression methods for real-world IV data sets}
\resizebox{2\columnwidth}{!}{
\begin{tabular}{c|c|c|c}\hline
         &  Systolic-Diastolic Blood  & \multicolumn{2}{c}{Cyber-Security Vulnerability Assessment Data Set~\cite{miller2016modelling}} \\\cline{3-4}
Methods & Pressure data set~\cite{blanco2013set} & Evade-4 & Evade-22 \\\hline

CM &$\hat{\beta}^c_0=1.886,\hat{\beta}^c_1=0.438$&$\hat{\beta}^c_0=25.979,\hat{\beta}^c_1=0.029$, $\hat{\beta}^c_2=-0.289,\hat{\beta}^c_3=0.448$&$\hat{\beta}^c_0=60.214,\hat{\beta}^c_1=0.052$, $\hat{\beta}^c_2=-0.526,\hat{\beta}^c_3=0.146$ \\\hline

MinMax &$\hat{\beta}^-_0=-0.769,\hat{\beta}^-_1=0.591$& $\hat{\beta}^-_0=-19.068,\hat{\beta}^-_1=-0.020,\hat{\beta}^-_2=-0.253,\hat{\beta}^-_3=0.412$& $\hat{\beta}^-_0=-19.06,\hat{\beta}^-_1=0.089,\hat{\beta}^-_2=-0.212,\hat{\beta}^-_3=0.141$\\
     &$\hat{\beta}^+_0=5.301, \hat{\beta}^+_1=0.303$&$\hat{\beta}^+_0=26.608, \hat{\beta}^+_1=0.078$,$\hat{\beta}^+_2=-0.243, \hat{\beta}^+_3=0.472$&$\hat{\beta}^+_0=73.163, \hat{\beta}^+_1=0.174$,$\hat{\beta}^+_2=-0.657, \hat{\beta}^+_3=0.279$ \\\hline

CRM &$\hat{\beta}^c_0=1.886, \hat{\beta}^c_1=0.438$&$\hat{\beta}^c_0=25.979, \hat{\beta}^c_1=0.029,\hat{\beta}^c_2=-0.289, \hat{\beta}^c_3=0.448$& $\hat{\beta}^c_0=60.214, \hat{\beta}^c_1=.0523,\hat{\beta}^c_2=-0.526, \hat{\beta}^c_3=0.146$\\
     &$\hat{\beta}^r_0=1.621, \hat{\beta}^r_1=-0.246$&$\hat{\beta}^r_0=3.511, \hat{\beta}^r_1=-0.366,\hat{\beta}^r_2=0.414,\hat{\beta}^r_3=0.695$&$\hat{\beta}^r_0=-0.289, \hat{\beta}^r_1=0.445,\hat{\beta}^r_2=0.433,\hat{\beta}^r_3=0.341$ \\\hline

CCRM &$\hat{\beta}^c_0=1.886, \hat{\beta}^c_1=0.438$&$\hat{\beta}^c_0=25.979, \hat{\beta}^c_1=0.029,\hat{\beta}^c_2=-0.289, \hat{\beta}^c_3=0.448$&$\hat{\beta}^c_0=60.214, \hat{\beta}^c_1=0.0523,\hat{\beta}^c_2=-0.526, \hat{\beta}^c_3=0.146$ \\
     &$\hat{\beta}^r_0=2.056, \hat{\beta}^r_1=0.0$&$\hat{\beta}^r_0=2.056, \hat{\beta}^r_1=0.0,\hat{\beta}^r_2=0.341,\hat{\beta}^r_3=0.608$& $\hat{\beta}^r_0=0.0, \hat{\beta}^r_1=0.437,\hat{\beta}^r_2=0.432,\hat{\beta}^r_3=0.332$\\\hline

CIM &$\hat{\beta}_0=5.619, \hat{\beta}_1=0.185$&$\hat{\beta}_0=27.589, \hat{\beta}_1=0.025, \hat{\beta}_2=-0.245,\hat{\beta}_3=0.388$& $\hat{\beta}_0=56.199, \hat{\beta}_1=0.031, \hat{\beta}_2=-0.416,\hat{\beta}_3=0.106$\\\hline

PM &$\hat{\beta}^-_0=-0.096, \hat{\alpha}^-_1=0.664, \hat{\omega}^-_1=-0.082$&$\hat{\beta}^-_0=22.376, \hat{\alpha}^-_1=0.147, \hat{\omega}^-_1=-0.178$& $\hat{\beta}^-_0=63.771, \hat{\alpha}^-_1=0.128, \hat{\omega}^-_1=-0.081$\\
   &$\hat{\beta}^+_0=4.380, \hat{\alpha}^+_1=0.324, \hat{\omega}^+_1=0.154$&$\hat{\alpha}^-_2=-0.331, \hat{\omega}^-_2=0.089$,$\hat{\alpha}^-_3=0.472, \hat{\omega}^-_3=-0.074$&$\hat{\alpha}^-_2=0.077, \hat{\omega}^-_2=-0.620$, $\hat{\alpha}^-_3=0.179, \hat{\omega}^-_3=-0.075$ \\
   &&$\hat{\beta}^+_0=19.204, \hat{\alpha}^+_1=0.534, \hat{\omega}^+_1=-0.495$& $\hat{\beta}^+_0=53.141, \hat{\alpha}^+_1=-0.311, \hat{\omega}^+_1=0.439$\\
   &&$\hat{\alpha}^+_2=-0.737, \hat{\omega}^+_2=-0.491$, $\hat{\alpha}^+_3=-0.183, \hat{\omega}^+_3=0.657$&$\hat{\alpha}^+_2=-0.342, \hat{\omega}^+_2=	-0.213$,$\hat{\alpha}^+_3=-0.151, \hat{\omega}^+_3=0.328$ \\\hline

LM$_c$ &$\hat{\eta}=0.522, \hat{\alpha}_1=0.617,\hat{\beta}_1=-0.087$&$\hat{\eta}=7.279,\hat{\alpha}_1=0.158,\hat{\beta}_1=-0.153$& $\hat{\eta}=58.746,\hat{\alpha}_1=0.131,\hat{\beta}_1=-0.044$\\
& $\hat{\theta}=3.241,\hat{\gamma}_1=0.246$&$\hat{\alpha}_2=-0.327,\hat{\beta}_2=0.083$, $\hat{\alpha}_3=0.492,\hat{\beta}_3=-0.056$&$\hat{\alpha}_2=0.085,\hat{\beta}_2=-0.633$, $\hat{\alpha}_3=0.184,\hat{\beta}_3=-0.044$\\
& &$\hat{\theta}=4.112, \hat{\gamma}_1=0.0,\hat{\gamma}_2=0.3414, \hat{\gamma}_3=0.608$&$\hat{\theta}=0.0, \hat{\gamma}_1=0.437,\hat{\gamma}_2=0.432, \hat{\gamma}_3=0.332$\\\hline

LM$_w$ &$\hat{\eta}=0.522, \hat{\alpha}_1=0.617,\hat{\beta}_1=-0.087$&$\hat{\eta}=17.279,\hat{\alpha}_1=0.158,\hat{\beta}_1=-0.153$& $\hat{\eta}=58.746,\hat{\alpha}_1=0.131,\hat{\beta}_1=-0.044$\\
& $\hat{\theta}=3.241,\hat{\gamma}_1=0.246$&$\hat{\alpha}_2=-0.327,\hat{\beta}_2=0.083$, $\hat{\alpha}_3=0.492,\hat{\beta}_3=-0.056$&$\hat{\alpha}_2=0.085,\hat{\beta}_2=-0.632$, $\hat{\alpha}_3=0.184,\hat{\beta}_3=-0.044$\\
& &$\hat{\theta}=7.022, \hat{\gamma}_1=-0.366,\hat{\gamma}_2=0.414, \hat{\gamma}_3=0.696$&$\hat{\theta}=-0.579, \hat{\gamma}_1=0.445,\hat{\gamma}_2=0.433, \hat{\gamma}_3=0.341$\\\hline
\end{tabular}}
\vspace{1mm}\\
\resizebox{1.5\columnwidth}{!}{
\begin{tabular}{c|c|c}\hline
         & \multicolumn{2}{c}{Food  Snacks  Purchase  Intention  Data  Set~\cite{ellerby2020insights}} \\\cline{2-3}
Methods & Ethics and Overall purchase intention & Taste and Overall purchase intention\\\hline

CM & $\hat{\beta}^c_0=23.9,\hat{\beta}^c_1=0.497$ & $\hat{\beta}^c_0=25.395,\hat{\beta}^c_1=0.462$\\\hline

MinMax & $\hat{\beta}^-_0=32.164,\hat{\beta}^-_1=0.3283,\hat{\beta}^+_0=18.696,\hat{\beta}^+_1=0.756$ &  $\hat{\beta}^-_0=15.877,\hat{\beta}^-_1=0.57,\hat{\beta}^+_0=36.864,\hat{\beta}^+_1=0.342$\\\hline

CRM & $\hat{\beta}^c_0=23.9,\hat{\beta}^c_1=0.497, \hat{\beta}^r_0=3.165,\hat{\beta}^r_1=0.12$ & $\hat{\beta}^c_0=25.395,\hat{\beta}^c_1=0.462, \hat{\beta}^r_0=5.989,\hat{\beta}^r_1=0.274$\\\hline

CCRM & $\hat{\beta}^c_0=23.9,\hat{\beta}^c_1=0.497, \hat{\beta}^r_0=3.165,\hat{\beta}^r_1=0.12$ & $\hat{\beta}^c_0=25.395,\hat{\beta}^c_1=0.462, \hat{\beta}^r_0=5.989,\hat{\beta}^r_1=0.274$\\\hline

CIM &$\hat{\beta}_0=26.927, \hat{\beta}_1=0.444$ & $\hat{\beta}_0=26.005, \hat{\beta}_1=0.45$\\\hline

PM & $\hat{\beta}^-_0=15.269, \hat{\alpha}^-_1=0.055, \hat{\omega}^-_1=0.461$  & $\hat{\beta}^-_0=17.245, \hat{\alpha}^-_1=0.656, \hat{\omega}^-_1=-0.092$\\
 & $\hat{\beta}^+_0=18.582, \hat{\alpha}^+_1=-0.051, \hat{\omega}^+_1=0.616$ &$\hat{\beta}^+_0=38.472, \hat{\alpha}^+_1=0.264, \hat{\omega}^+_1=0.105$\\\hline

LM$_c$ & $\hat{\eta}=13.761,\hat{\alpha}_1=0.062,\hat{\beta}_1=0.478$ &  $\hat{\eta}=21.869,\hat{\alpha}_1=0.597,\hat{\beta}_1=-0.131$\\
& $\hat{\theta}=6.329, \hat{\gamma}_1=0.12$ &$\hat{\theta}=11.98, \hat{\gamma}_1=0.274$\\\hline

LM$_w$ & $\hat{\eta}=13.761,\hat{\alpha}_1=0.062,\hat{\beta}_1=0.478$ & $\hat{\eta}=21.869,\hat{\alpha}_1=0.597,\hat{\beta}_1=-0.131$\\
& $\hat{\theta}=6.329, \hat{\gamma}_1=0.12$ & $\hat{\theta}=11.98, \hat{\gamma}_1=0.274$\\\hline
\end{tabular}}
\label{tab:my_label22}
\end{table*}

\begin{figure*}
\centering
\includegraphics[width=0.3\textwidth]{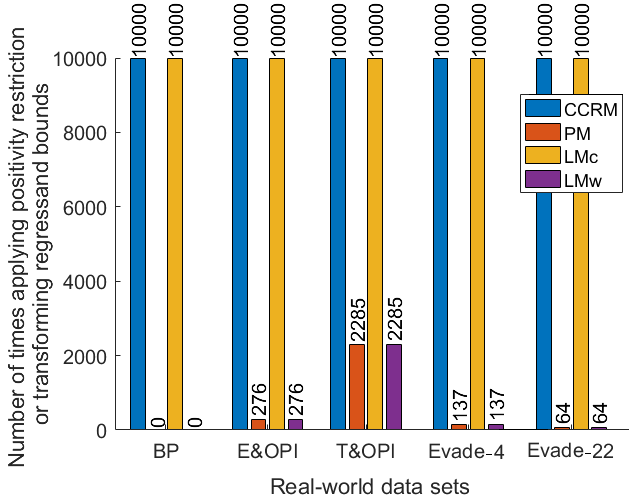}
\caption{Number of times in 10000 simulations the regression approaches apply positivity restriction/transformation regressand bounds for real-world cases. BP=blood pressure data set, E\&OPI=\emph{ethics} vs \emph{overall purchase intention} data set, T\&OPI=\emph{taste} vs \emph{overall purchase intention} data set.}
\label{fig:statistical analysis}
\end{figure*}

\vfill


\end{document}